\documentclass[10pt,twocolumn,letterpaper]{article}

\usepackage{cvpr}
\usepackage{times}
\usepackage{epsfig}
\usepackage{graphicx}
\usepackage{tabularx}
\usepackage{amsmath}
\usepackage{amssymb}
\usepackage{bm}
\usepackage{bbm}
\usepackage{algorithm}
\usepackage{algorithmic}
\usepackage{booktabs}
\usepackage{multirow}
\usepackage{tabularx}
\usepackage{xcolor}
\usepackage{subfigure}
\usepackage{stfloats}
\usepackage[toc,page]{appendix}
\colorlet{GRAY}{gray}

\usepackage{array}
\newcolumntype{?}{!{\vrule width 1pt}}
\newcolumntype{C}[1]{>{\centering\arraybackslash\hspace{0pt}}p{#1}}

\graphicspath{{images/}}

\usepackage{color}

\newcommand{\comment}[1]{}

\newif\ifdraft
\drafttrue
\ifdraft
 
 \newcommand{\BT}[1]{{\color{blue}BT:{\bf #1}}}
 \newcommand{\SU}[1]{{\color{cyan}SS:{\bf #1}}}

 \newcommand{\PF}[1]{{\color{red}PF:{\bf #1}}}
\else
 
 \newcommand{\BT}[1]{}
 
 \newcommand{\SU}[1]{}
 
 \newcommand{\PF}[1]{}
\fi

\newcommand{\bx}[0]{\mathbf{x}}

\newcommand{\bv}{\mathbf{v}}

\newcolumntype{M}[1]{>{\centering\arraybackslash}m{#1}}
\newcolumntype{R}[1]{>{\raggedleft\arraybackslash}m{#1}}
\newcolumntype{P}[1]{>{\centering\arraybackslash}p{#1}}

\definecolor{mygray}{gray}{0.2}

\usepackage[pagebackref=true,breaklinks=true,letterpaper=true,colorlinks,bookmarks=false]{hyperref}

\cvprfinalcopy
\def\cvprPaperID{3117} % *** Enter the CVPR Paper ID here

\ifcvprfinal\fi
\begin{document}

%%%%%%%%% TITLE
%Commented out by Sudipta \title{YOLO-POSE: Real-time 6 DOF Pose Estimation from a Single RGB Image}
% proposed by Sudipta, SS for single-shot and bb8 to indicate we predict bounding box corners .. but the name sounds odd!
%\title{SSBB8: Real-time 6D Object Detection and Pose Estimation in a Single RGB Image}
%\title{SSBB8: Real-time 6D Object Pose Estimation in a Single RGB Image}
%\title{Fast 6D Object Detection and Pose Estimation in a Single RGB Image}
%\title{SSP-Net: Real-time Single Shot 6D Object Pose Estimation from an RGB Image}
\title{Real-Time Seamless Single Shot 6D Object Pose Prediction}
%
%\author{Bugra Tekin \quad\quad\quad\quad Sudipta N. Sinha \quad\quad\quad\quad Pascal Fua \\ 
%	EPFL\quad\quad\quad\quad\quad\quad Microsoft Research \quad\quad\quad\quad EPFL\\
%	{\tt\scriptsize \{bugra.tekin\}@epfl.ch  \quad\quad \{sudipta.sinha\}@microsoft.com \quad\quad \{pascal.fua\}@epfl.ch }
%}

\author{Bugra Tekin\\
	EPFL\\
	{\tt\footnotesize bugra.tekin@epfl.ch}
	% For a paper whose authors are all at the same institution,
	% omit the following lines up until the closing ``}''.
	% Additional authors and addresses can be added with ``\and'',
	% just like the second author.
	% To save space, use either the email address or home page, not both
	\and
	Sudipta N. Sinha\\
	Microsoft Research\\
	{\tt\footnotesize sudipta.sinha@microsoft.com}
	\and
	Pascal Fua\\
	EPFL\\
	{\tt\footnotesize pascal.fua@epfl.ch}
}

\maketitle
%% -*- mode: latex; mode: reftex; mode: flyspell; mode: auto-fill; TeX-master: "top.tex"; -*-
% !TEX root = top.tex
% !TEX spellcheck = en-US

%\bt{What about: ``Real-time \textbf{S}ingle \textbf{P}ass 6D Object \textbf{P}ose Estimation from an RGB Image'' - We can maybe call it, in short, \textbf{SPP}. I think it might be good to highlight that it is real-time.\\}

%\su{Indeed, sounds better. but in this case, we cannot have both phrases "object detection and "pose estimation" in the title. probably ok, since half of the experiments are on object-specific networks.\\}

%\su{based on your previous suggestion. How about: "SSP-Net: Real-time \textbf{S}ingle \textbf{S}hot 6D Object \textbf{P}ose Estimation from an RGB Image" ?.  I think {\it single-shot} and {\it single-staged} are more common than {\t single-pass}.?\\}

\begin{abstract}

We propose a single-shot approach for simultaneously detecting an object in an RGB image and predicting its 6D pose {\it without} requiring multiple stages or having to examine multiple hypotheses. Unlike 
%an existing
a recently proposed
single-shot technique for this task~\cite{kehl2017} that only predicts an approximate 6D pose that must then be refined, ours is accurate enough not to require additional post-processing. As a result, it is much faster -- 50 fps on a Titan X (Pascal) GPU -- and more suitable for real-time processing.
The key component of our method is a new CNN architecture inspired by~\cite{redmon2016a,redmon2016b} that directly predicts the 2D image locations of the projected vertices of the object's 3D bounding box. The object's 6D pose is then estimated using a PnP algorithm.

For single object and multiple object pose estimation on the \textsc{LineMod} and  \textsc{Occlusion} datasets, our approach substantially outperforms other recent CNN-based approaches~\cite{kehl2017,rad2017} when they are all used without post-processing. During post-processing,
%classical methods
a pose refinement step can be used to boost the accuracy of these two methods, but at 10 fps or less, they are much slower than our method.
%~\cite{rad2017,kehl2017} but at the cost of being then
%but this also makes them run at 10 fps or less making them much slower than our method.
%It runs at 50fps on a Titan X GPU whereas the other two run at 10 or less.

\end{abstract} 
\vspace{4mm}
%%%%%%%%% BODY TEXT
%% -*- mode: latex; mode: reftex; mode: flyspell; mode: auto-fill; TeX-master: "top.tex"; -*-
% !TEX root = top.tex
% !TEX spellcheck = en-US

\section{Introduction}

Real-time object detection and 6D pose estimation is crucial for augmented reality,
virtual reality, and robotics. Currently, methods relying on depth data acquired by
RGB-D cameras are quite robust~\cite{brachmann2014,choi2012,choi2016,kehl2016,lai2011}.
However, active depth sensors are power hungry, which makes  6D object detection methods
for passive RGB images more attractive for mobile and wearable cameras. There are many
fast keypoint and edge-based methods~\cite{lowe1999, rothganger2006, wagner2008} that are
effective for textured objects. However, they have difficulty handling weakly textured
or untextured objects and processing low-resolution video streams, which are quite common
when dealing with cameras on wearable devices.

Deep learning techniques have recently been used to address these limitations~\cite{kehl2017,rad2017}.
BB8~\cite{rad2017} is a 6D object detection pipeline made of one CNN to coarsely segment the object and another to predict the 2D locations of the projections of the object's 3D bounding box given the segmentation, which are then used to compute the 6D pose using a PnP algorithm~\cite{pnp}. The method is effective but slow due to its multi-stage nature. SSD-6D~\cite{kehl2017} is a different pipeline that relies on the SSD architecture~\cite{liu2016} to predict 2D bounding boxes and a very rough estimate of the object's orientation in a single step. This is followed by an approximation to predict the object's depth from the size of its 2D bounding box in the image, to lift the  2D detections to 6D. Both BB8 and SSD-6D require a further pose refinement step for improved accuracy, which increases their running times linearly with the number of objects being detected.
%tracked.

In this paper, we propose a single-shot deep CNN architecture that takes the image as input and directly detects the 2D projections of the 3D bounding box vertices. It is end-to-end trainable and accurate even without any {\it a posteriori} refinement. And since, we do not need this refinement step, we also do not need a precise and detailed textured 3D object model that is needed by other methods ~\cite{kehl2017,rad2017}. We only need the 3D bounding box of the object shape for training. This can be derived from other easier to acquire and approximate 3D shape representations.

We demonstrate state-of-the-art accuracy on the \textsc{LineMod} dataset~\cite{hinterstoisser2012}, which has become a {\it de facto} standard benchmark for 6D pose estimation. However, we are much faster than the competing techniques by a factor of more than five, when dealing with a single object. Furthermore, we pay virtually no time-penalty when handling several objects and our running time remains constant whereas that of other methods grow proportional to the number of objects, which we demonstrate on the \textsc{Occlusion} dataset~\cite{brachmann2014}.

Therefore, our contribution is an architecture that yields a fast and accurate one-shot 6D pose prediction without requiring any post-processing. It extends single shot CNN architectures for 2D detection in a seamless and natural way to the 6D detection task.
Our implementation is based on YOLO~\cite{redmon2016b} but the approach is amenable to other single-shot detectors such as SSD~\cite{liu2016} and its variants.

%% -*- mode: latex; mode: reftex; mode: flyspell; mode: auto-fill; TeX-master: "top.tex"; -*-
% !TEX root = top.tex
% !TEX spellcheck = en-US

\section{Related Work}

We now review existing work on 6D pose estimation ranging from classical feature and template matching methods to newer end-to-end trainable CNN-based methods.

\vspace{-3mm}
\paragraph{Classical methods.}

Traditional RGB object instance recognition and pose estimation works used local keypoints and feature matching. Local descriptors needed by such methods were designed for invariance to changes in scale, rotation, illumination and viewpoints~\cite{lowe1999, rothganger2006, wagner2008}. Such methods are often fast and robust to occlusion and scene clutter. However, they only reliably handle textured objects in high resolution images~\cite{lepetit2005}. Other related methods include 3D model-based registration~\cite{li2011,lowe1991,vacchetti2004}, Hausdorff matching~\cite{huttenlocher1993}, oriented Chamfer matching for edges~\cite{liu2010} and 3D chamfer matching for aligning 3D curve-based models to images~\cite{ramnath2014}.

\vspace{-3mm}
\paragraph{RGB-D methods.}

The advent of commodity depth cameras has spawned many RGB-D object pose estimation methods~\cite{brachmann2014,choi2012,choi2016,kehl2016,lai2011b,michel2016,sock2017,zhang2017}. For example,
Hinterstoisser proposed template matching algorithms suitable for both color and depth images~\cite{hinterstoisser2011,hinterstoisser2012}. Rios et al.~\cite{rios2013} extended their work using discriminative learning and cascaded detections for higher accuracy and efficiency respectively. RGB-D methods were used on indoor robots for 3D object recognition, pose estimation, grasping and manipulation~\cite{choi2012,choi2016,collet2011,lai2011,lai2011b,zhu2014}. Brachmann et al.~\cite{brachmann2014} proposed using regression forests to predict dense object coordinates, to segment the object and recover its pose from dense correspondences. They also extended their method to handle uncertainty during inference and deal with RGB images~\cite{brachmann2016}. Zach et al.~\cite{zach2015} explored fast dynamic programming based algorithms for RGB-D images.

\vspace{-3mm}
\paragraph{CNN-based methods.}

In recent years, research in most pose estimation tasks has been dominated by CNNs.
Techniques such as Viewpoints and Keypoints~\cite{tulsiani2015} and Render for CNN~\cite{su2015}
cast object categorization and 3D pose estimation into classification tasks, specifically by discretizing the pose space.
In contrast, PoseNet~\cite{kendall2015} proposes using a CNN to directly regress from an RGB image to a 6D pose, albeit for camera pose estimation, a slightly different task. Since PoseNet outputs a translational and a rotational component, the two associated loss terms have to be balanced carefully by tuning a hyper-parameter during training.

To avoid this problem, the newer PoseCNN architecture~\cite{xiang2017} is trained to predict 6D object pose from a single RGB image in multiple stages, by decoupling the translation and rotation predictors. A geodesic loss function more suitable for optimizing over 3D rotations have been suggested in~\cite{mahendran2017}. Another way to address this issue has recently emerged. In~\cite{kehl2017,rad2017}, the CNNs do not directly predict object pose. Instead, they output 2D coordinates, 2D masks, or discrete orientation predictions from which the 6D pose can be inferred. Because all the predictions are in the 2D image, the problem of weighting different loss terms goes away. Also training becomes numerically more stable, resulting in better performance on the~\textsc{LineMod} dataset~\cite{hinterstoisser2012}. We also adopt this philosophy in our work.

In parallel to these developments, on the 2D object detection task, there has been a progressive trend towards single shot CNN frameworks as an alternative to two-staged methods such as Faster-RCNN~\cite{ren2015} that first find a few candidate locations in the image and then classifies them as objects or background. Recently, single shot architectures such as YOLO~\cite{redmon2016a,redmon2016b} and SSD~\cite{liu2016} have been shown to be fast and accurate. SSD has been extended to predict the object's identity, its 2D bounding box in the image and a discrete estimate of the object's orientation~\cite{kehl2017,poirson2016}.
In this paper, we go beyond such methods by extending a YOLO-like architecture~\cite{redmon2016b} to directly predict a few 2D coordinates from which the full 6D object pose can be accurately recovered.

%% -*- mode: latex; mode: reftex; mode: flyspell; mode: auto-fill; TeX-master: "top.tex"; -*-
% !TEX root = top.tex
% !TEX spellcheck = en-US

\section{Approach}
\label{sec:approach}

\begin{figure*}[t]
	\centering
	\scalebox{0.85}{
		\begin{tabular}{c}
			\includegraphics[width=\linewidth]{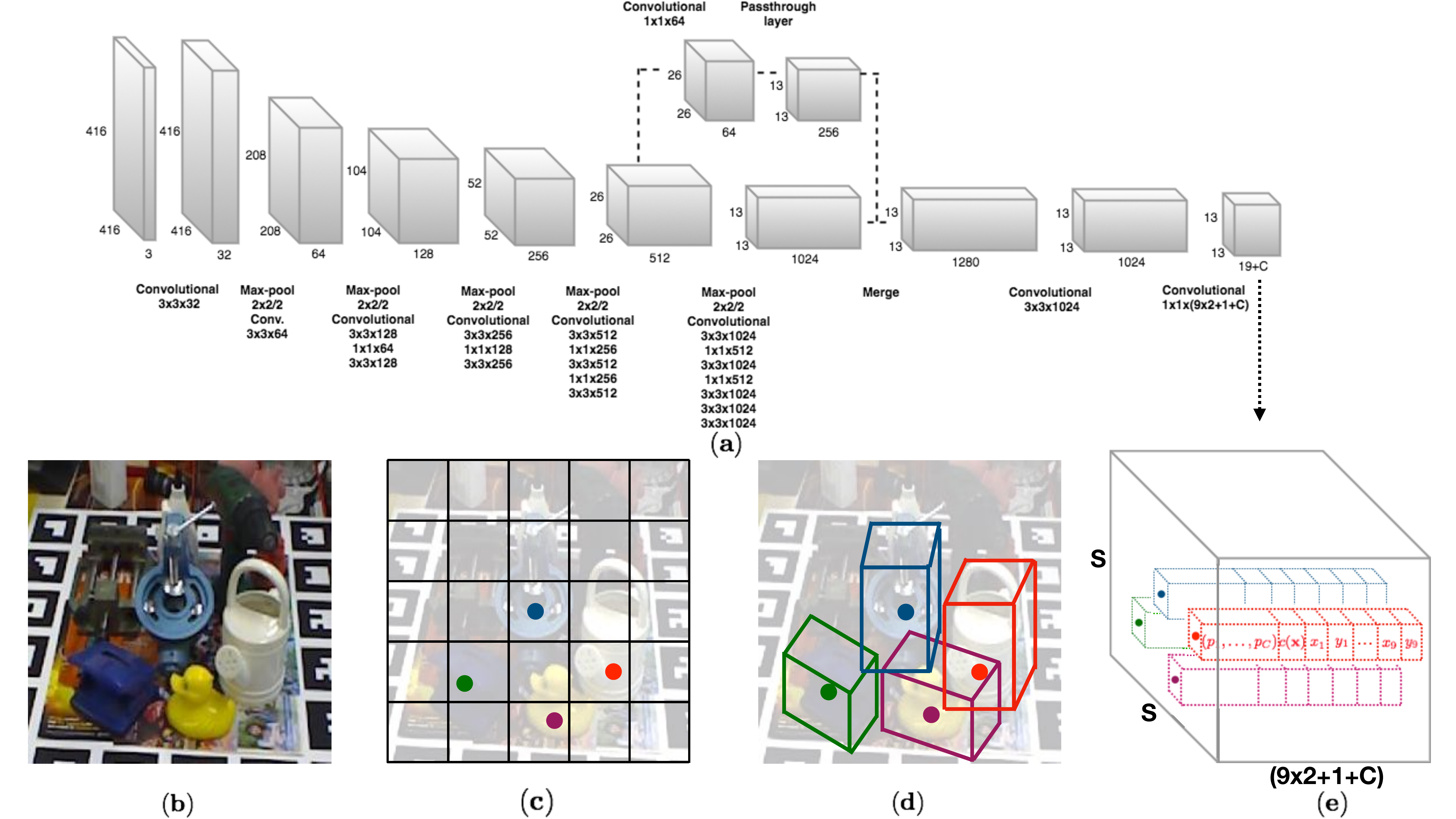}
		\end{tabular}
	} 	\\
	\vspace{-0.1mm}
	\caption{Overview: (a) The proposed CNN architecture. (b) An example input image with four objects. (c) The $S \times S$ grid showing cells responsible for detecting the four objects. (d) Each cell predicts 2D locations of the corners of the projected 3D bounding boxes in the image. (e) The 3D output tensor from our network, which represents for each cell a vector consisting of the 2D corner locations, the class probabilities and a confidence value associated with the prediction.}
	\vspace{-5mm}
	\label{fig:overview}
\end{figure*}

With our goal of designing an end-to-end trainable network that predicts the 6D pose in real-time,
we were inspired by the impressive performance of single shot 2D object detectors
such as YOLO~\cite{redmon2016a,redmon2016b}. This led us to design the CNN architecture~\cite{redmon2016a,redmon2016b} shown in Fig.~\ref{fig:overview}.
We designed our network to predict the 2D projections of the corners of the 3D bounding box around our objects. The main insight was that YOLO was
originally designed to regress 2D bounding boxes and to predict the projections of the 3D bounding box corners in the image, a few more 2D points
had to be predicted for each object instance in the image.
Then given these 2D coordinates and the 3D ground control points for the bounding box corners, the 6D pose can be calculated algebraically with an efficient PnP algorithm~\cite{pnp}.
BB8~\cite{rad2017} takes a similar approach. However, they first find a 2D segmentation mask around the object and present a cropped image to a second network that predicts the
eight 2D corners in the image. We now describe our network architecture and explain various aspects of our approach in details.

\subsection{Model}
\label{subsec:model}

We formulate the 6D pose estimation problem in terms of predicting the 2D image coordinates of virtual 3D control
points associated with the 3D models of our objects of interest. Given the 2D coordinate predictions, we calculate
the object's 6D pose using a PnP algorithm. We parameterize the 3D model of each object with 9 control points.
For these control points, we select the 8 corners of the tight 3D bounding box fitted to the 3D model, similar to~\cite{rad2017}.
In addition, we use the centroid of the object's 3D model as the 9th point. This parameterization is general
and can be used for any rigid 3D object with arbitrary shape and topology. In addition, these 9 control
points are guaranteed to be well spread out in the 2D image and could be semantically meaningful for many
man-made objects.

Our model takes as input a single full color image, processes it with a fully-convolutional architecture
shown in Figure~\ref{fig:overview}(a) and
divides the image into a 2D regular grid containing $S \times S$ cells as shown in Figure~\ref{fig:overview}(c).
In our model, each grid location in the 3D output tensor will be associated with a multidimensional vector, consisting of
predicted 2D image locations of the $9$ control points, the class probabilities of the object and an overall confidence value.
At test time, predictions at cells with low confidence values, ie. where the objects
of interest are not present, will be pruned.

The output target values for our network are stored in a 3D tensor of size $S \times S \times D$ visualized in
Fig.~\ref{fig:overview}(e). The target values for an object
at a specific spatial cell location $i \in S \times S$ is placed in the $i$-th cell in the 3D tensor in the form of a $D$ dimensional vector $\bv_i$.
When $N$ objects are present in different cells, we have $N$ such vectors, $\bv_1, \bv_2, \dots, \bv_n$ in the 3D tensor.
We train our network to predict these target values.
The $9$ control points in our case are the 3D object model's center and bounding box corners but could be defined in other ways as well.
To train our network, we only need to know the 3D bounding box of the object, not a detailed mesh or an associated texture map.

\vspace{-0.5mm}

\begin{figure}[t]
	\centering
	\scalebox{0.85}{
		\begin{tabular}{c}
			\includegraphics[width=\columnwidth]{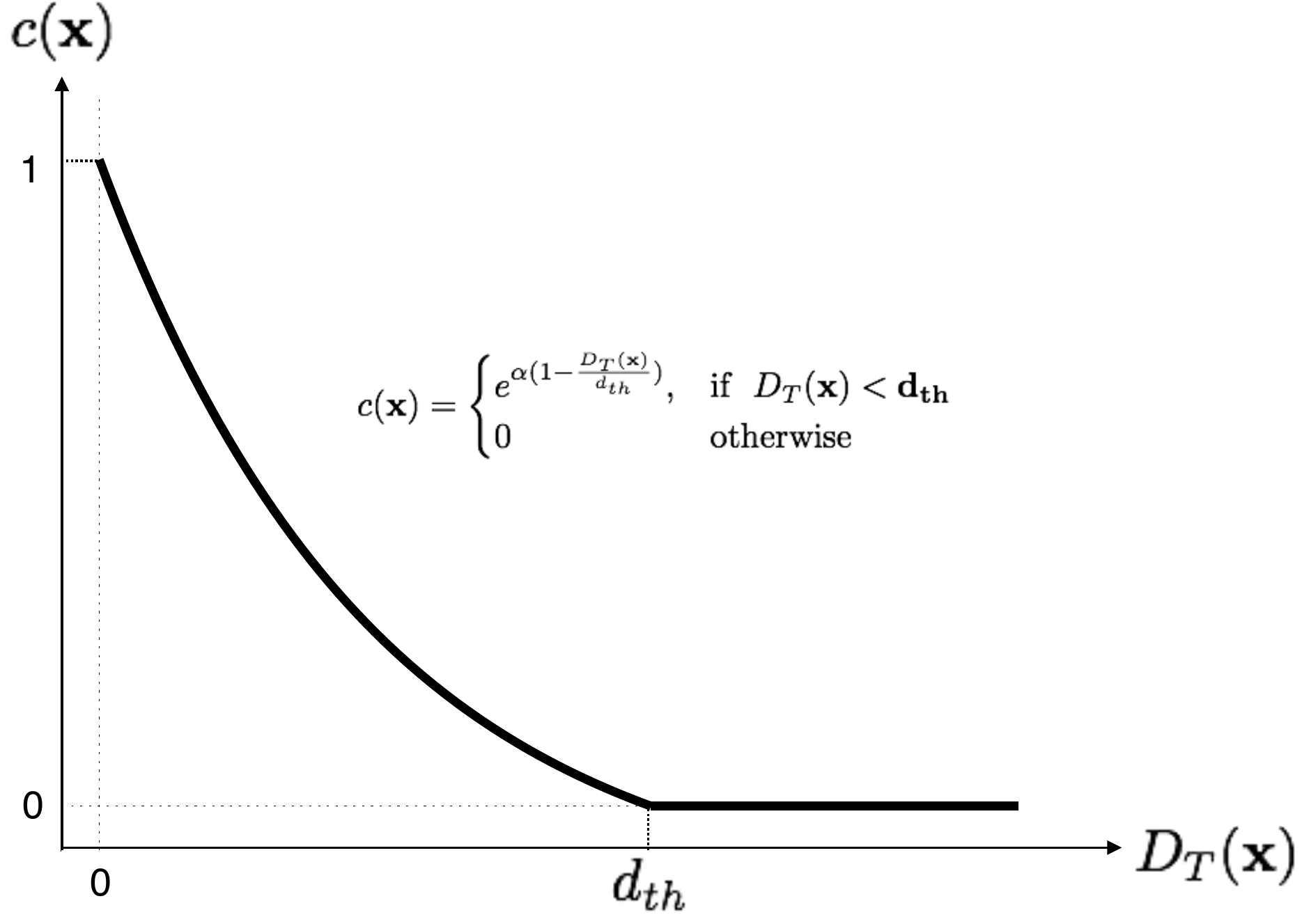}
		\end{tabular}
	} 	\\
	\vspace{-1mm}
	\caption{Confidence $c(\bx)$ as a function of the distance $D_T(\bx)$ between a predicted point and the true point.}
	\vspace{-5mm}
	\label{fig:conffunction}
\end{figure}

As in YOLO, it is crucial that a trained network is able to predict not only the precise 2D locations but also high confidence values
in regions where the object is present and low confidence where it isn't present.
In case of 2D object detection, YOLO uses for its confidence values, an intersection over union (IoU) score associated with the predicted (and true 2D rectangles)
in the image. In our case, the objects are in 3D and to compute an equivalent IoU score with two arbitrary cuboids, we would need to
calculate a 3D convex hull corresponding to their intersections. This would be tedious and would slow down training, as also analyzed in our supplemental material. Therefore, we take a different approach. We
model the predicted confidence value using a confidence function shown in Figure~\ref{fig:conffunction}.
The confidence function, $c(\bx)$, returns a confidence value for a predicted 2D point denoted by $\bx$ based on its distance
$D_T(\bx)$ from the ground truth i.e. target 2D point. Formally, we define the confidence function $c(\bx)$ as follows:

\begin{equation}
c(\bx)=
\begin{cases}
e^{\alpha (1 - \frac{D_T(\bx)}{d_{th}} )},& \text{if} ~~ D_T(\bx) < d_{th}  \\
0              & \text{otherwise}
\end{cases}
\label{eq:conffunction}
\end{equation}

The distance $D_T(\bx)$ is defined as the 2D Euclidean distance in the image space. To achieve precise localization with this function,
we choose a sharp exponential function with a cut-off value $d_{th}$ instead of a monotonically decreasing
linear function. The sharpness of the exponential function is defined by the parameter $\alpha$. In practice, we apply the confidence
function to all the control points and calculate the mean value and assign it as the confidence.
As mentioned earlier, we also predict $C$ conditional class probabilities at each cell. The class probability is conditioned on the cell containing an object.
Overall, our output 3D tensor depicted in Figure~\ref{fig:overview}(\textbf{e}) has dimension $S \times S \times D$, where the 2D spatial grid corresponding to the image
dimensions has $S \times S$ cells and each such cell has a $D$ dimensional vector.
Here, $D = 9 \times 2 + C + 1$, because we have 9 $(x_i, y_i)$ control points, $C$ class probabilities and one confidence value.

Our network architecture follows the fully convolutional YOLO v2 architecture~\cite{redmon2016b}. Thus, our network has 23 convolutional layers and 5 max-pooling layers. Similar to YOLO v2, we choose $S = 13$ and have a $13 \times 13$ 2D spatial grid on which we make our predictions.
We also allow higher layers of our network to use fine-grained features by adding a passthrough
layer. Specifically, we bring features from an earlier layer at resolution $26 \times 26$, apply batch normalization and resize the input image
during training on-the-fly. As the network downsamples the image by a factor of $32$, we change the input resolution to a multiple
of $32$ randomly chosen from the set $\{320, 352, \dots, 608\}$ to be robust to objects of different size.

\vspace{-1mm}
\subsection{Training Procedure}
\vspace{-1mm}

Our final layer outputs class probabilities, $(x, y)$ coordinate locations for the control points, and the
overall confidence score. During training, this confidence value is computed on the fly using the
function defined in Eq.~\ref{eq:conffunction} to measure the distance between the current coordinate predictions and the ground-truth, $D_T(\bx)$.
We predict offsets for the 2D coordinates with respect to $(c_x, c_y)$, the top-left corner of the associated grid cell.
For the centroid, we constrain this offset to lie between $0$ and $1$. However, for the corner points, we do not constrain the
network's output as those points should be allowed to fall outside the cell. The predicted control point ($g_x, g_y$) is defined as
\begin{align}
& g_x = f(x) + c_x \\
& g_y = f(y) + c_y
\label{eq:offsets}
\end{align}
\noindent where $f(\cdot)$ is chosen to be a 1D sigmoid function in case of the centroid and the identity function in case of the eight corner points.
This has the effect of forcing the network to first
find the approximate cell location for the object and later refine its eight corner locations. We minimize the following loss function to train
our complete network.

\begin{equation}
\mathcal{L} = \lambda_{pt} \mathcal{L}_{pt} + \lambda_{conf} \mathcal{L}_{conf} + \lambda_{id} \mathcal{L}_{id}
\label{eq:loss}
\end{equation}

\noindent Here, the terms $\mathcal{L}_{pt}$, $\mathcal{L}_{conf}$ and $\mathcal{L}_{id}$ denote the coordinate, confidence and the classification loss, respectively.
We use mean-squared error for the coordinate and confidence losses, and cross
entropy for the classification loss. As suggested in~\cite{redmon2016a}, we downweight the confidence
loss for cells that don't contain objects by setting $\lambda_{conf}$ to $0.1$. This improves model stability. For cells that
contain objects, we set $\lambda_{conf}$ to $5.0$. We set~$\lambda_{pt}$ and~$\lambda_{id}$ simply to $1$.

When multiple objects are located close to each other in the 3D scene,
they are more likely to appear close together in the images or be occluded by each other.
In these cases, certain cells might contain multiple objects. To be able to predict the pose of such multiple objects that lie in the
same cell, we allow up to $5$ candidates per cell and therefore predict five sets of control points per cell. This essentially means that
we assumed that at most 5 objects could occlude each other in a single grid cell. This is a reasonable assumption to make in practical pose estimation scenarios. As in~\cite{redmon2016b}, we precompute with k-means, five anchor boxes that define the size, ie. the width and height of a 2D rectangle tightly fitted to
a masked region around the object in the image. During training, we assign whichever anchor box has the most similar size
to the current object as the responsible one to predict the 2D coordinates for that object.

\subsection{Pose Prediction}

We detect and estimate the pose of objects in 6D by invoking our network only once. At test time, we estimate the class-specific
confidence scores for each object by multiplying the class probabilities and the score returned by the confidence
function. Each grid cell produces predictions in one network evaluation and cells with predictions with low confidence are pruned using
a confidence threshold. For large objects and objects whose projections lie at the intersection of two cells, multiple cells are likely
to predict highly confident detections. To obtain a more robust and well localized pose estimate, we inspect the cells in the $3 \times 3$
neighborhood of the cell which has the maximum confidence score. We combine the individual corner predictions of these adjacent cells by
computing a weighted average of the individual detections, where the weights used are the confidence scores of the associated cells.

At run-time, the network gives the 2D projections of the object's centroid and corners of
its 3D bounding box along with the object identity. We estimate the 6D pose from the correspondences between
the 2D and 3D points using a Perspective-n-Point (PnP) pose estimation method~\cite{pnp}. In our case, PnP uses only 9 such control
point correspondences and provides an estimate of the 3D rotation $\bf{R}$ and 3D translation $\bf{t}$ of the object in camera coordinates.

%% -*- mode: latex; mode: reftex; mode: flyspell; mode: auto-fill; TeX-master: "top.tex"; -*-
% !TEX root = top.tex
% !TEX spellcheck = en-US

\vspace{-2mm}

\section{Implementation Details}

We initialize the parameters of our network by training the original network on the
ImageNet classification task. As the pose estimates in the early stages of training
are inaccurate, the confidence values computed using Eq.~\ref{eq:conffunction} are
initially unreliable. To remedy this, we pretrain our network parameters by
setting the regularization parameter for confidence to $0$. Subsequently, we train
our network by setting $\lambda_{conf}$ to 5 for the cells that contain an object, 
and to 0.1 otherwise, to have more reliable confidence estimates in the early stages 
of the network. In practice, we set the sharpness of the confidence function $\alpha$ 
to 2 and the distance threshold to $30$ pixels. We use
stochastic gradient descent for optimization. We start with a learning rate of $0.001$
and divide the learning rate by $10$ at every $100$ epochs. To avoid overfitting, we
use extensive data augmentation by randomly changing the hue, saturation and exposure
of the image by up to a factor of $1.5$. We also randomly scale and translate the image
by up to a factor of $20\%$ of the image size. Our implementation is based on PyTorch. We will
make our code publicly available for the sake of reproducibility.

\section{Experiments}
\label{sec:results}

\vspace{-2mm}

We first evaluate our method for estimating the 6D pose of single objects
and then we evaluate it in the case where multiple objects are present in the image.
We use the same datasets and evaluation
protocols as in~\cite{brachmann2016,kehl2017,rad2017}, which we review below. We then
present and compare our results to the state of the art methods.

\subsection{Datasets}
\label{sec:datasets}

We test our approach on two datasets that were designed explicitly to benchmark 6D object pose estimation
algorithms. We describe them briefly below.

\vspace{-3mm}
\paragraph{LineMod~\cite{hinterstoisser2012}} has become a {\it de facto} standard benchmark for 6D object pose
estimation of textureless objects in cluttered scenes. The central object in each RGB image  is assigned a ground-truth
rotation, translation, and ID. A full 3D mesh representing the object is also provided. There are $15783$ images
in~\textsc{LineMod} for $13$ objects. Each object features in about $1200$ instances.

\vspace{-3mm}
\paragraph{OCCLUSION~\cite{brachmann2014}} is a multi-object detection and pose estimation
dataset that contains additional annotations for {\it all} objects in a subset of the \textsc{LineMod} images.
As its name suggests, several objects in the images are severely occluded due to scene clutter, which
makes pose estimation extremely challenging. With the exception of~\cite{kehl2017,rad2017}, it has
primarily been used  to test algorithms that require depth images.

\subsection{Evaluation Metrics}
\label{subsec:eval}

We use three standard metrics to evaluate 6D pose accuracy, namely -- 2D reprojection error, IoU score and average 3D distance of model vertices (referred to as ADD metric) as in~\cite{brachmann2016,kehl2017,rad2017}. In all cases, we calculate the accuracy as
the percentage of {\it correct} pose estimates for certain error thresholds.

When using the reprojection error, we consider a pose estimate to be correct when the  mean distance between
the 2D projections of the object's 3D mesh vertices using the estimate and the ground truth pose
is less than 5 pixels~\cite{brachmann2016}. This measures the closeness of the true image projection of the
object to that obtained by using the estimated pose. This metric is
suitable for augmented reality applications. 

To compute  the IoU score, we measure the overlap between the projections of the 3D model given
the ground-truth and predicted pose and accept a pose as correct if the overlap is  larger than $0.5$, as in~\cite{kehl2017}.

\begin{table}[t]
	\begin{center}
		\scalebox{0.8}{
			\begin{tabular}{|c|ccc|cc|}
				\hline
				\textbf{Method} & \multicolumn{3}{c|}{w/o Refinement}   							            							 & \multicolumn{2}{c|}{w/ Refinement} 			 \\ \hline
				& Brachmann				     	 		 & BB8 			   			   &  OURS				& Brachmann       	   & BB8 				    			\\
				\textbf{Object} & \cite{brachmann2016}  			  	 & \cite{rad2017} 	  		   &  			 		& \cite{brachmann2016} & \cite{rad2017} 		       \\ \hline
				Ape		    	& -                   		 	 		 & \textbf{\emph{95.3}}        & 92.10   			& 85.2       	 	  &    \textbf{96.6}     					   \\
				Benchvise    	& -          			 		  	     & 80.0            	 		   & \textbf{95.06}     & 67.9                &    90.1                           \\
				Cam    			& -	                	 		  		 & 80.9                        & \textbf{93.24}     & 58.7                &    86.0                            \\
				Can    			& -                   	 				 & 84.1                        & \textbf{97.44}& 70.8                     & 91.2                        	\\
				Cat    			& -	                     		  		 & 97.0                        & \textbf{97.41}     & 84.2                &    98.8                         \\
				Driller         & -                   	 		 		 & 74.1	                       & \textbf{\emph{79.41}}& 73.9              & \textbf{80.9}                        \\
				Duck        	& -		                     		     & 81.2                        & \textbf{94.65}     & 73.1                &    92.2                       \\
				Eggbox        	& -		                 		  		 & 87.9                        & \textbf{90.33}     & 83.1                &    91.0                      \\
				Glue        	& -                   		 		     & 89.0                        & \textbf{96.53}     & 74.2                &    92.3                      \\
				Holepuncher     & -                  	 		  		 & 90.5                        & \textbf{92.86}     & 78.9                &    95.3                         \\ 		
				Iron        	& -                   	 		 		 & 78.9                        & \textbf{\emph{82.94}}& 83.6              &  \textbf{84.8}                          \\			
				Lamp        	& -                       		 		 & 74.4                        & \textbf{76.87}     & 64.0                &    75.8                         \\ 						
				Phone        	& -                   	 		  		 & 77.6                        & \textbf{86.07}		& 60.6                &    85.3                            \\ \hline
				Average         & 69.5                       		     & 83.9                        & \textbf{90.37}     & 73.7                &    89.3                            \\ \hline					
			\end{tabular}
		}
	\end{center}
	\vspace{-3mm}
	\caption{Comparison of our approach with state-of-the-art algorithms on \textsc{LineMod} in terms of 2D reprojection error. We report percentages of correctly estimated poses. In Tables~\ref{tab:projerror},~\ref{tab:vertexerror} and~\ref{tab:iouerror}~\textbf{bold face} numbers denote the best overall methods, \textbf{\emph{bold italic}}
		numbers denote the best methods among those that do not use refinement as opposed to the ones that use, if different.
		Note that even though we do not rely on the knowledge of a detailed 3D object model our method consistently outperforms the baselines.}
	\vspace{-7mm}
	\label{tab:projerror}
\end{table}

When comparing 6D poses using the ADD metric, we take a pose estimate to be correct if the mean distance between the
true coordinates of 3D mesh vertices and those estimated given the pose is less than $10\%$
of the object's diameter~\cite{hinterstoisser2012}. For most objects, this is approximately a 2cm threshold
but for smaller objects, such as {\it ape}, the threshold drops to about 1cm. For rotationally symmetric objects
whose pose can only be computed up to one degree of rotational freedom, we modify slightly the metric as
in~\cite{brachmann2016,hinterstoisser2012} and compute
\begin{equation}
s = \frac{1}{|\mathcal{M}|} \sum_{x_1 \in \mathcal{M}} \min_{\mathcal{M}} \|  (\bf{R} \bf{x} + \bf{t}) - (\hat{\bf{R}} \bf{x} + \hat{\bf{t}}) \| \; ,
\label{eq:adi}
\end{equation}
where $(\bf{R}, \bf{t})$ are the ground-truth  rotation and translation, $(\hat{\bf{R}}, \hat{\bf{t}})$ the predicted ones, and $\mathcal{M}$ the
vertex set of the 3D model. We use this metric when evaluating the pose accuracy for the rotationally invariant objects, \emph{eggbox} and \emph{glue}
as in \cite{brachmann2016,hinterstoisser2012}.

\subsection{Single Object Pose Estimation}
\label{subsec:singleobj}

We first estimate the 6D pose of the central object in the RGB only \textsc{LineMod} images, without reference to the depth ones. We  compare our approach to those of~\cite{brachmann2016,kehl2017,rad2017}, which operate under similar conditions.

In this dataset, the training images are selected such that the relative orientation between corresponding pose
annotations are larger than a  threshold. As in~\cite{brachmann2016,kehl2017,rad2017}, to avoid being influenced by 
the scene context and overfitting to the background, we segment the training images using 
the segmentation masks provided with the dataset and replace the background by a random image from the PASCAL
VOC dataset~\cite{everingham10}. 

We use exactly the same training/test splits as in~\cite{rad2017}. We report our results in terms of
2D reprojection error in Table~\ref{tab:projerror}, 6D pose error in Table~\ref{tab:vertexerror} and IoU metric in Table~\ref{tab:iouerror}.
We provide example pose predictions of our approach in Figure~\ref{fig:linemodqual}.

\vspace{-3mm}

\subsubsection{Comparative Accuracy}

\paragraph{6D Accuracy in terms of projection error.}

In Table~\ref{tab:projerror}, we compare our results to  those of Brachmann et al.~\cite{brachmann2016}
and to BB8~\cite{rad2017}. Both of these competing methods involve a multi-stage pipeline
that comprises a 2D detection step followed by pose prediction and refinement. Since we do not have a refinement
stage, we show in the table their results without and with it. In both cases, we achieve better
6D pose estimation accuracies.

In Table~\ref{tab:iouerror}, we perform a similar comparison with SSD-6D~\cite{kehl2017}, whose
authors report their projection accuracy in terms of the IoU metric. That method also requires
{\it a posteriori} refinement and our results are again better in both cases, even though SSD-6D
 relies on a large training set of rendered images that are sampled over a wide range
 of viewpoints and locations.

\paragraph{6D Accuracy in terms of the ADD metric.}

%In Tables~\ref{tab:vertexerror} and~\ref{tab:ssd6dcomparison}, we compare our methods against
%the other in terms of the average of the 3D distances, as described in Section~\ref{subsec:eval}.
%Again we give numbers before and after refinement for the competing methods. Before refinement,
%we outperform all the methods by a significant margin of at least $12\%$. After refinement, our pose
%estimates are still better than Brachmann et al.~\cite{brachmann2016} but less accurate than
%BB8~\footnote{The authors do not report results without refinement, however they provided us with the accuracy numbers reported in Table~\ref{tab:vertexerror}.} and SSD-6D~\footnote{The authors were not able
%to provide their accuracy numbers without refinement for this metric, but made their code publicly available.
%We ran their code with provided pretrained models to obtain the 6D pose errors.}. Of course, we could
%perform a refinement stage as they do but this comes at a heavy computational penalty as we describe below.

In Tables~\ref{tab:vertexerror} and~\ref{tab:ssd6dcomparison}, we compare our methods against
the other in terms of the average of the 3D distances, as described in Section~\ref{subsec:eval}.
In Table~\ref{tab:vertexerror}, we give numbers before and after refinement for the competing methods. Before refinement,
we outperform all the methods by a significant margin of at least $12\%$. After refinement, our pose
estimates are still better than Brachmann et al.~\cite{brachmann2016}. By assuming the additional knowledge of a full 3D CAD model
and using it to further refine the pose, BB8~\footnote{The authors do not report results without refinement, 
	however they provided us with the accuracy numbers reported in Table~\ref{tab:vertexerror}.} and SSD-6D~\footnote{The authors were not able
	to provide their accuracy numbers without refinement for this metric, but made their code publicly available.
	We ran their code with provided pretrained models to obtain the 6D pose errors.} boost their pose estimation accuracy. 

Without any bells and whistles, our approach achieves state-of-the-art pose estimation accuracy in all the metrics without refinement.
When compared against methods that rely on the additional knowledge of full 3D CAD models and pose refinement, it still achieves state-of-the-art
performance in 2D projection error and IoU metrics and yields comparable accuracy in the ADD metric. Our approach could be used in conjunction 
with such refinement strategies to further increase the accuracy however this comes at a heavy computational cost
as we describe below.

\begin{table}[t]
	\centering
	\tabcolsep=0.07cm
	\scalebox{0.75}{
		\begin{tabular}{|c|cccc|ccc|}
			\hline
			\textbf{Method} & \multicolumn{4}{c|}{w/o Refinement}   							            							    & \multicolumn{3}{c|}{w/ Refinement} 			 	\\ \hline
			& Brachmann				     	 		 & BB8 			   		& SSD-6D	        &  OURS						& Brachmann       	   & BB8 				 & SSD-6D  			\\
			\textbf{Object} & \cite{brachmann2016}  			  	 & \cite{rad2017} 	  	& \cite{kehl2017}	&  			 				& \cite{brachmann2016} & \cite{rad2017} 	 & \cite{kehl2017}	       		\\ \hline
			Ape		    	& -                   		 	 		 & \textbf{\emph{27.9}}	& 0    			    &  21.62 					& 33.2      	 	   & 40.4                &\textbf{65}					   \\
			Benchvise    	& -          			 		  	     & 62.0            	    & 0.18 				&  \textbf{\emph{81.80}} 	& 64.8                 & \textbf{91.8}       &80              \\
			Cam    			& -	                	 		  		 & \textbf{\emph{40.1}} & 0.41 				&  36.57 					& 38.4                 & 55.7                &\textbf{78}               \\
			Can    			& -                   	 				 & 48.1                 & 1.35 				&  \textbf{\emph{68.80}} 	& 62.9                 & 64.1                &\textbf{86}               	\\
			Cat    			& -	                     		  		 & \textbf{\emph{45.2}} & 0.51 				&  41.82 					& 42.7                 & 62.6                &\textbf{70}            \\
			Driller         & -                   	 		 		 & 58.6                 & 2.58 				&  \textbf{\emph{63.51}} 	& 61.9                 & \textbf{74.4}       &73               \\
			Duck        	& -		                     		     & \textbf{\emph{32.8}} & 0    				&  27.23 					& 30.2                 & 44.3                &\textbf{66}          \\
			Eggbox        	& -		                 		  		 & 40.0                 & 8.9  				&  \textbf{\emph{69.58}} 	& 49.9                 & 57.8                &\textbf{100}         \\
			Glue        	& -                   		 		     & 27.0                 & 0    				&  \textbf{\emph{80.02}} 	& 31.2                 & 41.2                &\textbf{100}         \\
			Holepuncher     & -                  	 		  		 & 42.4                 & 0.30				&  \textbf{\emph{42.63}} 	& 52.8                 & \textbf{67.2}       &49                   \\ 		
			Iron        	& -                   	 		 		 & 67.0                 & 8.86 				&  \textbf{\emph{74.97}}  	& 80.0                 & \textbf{84.7}       &78                    \\			
			Lamp        	& -                       		 		 & 39.9                 & 8.20 				&  \textbf{\emph{71.11}}  	& 67.0                 & \textbf{76.5}       &73                   \\ 			
			Phone        	& -                   	 		  		 & 35.2                 & 0.18 				&  \textbf{\emph{47.74}} 	& 38.1                 & 54.0                &\textbf{79}                      \\ \hline
			Average         & 32.3                       		     & 43.6                 & 2.42 				&  \textbf{\emph{55.95}} 	& 50.2                 & 62.7                &\textbf{79}               \\ \hline					
		\end{tabular}
	}
	\vspace{-0.2mm}
	\caption{Comparison of our approach with state-of-the-art algorithms on \textsc{LineMod} in terms of ADD metric. We report percentages of
		correctly estimated poses.}
	\vspace{-2mm}
	\label{tab:vertexerror}
\end{table}

\begin{table}[t]
	\begin{center}
		\scalebox{0.7}{
			\begin{tabular}{|c|cc|cc|cc|}
				\hline
				\textbf{Threshold}  & \multicolumn{2}{c|}{10\%}   							            									& \multicolumn{2}{c|}{30\%} 			& \multicolumn{2}{c}{50\%} \\ \hline
				\textbf{Object} & \cite{kehl2017}  			  	 			 & OURS 	  					 & \cite{kehl2017} 			 & OURS 		 			    & \cite{kehl2017}          & OURS    \\ \hline
				Ape		    	& 0                   		 	 			 & \textbf{21.62}                        &   5.62   	 			 &    \textbf{70.67}     				& 19.95       			   &   \textbf{88.10}    \\
				Benchvise    	& 0.18          			 		  	     & \textbf{81.80}            	 		 &   2.07                    &    \textbf{91.07}                     & 10.62                    &   \textbf{98.85}      \\
				Cam    			& 0.41	                	 		  		 & \textbf{36.57}                        &   34.52               	 &    \textbf{81.57}                     & 63.54                    &   \textbf{94.80}        \\
				Can    			& 1.35                   	 				 & \textbf{68.80}                        &   61.43              	 &    \textbf{99.02}                     & 85.49               	   &   \textbf{99.90} \\
				Cat    			& 0.51	                     		  		 & \textbf{41.82}                        &   36.87               	 &    \textbf{90.62}                     & 64.04                    &   \textbf{98.80} \\
				Driller         & 2.58                   	 		 		 & \textbf{63.51}	                     &   56.01                   &    \textbf{99.01}                     & 84.86                    &   \textbf{99.80} \\
				Duck        	& 0		                     		  		 & \textbf{27.23}                        &   5.56               	 &    \textbf{70.70}                     & 32.65                    &    \textbf{89.39}  \\
				Eggbox        	& 8.9		                 		  		 & \textbf{69.58}                        &   24.61              	 &    \textbf{81.31}                     & 48.41                    &    \textbf{98.31}  \\
				Glue        	& 0                   		 		  		 & \textbf{80.02}                        &   14.18               	 &    \textbf{89.00}                     & 26.94               	   &    \textbf{97.20}  \\
				Holepuncher     & 0.30                  	 		  		 & \textbf{42.63}                        &   18.23               	 &    \textbf{85.54}                     & 38.75                    &    \textbf{96.29}  \\ 		
				Iron        	& 8.86                   	 		 		 & \textbf{74.97}                        &   59.26               	 &    \textbf{98.88}                     & 88.31                	   &   \textbf{99.39}  \\			
				Lamp        	& 8.20                       		 		 & \textbf{71.11}                        &   57.64                   &    \textbf{98.85}                     & 81.03                    &    \textbf{99.62}  \\ 						
				Phone        	& 0.18                   	 		  		 & \textbf{47.74}                        &   35.55                   &    \textbf{91.07}                     & 61.22                    &    \textbf{98.85}  \\ \hline
				Average         & 2.42                       		  		 & \textbf{55.95}                        &   31.65                   &    \textbf{88.25}                     & 54.29                    &	\textbf{96.78} \\ \hline					
			\end{tabular}
		}
	\end{center}
	\vspace{-0.1cm}
	\caption{Comparison of our approach with SSD-6D~\cite{kehl2017} without refinement using different thresholds for the 6D pose metric.}
	\label{tab:ssd6dcomparison}
	\vspace{-0.3cm}
\end{table}

\subsubsection{Accuracy / Speed Trade-off}
\vspace{-2mm}

In Table~\ref{tab:runtime}, we report the computational efficiency of our approach for single object pose estimation in comparison to
the state-of-the-art approaches~\cite{brachmann2016,kehl2017,rad2017}. Our approach runs at real-time performance in contrast to the existing approaches which
fall short of it. In particular, our algorithm runs at least $5$ times faster than the state-of-the-art techniques for single object pose estimation.

As can be seen in Table~\ref{tab:vertexerror}, pose refinement in Brachmann et al. increase the accuracy
significantly by $17.9\%$ at an additional run-time of $100$ miliseconds per object. BB8 also gets a substantial improvement of $19.1\%$ in
accuracy at an additional run-time of $21$ miliseconds per object. Even without correcting for the pose error, our approach
outperforms Brachmann et al. and yields close accuracy to BB8 while being $16$ times faster for single object
pose estimation. As discussed also in~\cite{kehl2017}, the unrefined poses computed
from the bounding boxes of the SSD 2D object detector, are rather approximate. We confirmed this by
running their publicly available code with the provided pretrained models. We report the accuracy numbers without the refinement
using the ADD metric in Table~\ref{tab:ssd6dcomparison} for different thresholds. While providing a good
initialization for the subsequent pose processing, the pose estimates of SSD-6D without refinement are much less accurate than
our approach. The further refinement increases the pose estimation accuracy significantly, however at the cost of a computational
time of 24 miliseconds per object. Moreover, in contrast to our approach, the refinement requires the knowledge of the full 3D object CAD model.

\begin{figure*}[t]
	\centering
	\scalebox{0.3}{
		\begin{tabular}{ccccc}
			\includegraphics[width=0.4\linewidth]{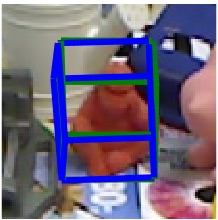}
			&\includegraphics[width=0.4\linewidth]{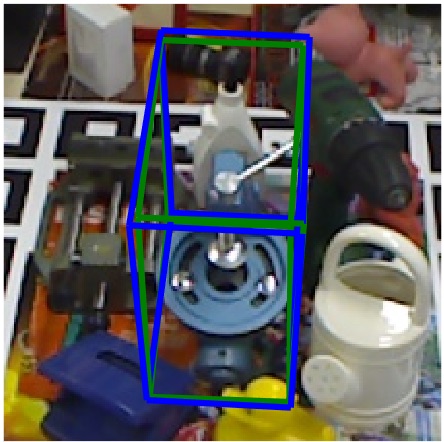}
			&\includegraphics[width=0.4\linewidth]{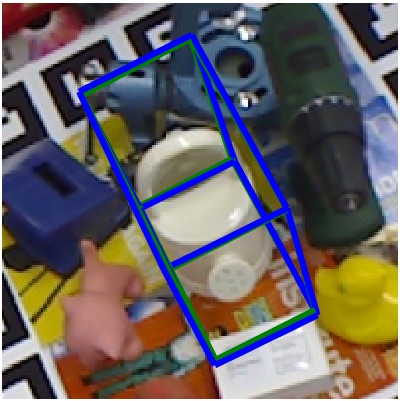}
			&\includegraphics[width=0.4\linewidth]{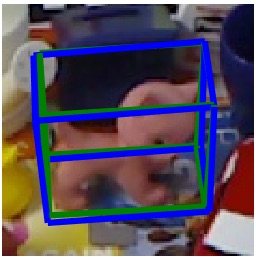}
			&\includegraphics[width=0.4\linewidth]{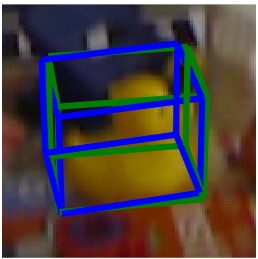} \\
			\includegraphics[width=0.4\linewidth]{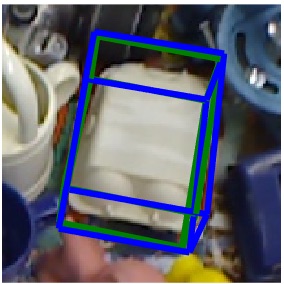}
			&\includegraphics[width=0.4\linewidth]{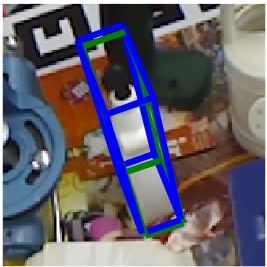}
			&\includegraphics[width=0.4\linewidth]{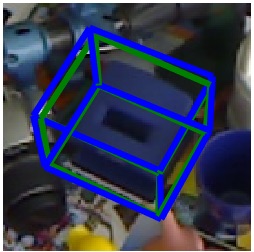}
			&\includegraphics[width=0.4\linewidth]{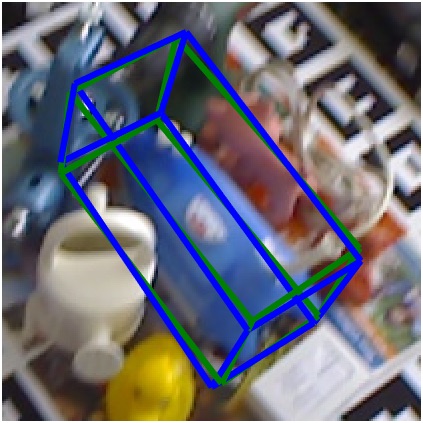}
			&\includegraphics[width=0.4\linewidth]{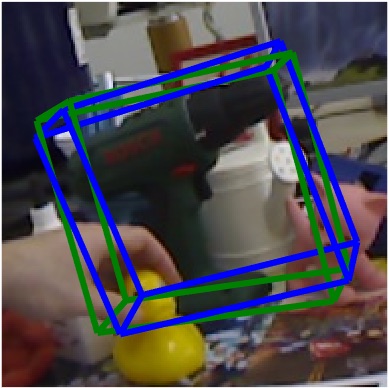} \\
			\includegraphics[width=0.4\linewidth]{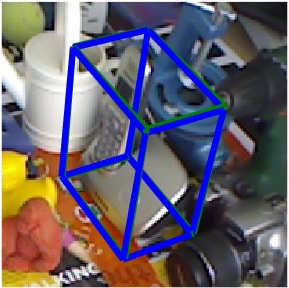}
			&\includegraphics[width=0.4\linewidth]{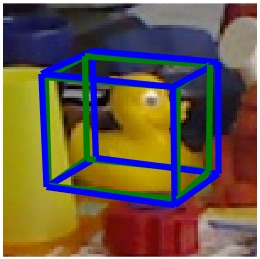}
			&\includegraphics[width=0.4\linewidth]{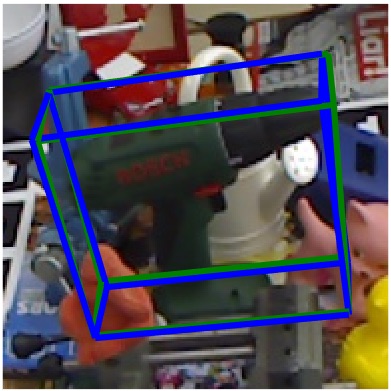}
			&\includegraphics[width=0.4\linewidth]{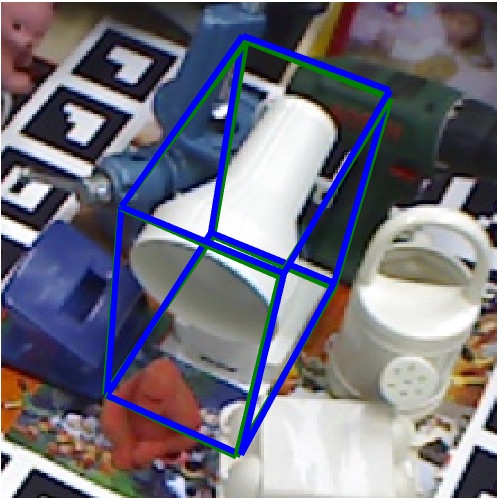}
			&\includegraphics[width=0.4\linewidth]{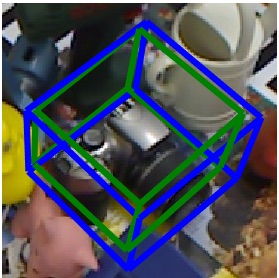} \\
%			&\includegraphics[width=0.4\linewidth]{failurecase_cat1} \\
		\end{tabular}
	} \vspace{-0mm}
	\caption{Pose  estimation  results  of our approach. Note that our method can  recover
		the  6D  pose  in these  challenging  scenarios,  which involve significant amounts of clutter,
		occlusion and orientation ambiguity. In the last column, we show failure cases due to motion blur,
		severe occlusion and  specularity (this figure is best viewed on a computer screen).}
	\vspace{-4mm}
	\label{fig:linemodqual}
\end{figure*}

In Figure~\ref{fig:linemodqual}, we show example results of our method on the
\textsc{LineMod}. We include more visual results of our method in the supplementary material.

\begin{table}[t]
	\begin{center}
		\scalebox{0.65}{
			\begin{tabular}{|c|cc|c|}
				\hline
				\textbf{Method} & \multicolumn{2}{c|}{w/o Refinement}   							            			      & \multicolumn{1}{c|}{w/ Refinement} 			 \\ \hline
				& SSD-6D				     	 		  			   			       &  OURS				      & SSD-6D      				    			\\
				\textbf{Object} & \cite{kehl2017}  			  	  	  		   				   		   &  			 		  	  & \cite{kehl2017} 	       \\ \hline
				Ape		    	& 98.46                   		 	 		 						   & \textbf{99.81} 		  & 99      					   \\
				Benchvise    	& \textbf{100}            			 		  	     				   & 99.90 					  & \textbf{100}                   \\
				Cam    			& 99.53	                	 		  		 						   & \textbf{100}   		  & 99 	                   \\
				Can    			& \textbf{100}                     	 				 				   & 99.81	  				  & \textbf{100}                      	\\
				Cat    			& 99.34	                     		  		 						   & \textbf{99.90} 		  & 99 	                \\
				Duck        	& 99.04		                     		     						   & \textbf{100}   		  & 98 	              \\
				Glue        	& 97.24                   		 		     						   & \textbf{99.81} 		  & 98	             \\
				Holepuncher     & 98.95                  	 		  		 						   & \textbf{99.90} 		  & 99 	                \\ 		
				Iron        	& 99.65                   	 		 		 						   & \textbf{100}   		  & 99 	                        \\			
				Lamp        	& 99.38                       		 		 						   & \textbf{100}   		  & 99 	                       \\ 						
				Phone        	& 99.91                   	 		  		 						   & \textbf{100}   		  & 100                           \\ \hline
				Average         & 99.22                       		     						       & \textbf{99.92} 		  & 99.4                   \\ \hline		
				Driller         & -                   	 		 		 						       & \textbf{100}   		  & 99 	                     \\
				Eggbox        	& -		                 		  		 						       & \textbf{99.91} 		  & 99 	                   \\ \hline
				
			\end{tabular}
		}
	\end{center}
	\vspace{-1mm}
	\caption{Comparison of our approach against~\cite{kehl2017} on \textsc{LineMod} using IoU metric. The authors of~\cite{kehl2017} were able to provide us the results of our approach w/o the refinement. }
	\vspace{-3mm}
	\label{tab:iouerror}
\end{table}

\begin{table}[t]
	\centering
	\tabcolsep=0.3cm
	\scalebox{0.85}{
		\begin{tabular}[b]{lcc}
			\toprule
			Method 									& Overall Speed    & Refinement runtime     \\
			\midrule
			Brachmann et al.~\cite{brachmann2016} 	& ~2  fps          & 100 ms/object        	\\
			Rad \& Lepetit~\cite{rad2017}			& ~3  fps  		   & 21  ms/object    		\\
			Kehl et al.~\cite{kehl2017}				& ~10 fps  		   & 24  ms/object   		\\
			OURS  					                & ~50 fps  		   & -    			     	\\
			\bottomrule
		\end{tabular}
	}
	\caption{Comparison of the overall computational runtime of our approach in comparison to~\cite{brachmann2016,kehl2017,rad2017}. We further provide
		the computational runtime induced by the pose refinement stage of~\cite{brachmann2016,kehl2017,rad2017}}
	\vspace{-1mm}
	\label{tab:runtime}
\end{table}

\subsection{Multiple Object Pose Estimation}

\begin{figure}[t]
	\centering
	\scalebox{0.6}{
		\begin{tabular}{c}
			\includegraphics[width=\linewidth]{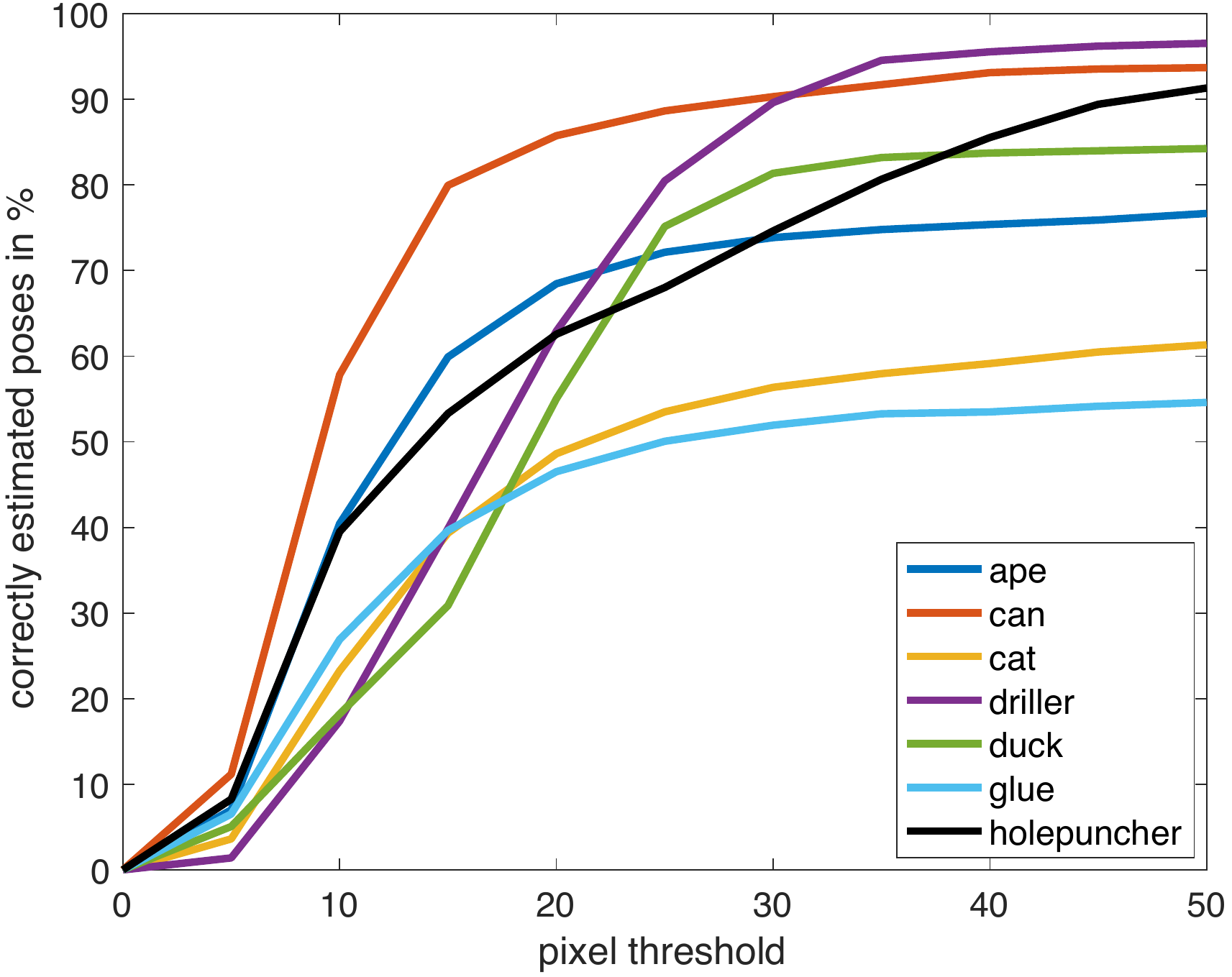}
		\end{tabular}
	}
	\vspace{-1mm}
	\caption{Percentage of correctly estimated poses as a function of the projection error for different objects of the Occlusion dataset~\cite{brachmann2016}.}
	\vspace{-3mm}
	\label{fig:occlusionpose}
\end{figure}

We use the OCCLUSION dataset to compare our approach to Brachmann et al.~\cite{brachmann2016} for multi-object detection
and report pose estimation accuracy as in~\cite{rad2017}. The identity of the objects cannot be assumed to be
known {\it a priori} and has to be guessed. To this end, the method of~\cite{rad2017} assumes that it has access to image crops
based on the ground-truth 2D bounding boxes~\footnote{This it is not explicitly stated in \cite{rad2017}, but the authors
	confirmed this to us in private email communication.}. We make no such assumptions. Instead, we jointly
detect the object in 2D, estimate its identity and predict its 6D pose. We generate our training images with the approach explained in Section~\ref{subsec:eval}.
We further augment the \textsc{LineMod} training data by adding into  the images objects extracted from other training sequences. We report our pose estimation accuracy in Figure~\ref{fig:occlusionpose} and demonstrate
that even without assuming ground-truth information as in the case of~\cite{rad2017}, our method yields satisfactory pose accuracy in the case of severe occlusions.
For object detection purposes, we consider an estimate to be correct if its detection IoU is larger than $0.5$. Note that here the
detection IoU corresponds to the overlap of the 2D bounding boxes of the object, rather than the overlap of
the projected masks as is the case for the IoU metric defined in Sec~\ref{subsec:eval}. In Table~\ref{tab:occlusiondetection}, we report
a mean average precision (MAP) of $0.48$ which is similar to the accuracy reported by~\cite{brachmann2016} and outperforms
the ones reported by~\cite{hinterstoisser2011,kehl2017}.
\vspace{-3mm}

\begin{table}[ht!]
	\begin{center}
		\begin{tabular}{cc}
			\parbox[c]{5cm}{
				\begin{small}
					\begin{tabular}[b]{lc}
						\toprule
						Method 		                			        & MAP   \\
						\midrule
						Hinterstoisser et al.~\cite{hinterstoisser2011} & 0.21    \\
						Brachmann et al.~\cite{brachmann2016}    	    & 0.51        \\
						Kehl et al.~\cite{kehl2017}					    & 0.38        \\
						OURS										    & 0.48        \\
						\bottomrule
					\end{tabular}
				\end{small}
			}& \hspace{-1cm}     			
			\parbox[c]{4.2cm}{
				\includegraphics[width=0.45\columnwidth]{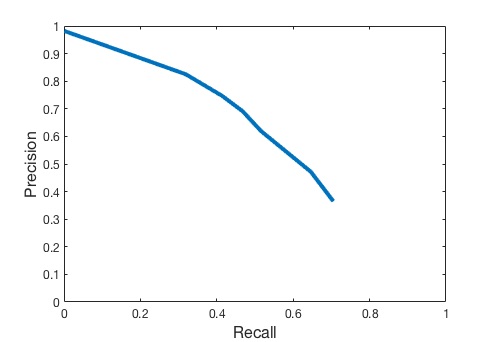}}\\
			%			(a)&(b)
		\end{tabular}
	\end{center}
	\vspace{-2mm}
	\caption{The detection experiment on the Occlusion dataset~\cite{brachmann2016}. (Left) Precision-recall plot. (Right)}
	\vspace{-4mm}
	\label{tab:occlusiondetection}
\end{table}

\begin{figure}[t]
	\centering
	\scalebox{0.5}{
		\begin{tabular}{c}
			\includegraphics[width=\columnwidth]{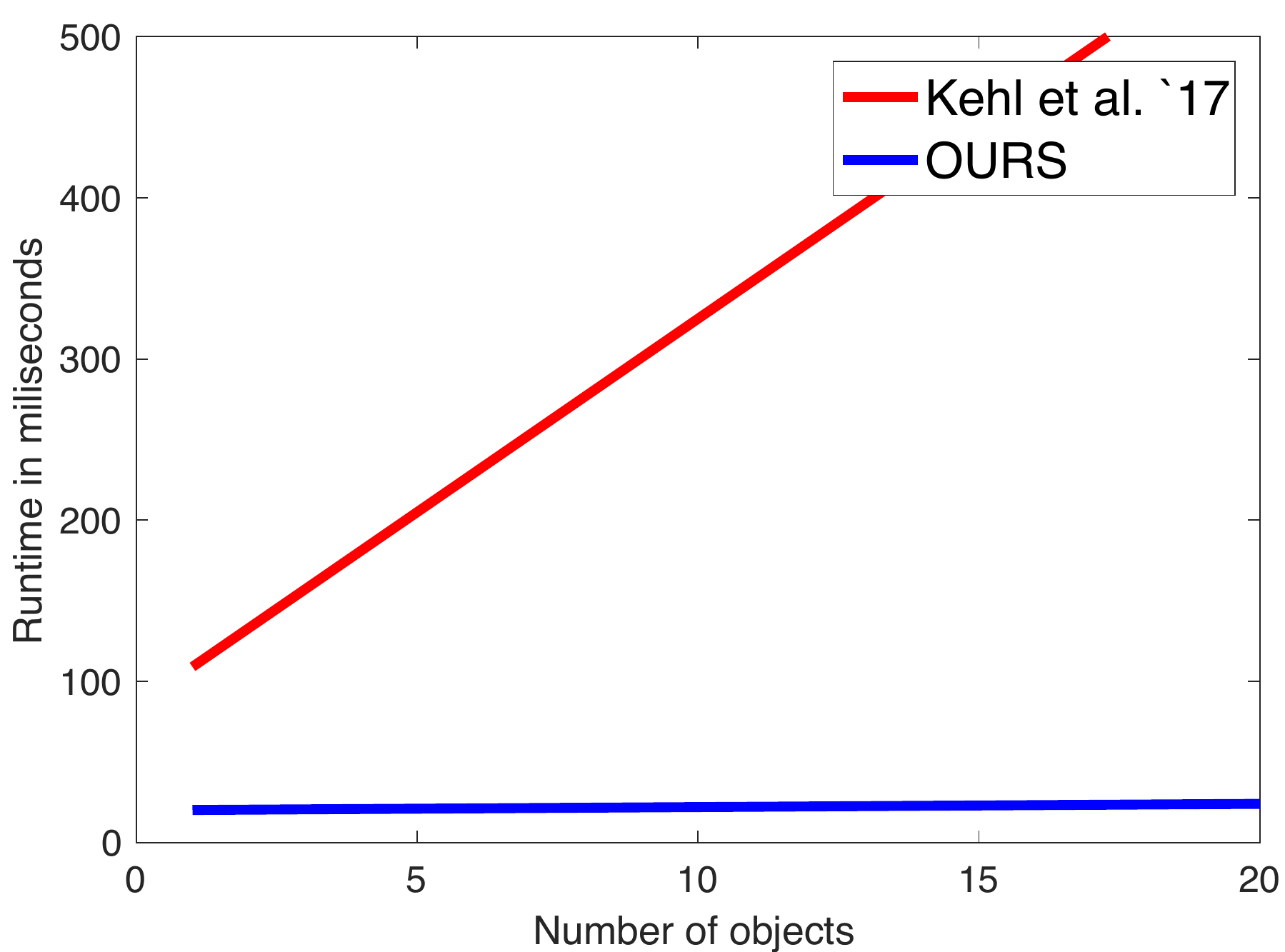}
		\end{tabular}
	}
	\vspace{-1mm}
	\caption{The runtime of our approach with increasing number of objects as compared to that of~\cite{kehl2017}.}
	\vspace{-3mm}
	\label{fig:runtimeperobject}
\end{figure}

Our approach provides accurate 6D poses with real-time performance. Upon one network invocation, our
only computational overhead is an efficient PnP algorithm which operates on just $9$ points per object.
Furthermore we do not require full 3D colored object models to further refine our initial pose estimates.
Our approach is therefore scalable to handle multiple objects as shown in Figure~\ref{fig:runtimeperobject} and
has only a negligible computational overhead of  PnP (0.2 miliseconds/object) while the competing approaches~\cite{kehl2017}
have a linear runtime growth.

We also evaluated the accuracy and speed of our approach for different input resolutions.
As explained in Section~\ref{subsec:model}, we adopt a multi-scale training procedure and change the
input resolution during training randomly as in~\cite{redmon2016b}. This allows us to
be able to change the input resolution at test-time and predict from images with
higher resolution. This is especially useful for predicting the pose of small objects
more robustly. As we do not have an initial step for 2D object detection and produce image crops
which are then resized to higher resolutions for pose prediction as in~\cite{rad2017}, our approach
requires better handling of the small objects. In Table~\ref{tab:resolutions}, we compare the
accuracy and computational efficiency of our approach for different input resolutions. With only
$1\%$ decrease in accuracy the average runtime per image is 94 ms and the runtime virtually
remains the same for estimating the pose of multiple objects.

\begin{table}[tbph]
	\centering
	\tabcolsep=0.1cm
	\scalebox{0.8}{
		\begin{tabular}[b]{lcc}
			\toprule
			Resolution 		 	      & 2D projection metric & Speed \\
			\midrule
			416 $\times$ 416      & 89.71    		 & 94 fps   \\
			480 $\times$ 480 	  & 90.00      		 & 67 fps   \\
			544 $\times$ 544	  & 90.37  			 & 50 fps   \\
			688 $\times$ 688 	  & 90.65 	 		 & 43 fps 	  	\\
			\bottomrule
		\end{tabular}
	}
	\vspace{-1mm}
	\caption{Accuracy/speed trade-off of our method on the \textsc{LineMod} dataset. Accuracy reported is the
		percentage of correctly estimated poses w.r.t the 2D projection error. The same network model
		is used for all four input resolutions. Timings are on a Titan X (Pascal) GPU.}
	\vspace{-4mm}
	\label{tab:resolutions}
\end{table}

%% -*- mode: latex; mode: reftex; mode: flyspell; mode: auto-fill; TeX-master: "top.tex"; -*-

\section{Conclusion}

We have proposed a new CNN architecture for fast and accurate single-shot 6D pose prediction that naturally extends the single shot 2D object detection paradigm to 6D object detection. 
Our network predicts 2D locations of the projections of the object’s 3D bounding box corners which involves predicting just a few more 2D points than for 2D bounding box regression. 
Given the predicted 2D corner projections, the 6D pose is computed via an efficient PnP method. For high accuracy, existing CNN-based 6D object detectors all refine their pose estimates 
during post-processing, a step that requires an accurate 3D object model and also incurs a runtime overhead per detected object. In contrast, our single shot predictions are very 
accurate which alleviates the need for refinement. Due to this, our method is not dependent on access to 3D object models and there is virtually no overhead when estimating the pose of multiple 
objects. Our method is real-time; it runs at 50 -- 94 fps depending on the image resolution. This makes it substantially faster than existing methods.

\vspace{-4mm}

\paragraph{Acknowledgements.} This work was supported in part by the Swiss National Science Foundation. We would like to thank Mahdi Rad, Vincent Lepetit, Wadim Kehl, Fabian Manhardt and Slobodan Ilic for helpful discussions.

{\fontsize{8}{9.6}\selectfont
	\bibliographystyle{ieee}
	\bibliography{ref}
}

%{\footnotesize
%\bibliographystyle{ieee}
%%\bibliography{short,vision,learning}
%\bibliography{ref}
%}

%% -*- mode: latex; mode: reftex; mode: flyspell; mode: auto-fill; TeX-master: "top.tex"; -*-

\makeatletter
\renewcommand\thesection{}
\renewcommand\thesubsection{\@arabic\c@section.\@arabic\c@subsection}
\makeatother	
\twocolumn[\vspace{20mm}\centering{\section*{Supplemental Material: \\ ``Real-Time Seamless Single Shot 6D Object Pose Prediction''}}\vspace{20mm}]

In the supplemental material, we provide details on how the training images were prepared and  on the proposed confidence function and the weighted prediction step. We also present qualitative results on \textsc{Occlusion}~\cite{brachmann2014} and \textsc{LineMod}~\cite{hinterstoisser2012}.

\vspace{-0.1mm}

\paragraph{Training Images.}

\begin{figure}[tbph]
	\centering
	\scalebox{0.7}{
		\begin{tabular}{cccc}
			\hspace{-4mm}
			\includegraphics[height=2.7cm,width=2.7cm]{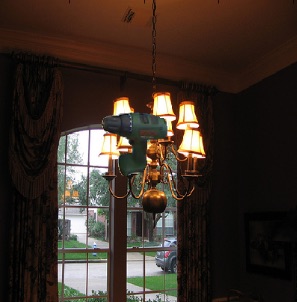}
			&\includegraphics[height=2.7cm,width=2.7cm]{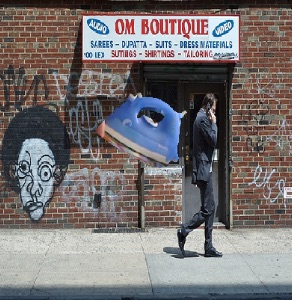}
			&\includegraphics[height=2.7cm,width=2.7cm]{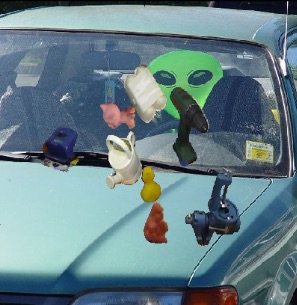}
			&\includegraphics[height=2.7cm,width=2.7cm]{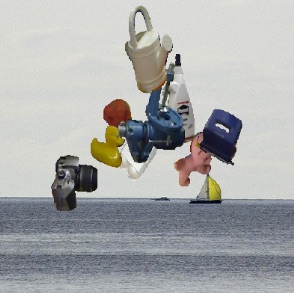}\\
		\end{tabular}
	}
	\caption{Using segmentation masks given in \textsc{LineMod}, we extract the foreground objects in our training images and composite
		them over random images from PASCAL VOC~\cite{everingham10}. We also augment the training set by combining images of multiple objects taken from different training images.}
	\vspace{-5mm}
	\label{fig:training_imgs}
\end{figure}

As discussed in the main paper, we segment the foreground object in the images of the training set,
using the segmentation masks provided and paste the segmented image over a random image as in~\cite{brachmann2016,kehl2017,rad2017}. Examples of such images, which are given as input
to the network at training time are shown in Figure~\ref{fig:training_imgs}. This operation of removing the actual background prevents the network from 
overfitting to the background, which is similar for training and test images of~\textsc{LineMod}. When we train a model without eliminating the
background, in practice, we observe about $1$\% improvement in the 2D projection score.

\vspace{-2mm}
\vspace{-4mm}

\begin{figure}[ht!]
	\begin{center}\scalebox{0.8}{
			\begin{tabular}{cc}
				\hspace{-1.8cm}
				\parbox[c]{5.2cm}{
					\includegraphics[width=0.54\columnwidth]{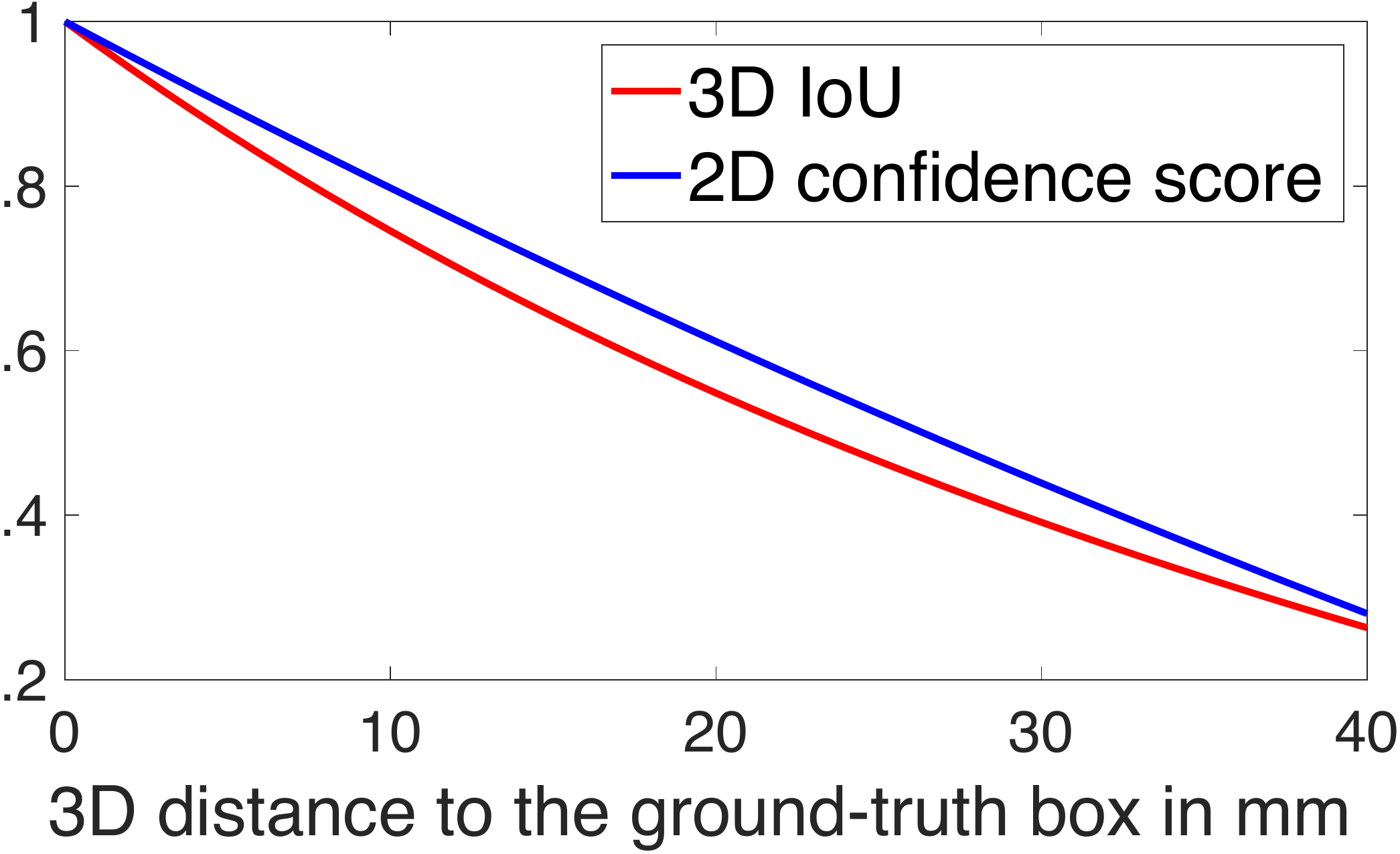}}
				&   \hspace{-1.1cm}		
				\parbox[c]{4cm}{
					\begin{small}
						\scalebox{0.95}{
							\begin{tabular}[b]{lc}
								\hline
								Method 		                & Runtime per object (ms)   \\
								\hline
								3D IoU					    & 5.37           \\
								2D Conf. Score				& 0.18           \\
								\hline
						\end{tabular} }
					\end{small}
				}\\
		\end{tabular}}
	\end{center}
	\vspace{-2mm}
	\caption{Comparison of the 3D IoU and our 2D confidence score in value (Left) and runtime (Right). The model for the~\emph{Cam} object is 
		shifted in x-dimension synthetically to produce a distorted prediction and projected on the image plane with randomly chosen 20 transformation matrices 
		from \textsc{LineMod}. Scores are computed between the ground-truth references and distorted predictions. Results are averaged over all the trials. The runtime 
		for 3D IoU is computed using the optimized PyGMO library that relies on \cite{Bringmann10}.}
	\vspace{-2.2mm}
	\label{fig:confvsiou}
\end{figure}

\paragraph{Confidence function.}
We analyze in Figure~\ref{fig:confvsiou} our confidence function in comparison to 3D cube IoU in terms of its value and runtime. We show that our confidence function closely approximates the actual 3D cube IoU while being much faster to compute.

\vspace{-0.2mm}

\vspace{-1mm}

\vspace{-2mm}
\begin{figure}[tbph]
	\centering
	\includegraphics[height=2.4cm,width=2.4cm]{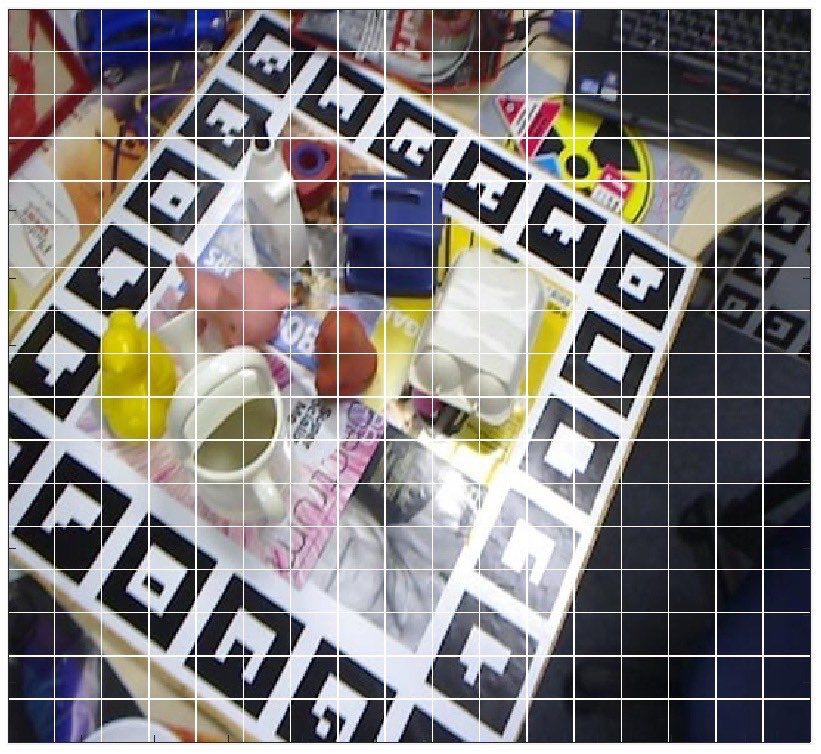}
	\hspace{1mm}
	\includegraphics[height=2.4cm,width=2.8cm]{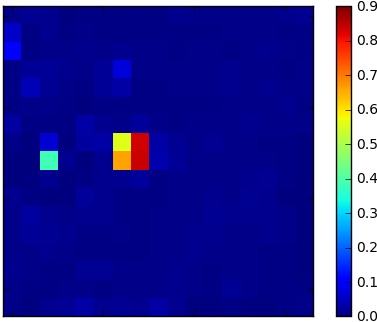}
	\hspace{1mm}
	\includegraphics[height=2.4cm,width=2.4cm]{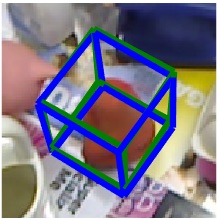}
	\caption{(Left) The 17$\times$17 grid on a 544$\times$544 image. (Middle) Confidence values for predictions of the \emph{ape} object on the grid. (Right) Cropped view of our pose estimate (shown in blue) and the ground truth (shown in green). Here, three cells next to the best cell have good predictions and their combination gives a more accurate pose than the best prediction alone (best viewed in color).}
	\vspace{-4mm}
	\label{fig:weightedconfidence}
\end{figure}

\paragraph{Confidence-weighted prediction.}
In the final step of our method, we compute a weighted sum of multiple sets of predictions for the corners and the centroid, using associated confidence values as weights.
On \textsc{LineMod}, this gave a 1--2\% improvement in accuracy with the 2D projection metric.
The first step involves scanning the full 17$\times$17 grid to find the cell with the highest confidence for each potential object. 
We then consider a $3 \times 3$ neighborhood around it on the grid and prune the cells with confidence values lower than the detection threshold of $0.5$.
On the remaining cells, we compute a confidence-weighted average of the associated predicted 18-dimensional vectors, where
the eight corner points and the centroid have been stacked to form the vector. The averaged coordinates are then used in the PnP method.
This {\emph sub-pixel refinement} on the grid usually improves the pose of somewhat large objects that occupy several adjoining cells in the grid.
Figure~\ref{fig:weightedconfidence} shows an example where the \emph{ape} object lies between two adjoining cells
and the confidence weighting improves the pose accuracy.

\vspace{-4mm}

\begin{figure*}[!htb]
	\centering
	\begin{tabular}[t]{ccc}
		%1
		\begin{subfigure} %{width=0.32\linewidth}
			\centering
			\smallskip
			\includegraphics[height=4cm,width=0.3\linewidth]{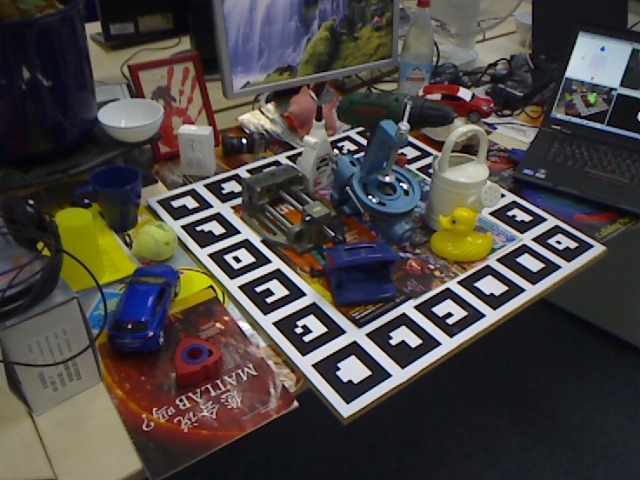}
		\end{subfigure}
		&
		\begin{subfigure} %{width=0.32\linewidth}
			\centering
			\smallskip
			\includegraphics[height=4cm,width=0.3\linewidth]{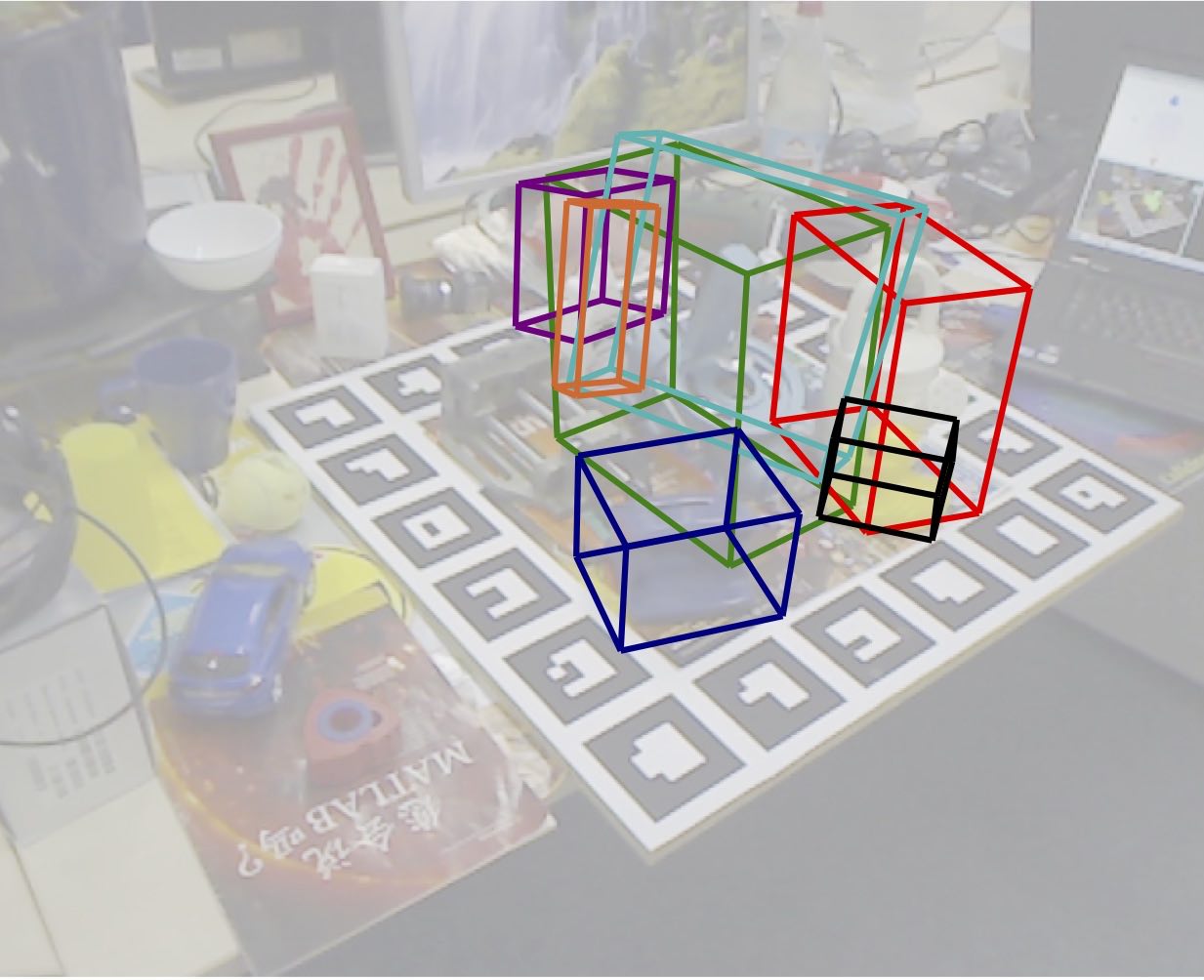}
		\end{subfigure}
		&
		\raisebox{1.8cm}{
			\begin{tabular}{ccc}% if you add [t], than sub images are pushed down
				\smallskip
				\begin{subfigure} %{width=0.1\linewidth}
					\centering
					\includegraphics[height=1.85cm,width=0.1\linewidth]{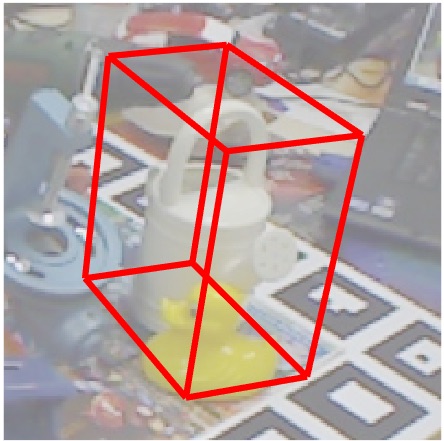}
				\end{subfigure} \hspace{-6mm}
				&\begin{subfigure} %{width=0.1\linewidth}
					\centering
					\includegraphics[height=1.85cm,width=0.1\linewidth]{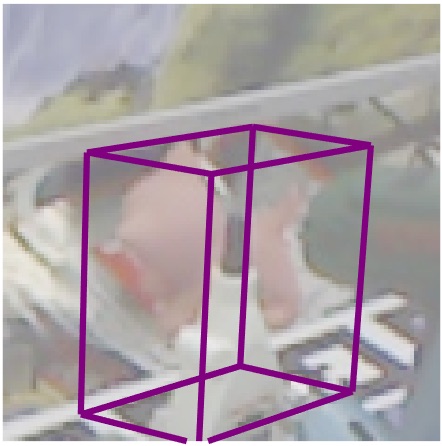}
				\end{subfigure} \hspace{-6mm}
				&\begin{subfigure}%{width=0.1\linewidth}
					\centering
					\includegraphics[height=1.85cm,width=0.1\linewidth]{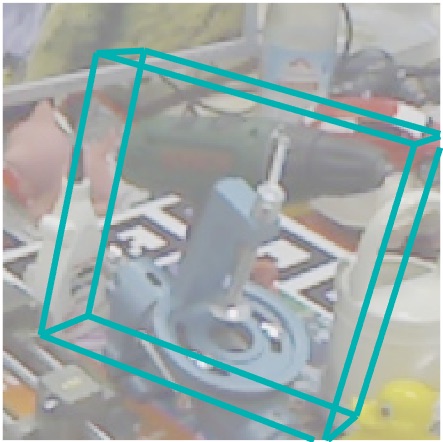}
				\end{subfigure}
				\\
				\begin{subfigure} %{width=0.1\linewidth}
					\centering
					\includegraphics[height=1.85cm,width=0.1\linewidth]{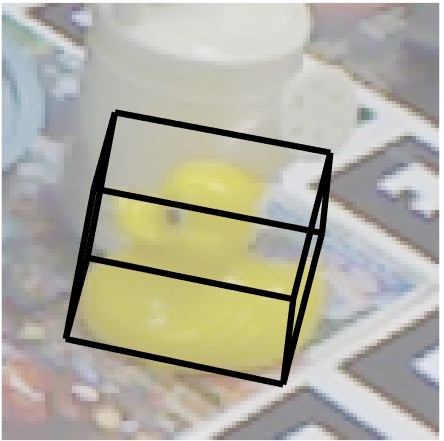}
				\end{subfigure} \hspace{-6mm}
				&\begin{subfigure} %{width=0.1\linewidth}
					\centering
					\includegraphics[height=1.85cm,width=0.1\linewidth]{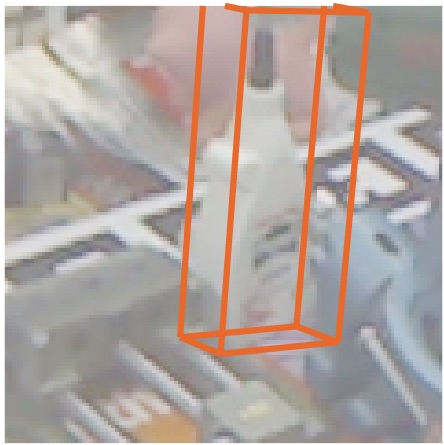}
				\end{subfigure} \hspace{-6mm}
				&\begin{subfigure} %{width=0.1\linewidth}
					\centering
					\includegraphics[height=1.85cm,width=0.1\linewidth]{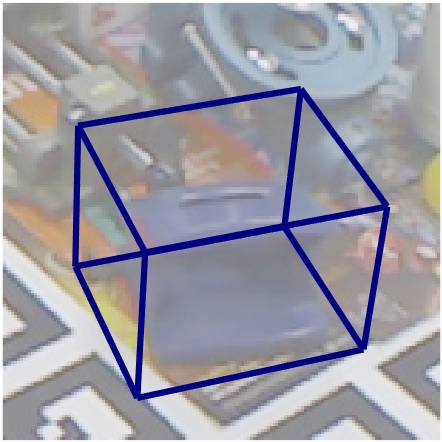}
				\end{subfigure}
		\end{tabular}}\\ %\vspace{2mm}
		%2	
		\begin{subfigure} %{width=0.32\linewidth}
			\centering
			\smallskip
			\includegraphics[height=4cm,width=0.3\linewidth]{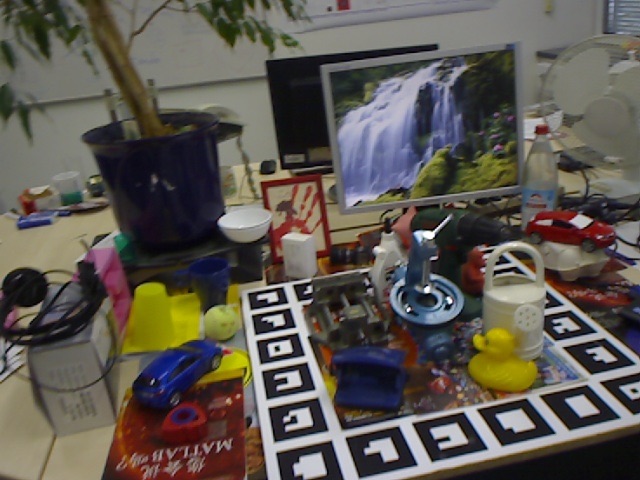}
		\end{subfigure}
		&
		\begin{subfigure} %{width=0.32\linewidth}
			\centering
			\smallskip
			\includegraphics[height=4cm,width=0.3\linewidth]{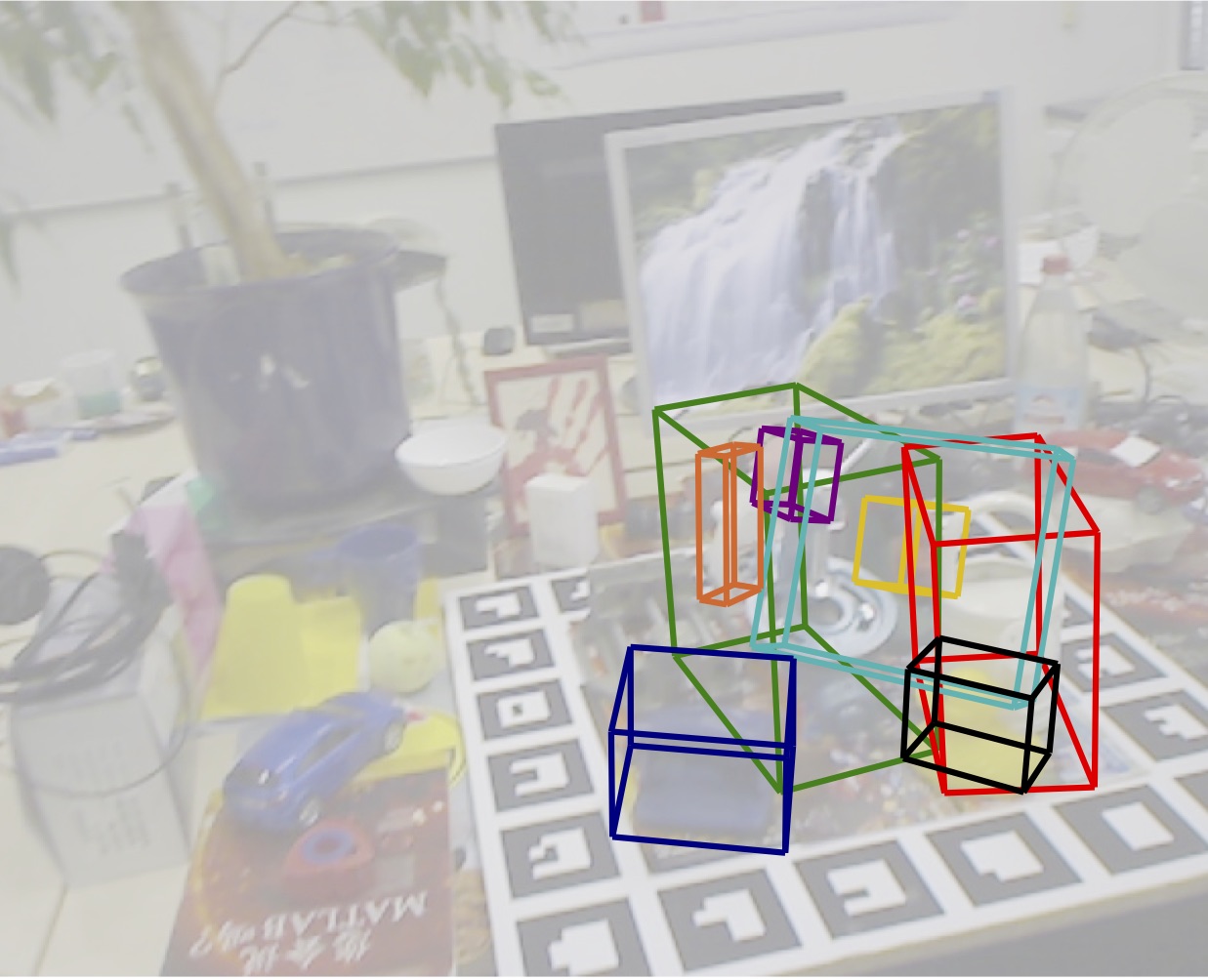}
		\end{subfigure}
		&
		\raisebox{1.8cm}{
			\begin{tabular}{ccc}% if you add [t], than sub images are pushed down
				\smallskip
				\begin{subfigure} %{width=0.1\linewidth}
					\centering
					\includegraphics[height=1.85cm,width=0.1\linewidth]{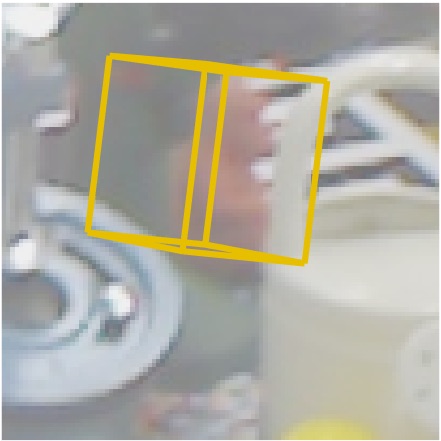}
				\end{subfigure} \hspace{-6mm}
				&\begin{subfigure} %{width=0.1\linewidth}
					\centering
					\includegraphics[height=1.85cm,width=0.1\linewidth]{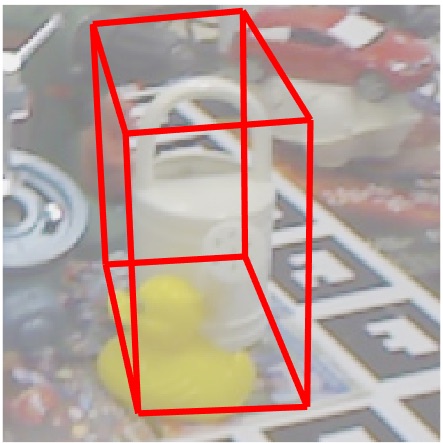}
				\end{subfigure} \hspace{-6mm}
				&\begin{subfigure}%{width=0.1\linewidth}
					\centering
					\includegraphics[height=1.85cm,width=0.1\linewidth]{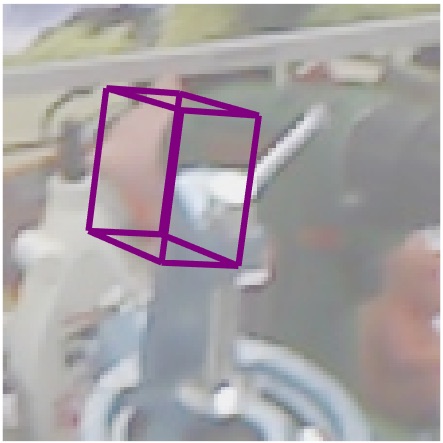}
				\end{subfigure}
				\\
				\begin{subfigure} %{width=0.1\linewidth}
					\centering
					\includegraphics[height=1.85cm,width=0.1\linewidth]{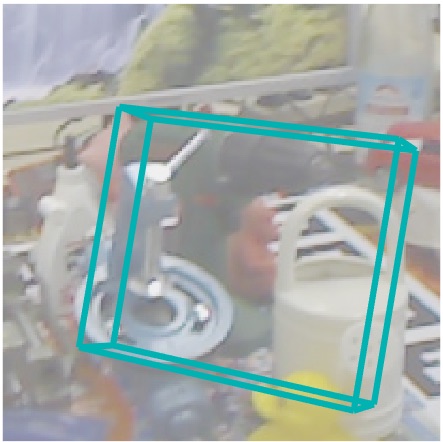}
				\end{subfigure} \hspace{-6mm}
				&\begin{subfigure} %{width=0.1\linewidth}
					\centering
					\includegraphics[height=1.85cm,width=0.1\linewidth]{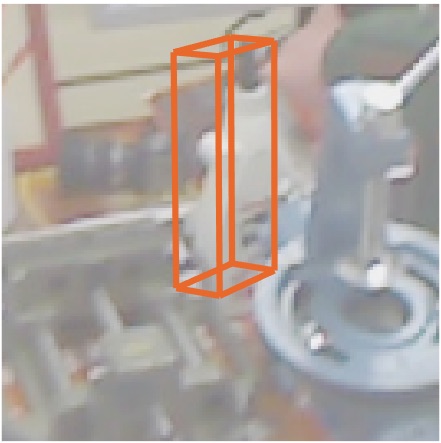}
				\end{subfigure} \hspace{-6mm}
				&\begin{subfigure} %{width=0.1\linewidth}
					\centering
					\includegraphics[height=1.85cm,width=0.1\linewidth]{occlusion48holepuncher}
				\end{subfigure}
		\end{tabular}}\\ %\vspace{2mm}
		%3
		\begin{subfigure} %{width=0.32\linewidth}
			\centering
			\smallskip
			\includegraphics[height=4cm,width=0.3\linewidth]{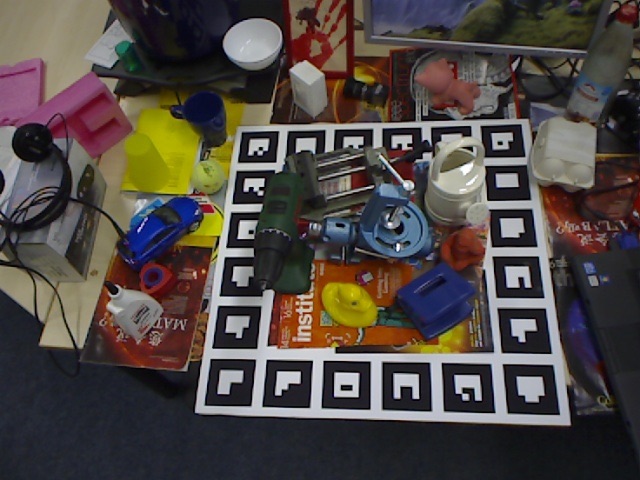}
		\end{subfigure}
		&
		\begin{subfigure} %{width=0.32\linewidth}
			\centering
			\smallskip
			\includegraphics[height=4cm,width=0.3\linewidth]{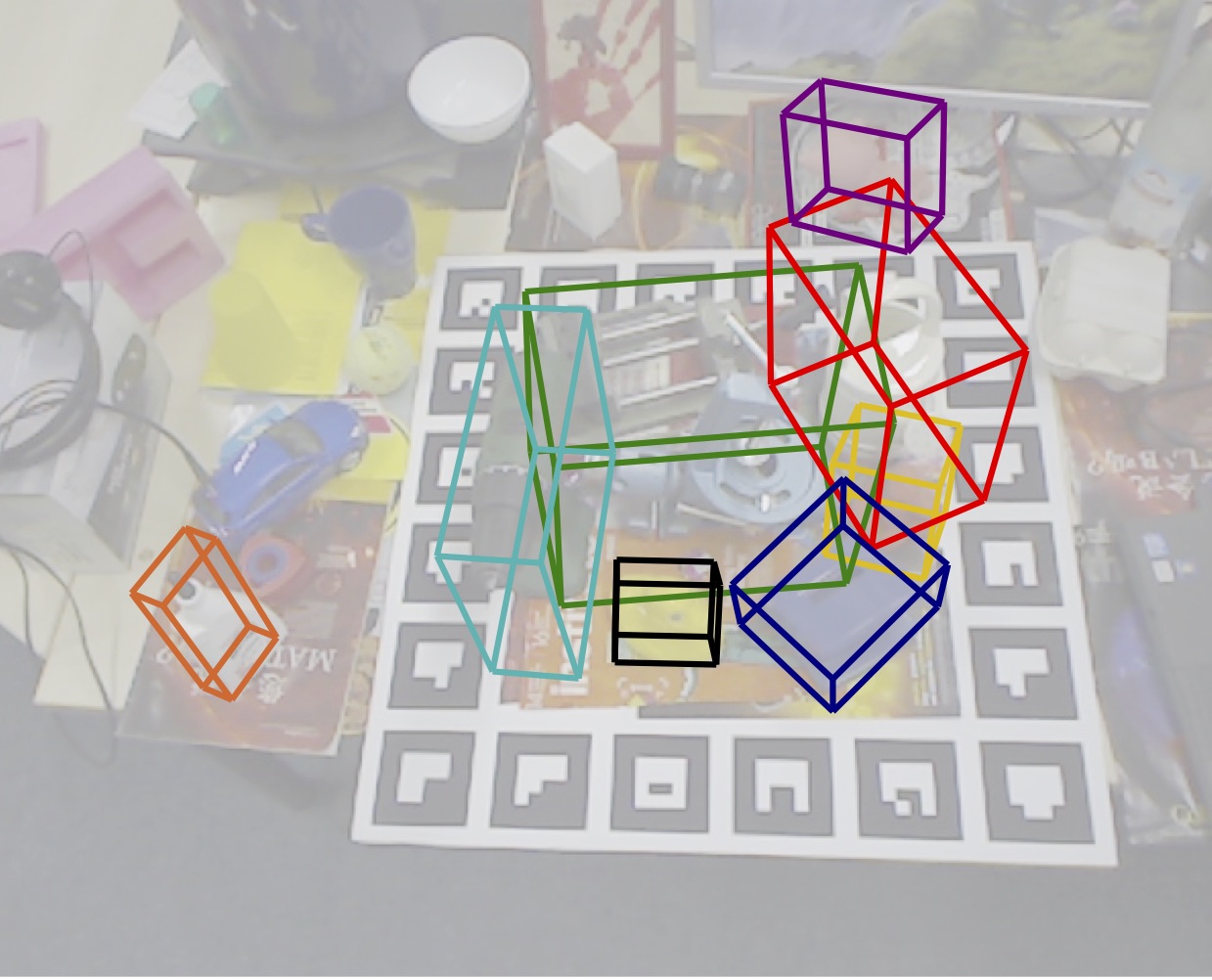}
		\end{subfigure}
		&
		\raisebox{1.8cm}{
			\begin{tabular}{ccc}% if you add [t], than sub images are pushed down
				\smallskip
				\begin{subfigure} %{width=0.1\linewidth}
					\centering
					\includegraphics[height=1.85cm,width=0.1\linewidth]{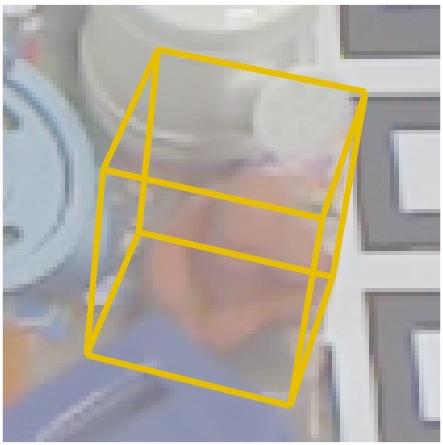}
				\end{subfigure} \hspace{-6mm}
				&\begin{subfigure} %{width=0.1\linewidth}
					\centering
					\includegraphics[height=1.85cm,width=0.1\linewidth]{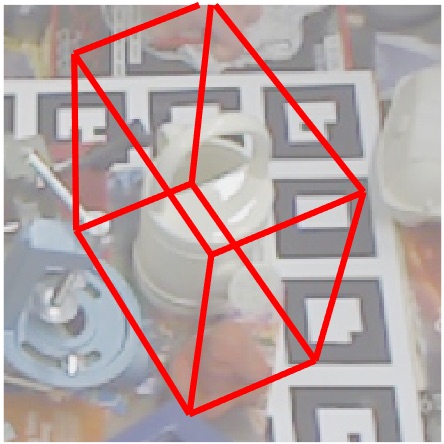}
				\end{subfigure} \hspace{-6mm}
				&\begin{subfigure}%{width=0.1\linewidth}
					\centering
					\includegraphics[height=1.85cm,width=0.1\linewidth]{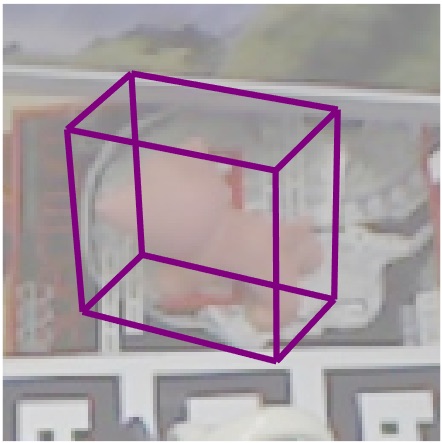}
				\end{subfigure}
				\\
				\begin{subfigure} %{width=0.1\linewidth}
					\centering
					\includegraphics[height=1.85cm,width=0.1\linewidth]{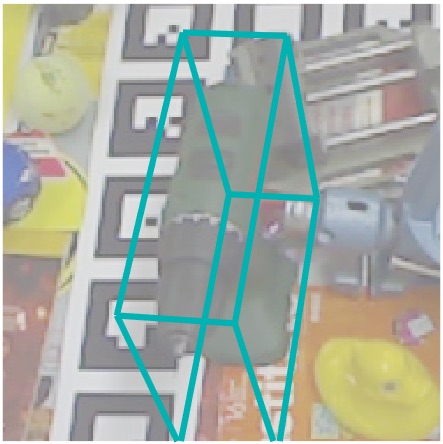}
				\end{subfigure} \hspace{-6mm}
				&\begin{subfigure} %{width=0.1\linewidth}
					\centering
					\includegraphics[height=1.85cm,width=0.1\linewidth]{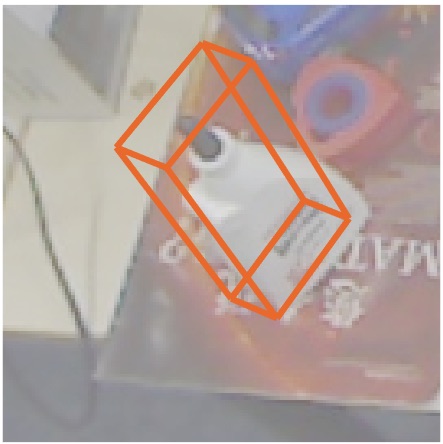}
				\end{subfigure} \hspace{-6mm}
				&\begin{subfigure} %{width=0.1\linewidth}
					\centering
					\includegraphics[height=1.85cm,width=0.1\linewidth]{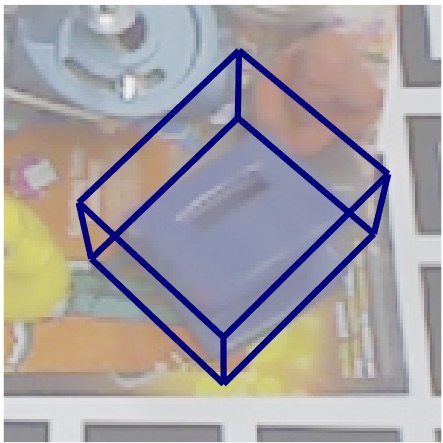}
				\end{subfigure}
		\end{tabular}}\\ %\vspace{2mm}
		%4
		\begin{subfigure} %{width=0.32\linewidth}
			\centering
			\smallskip
			\includegraphics[height=4cm,width=0.3\linewidth]{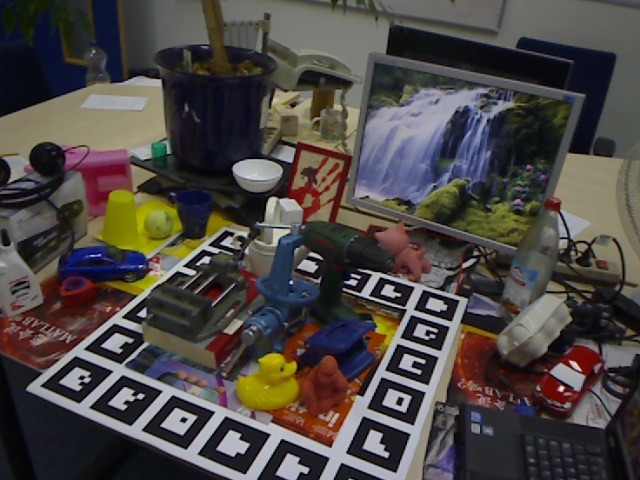}
		\end{subfigure}
		&
		\begin{subfigure} %{width=0.32\linewidth}
			\centering
			\smallskip
			\includegraphics[height=4cm,width=0.3\linewidth]{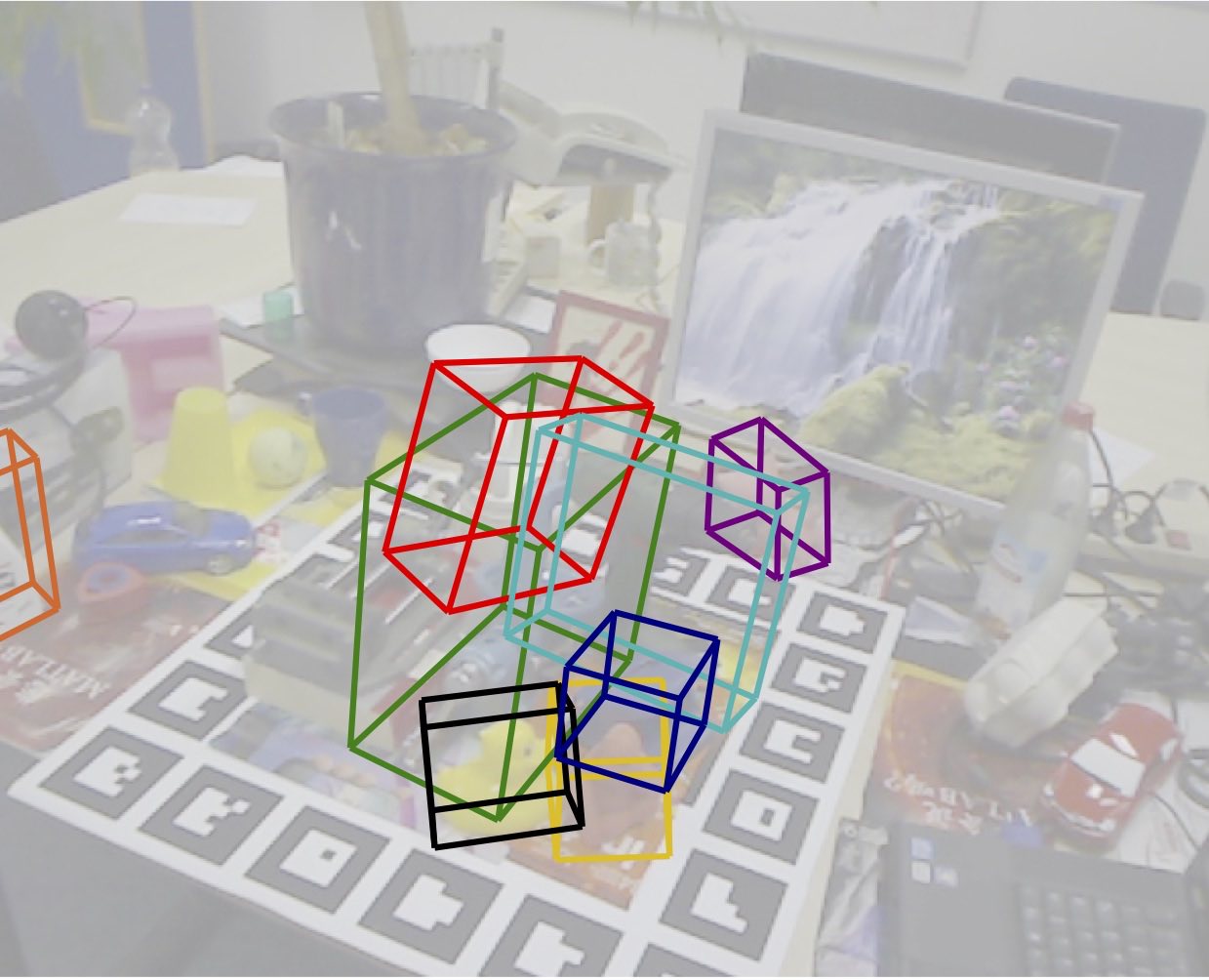}
		\end{subfigure}
		&
		\raisebox{1.8cm}{
			\begin{tabular}{ccc}% if you add [t], than sub images are pushed down
				\smallskip
				\begin{subfigure} %{width=0.1\linewidth}
					\centering
					\includegraphics[height=1.85cm,width=0.1\linewidth]{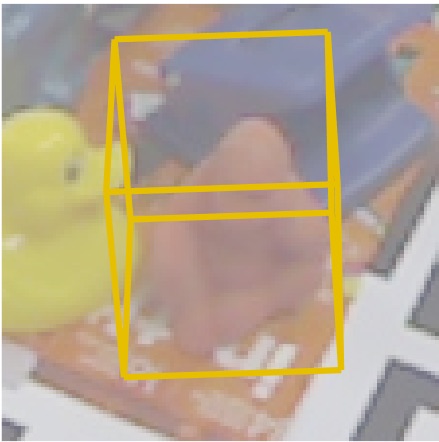}
				\end{subfigure} \hspace{-6mm}
				&\begin{subfigure} %{width=0.1\linewidth}
					\centering
					\includegraphics[height=1.85cm,width=0.1\linewidth]{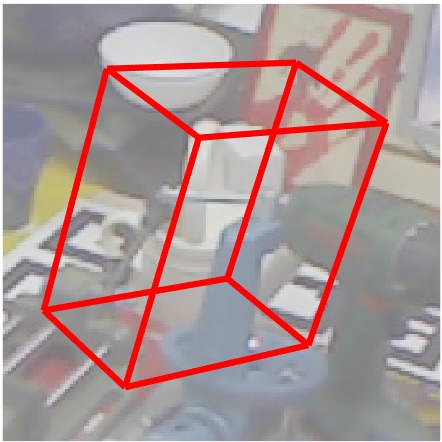}
				\end{subfigure} \hspace{-6mm}
				&\begin{subfigure}%{width=0.1\linewidth}
					\centering
					\includegraphics[height=1.85cm,width=0.1\linewidth]{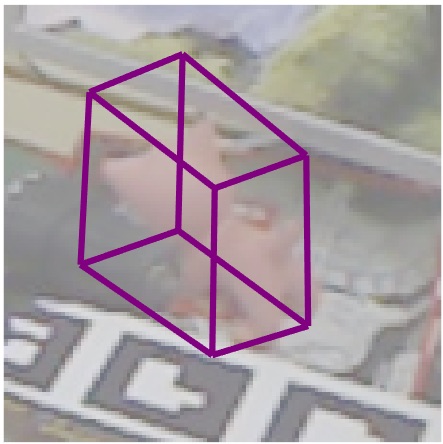}
				\end{subfigure}
				\\
				\begin{subfigure} %{width=0.1\linewidth}
					\centering
					\includegraphics[height=1.85cm,width=0.1\linewidth]{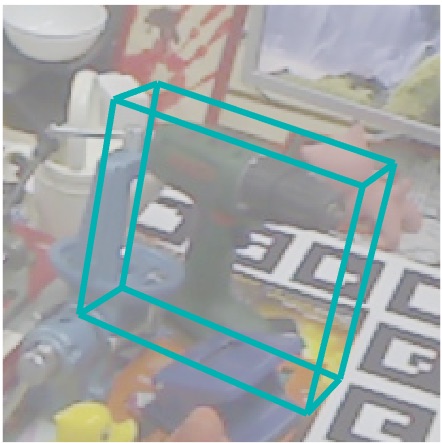}
				\end{subfigure} \hspace{-6mm}
				&\begin{subfigure} %{width=0.1\linewidth}
					\centering
					\includegraphics[height=1.85cm,width=0.1\linewidth]{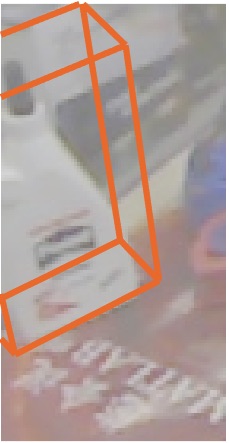}
				\end{subfigure} \hspace{-6mm}
				&\begin{subfigure} %{width=0.1\linewidth}
					\centering
					\includegraphics[height=1.85cm,width=0.1\linewidth]{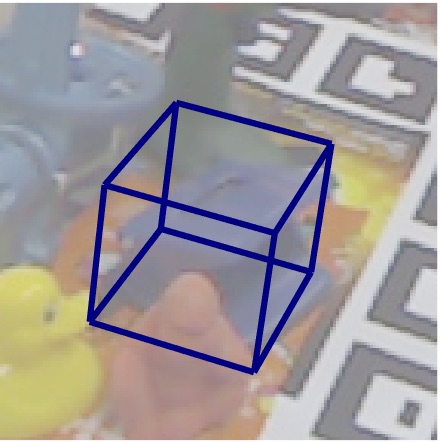}
				\end{subfigure}
		\end{tabular}} \\
		\vspace{-7.5mm}
	\end{tabular}
	\caption{Results on the \textsc{Occlusion} dataset. Our method is quite robust against severe occlusions in the presence of scene clutter and rotational pose ambiguity for symmetric objects. (left) Input images, (middle) 6D pose predictions of multiple objects, (right) A magnified view of the individual 6D pose estimates of six different objects is shown for clarity.
		In each case, the 3D bounding box is rendered on the input image. The following color coding is used -- \textsc{Ape} (gold), \textsc{Benchvise} (green), \textsc{Can} (red), \textsc{Cat} (purple), \textsc{Driller} (cyan), \textsc{Duck} (black), \textsc{Glue} (orange), \textsc{Holepuncher} (blue). In addition to the objects from the~\textsc{Occlusion} dataset, we 
		also visualize the pose predictions of the~\emph{Benchvise} object from the~\textsc{LineMod} dataset. As in~\cite{rad2017}, we do not evaluate on the \emph{Eggbox} object, as more than $70 \%$ of close poses are not seen in the training sequence. This image is best viewed on a computer screen.}
	\vspace{-5mm}
	\label{fig:supp_occlusion1}
\end{figure*}

\begin{figure*}[!htb]
	\centering
	\begin{tabular}[t]{ccc}
		%1
		\begin{subfigure} %{width=0.32\linewidth}
			\centering
			\smallskip
			\includegraphics[height=4cm,width=0.3\linewidth]{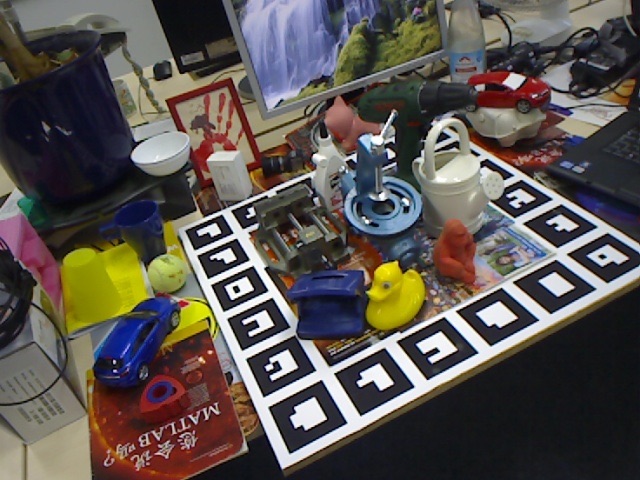}
		\end{subfigure}
		&
		\begin{subfigure} %{width=0.32\linewidth}
			\centering
			\smallskip
			\includegraphics[height=4cm,width=0.3\linewidth]{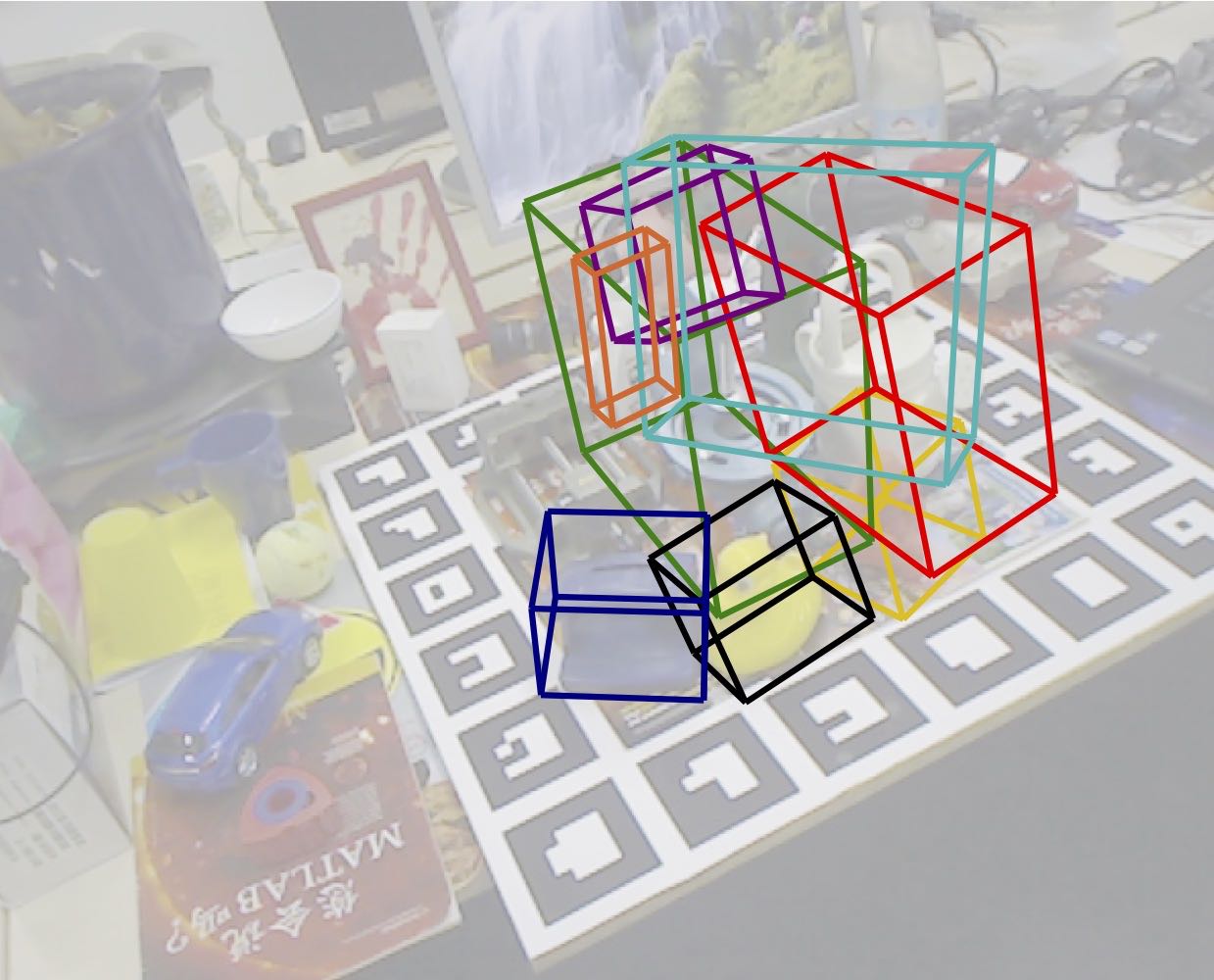}
		\end{subfigure}
		&
		\raisebox{1.8cm}{
			\begin{tabular}{ccc}% if you add [t], than sub images are pushed down
				\smallskip
				\begin{subfigure} %{width=0.1\linewidth}
					\centering
					\includegraphics[height=1.85cm,width=0.1\linewidth]{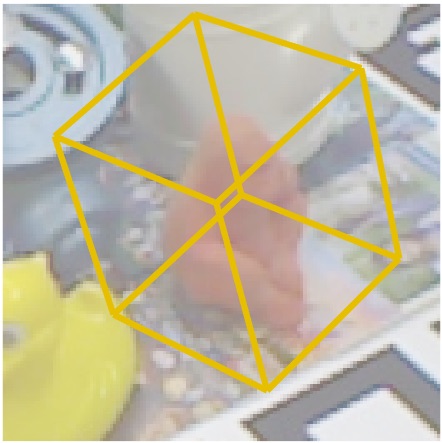}
				\end{subfigure} \hspace{-6mm}
				&\begin{subfigure} %{width=0.1\linewidth}
					\centering
					\includegraphics[height=1.85cm,width=0.1\linewidth]{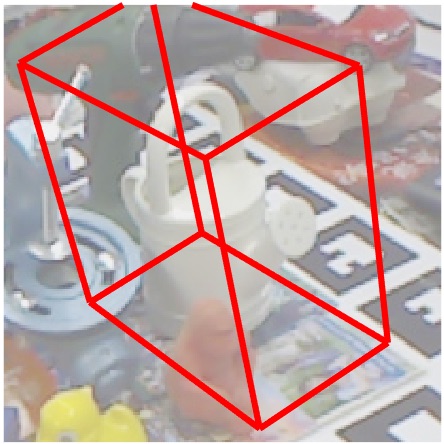}
				\end{subfigure} \hspace{-6mm}
				&\begin{subfigure}%{width=0.1\linewidth}
					\centering
					\includegraphics[height=1.85cm,width=0.1\linewidth]{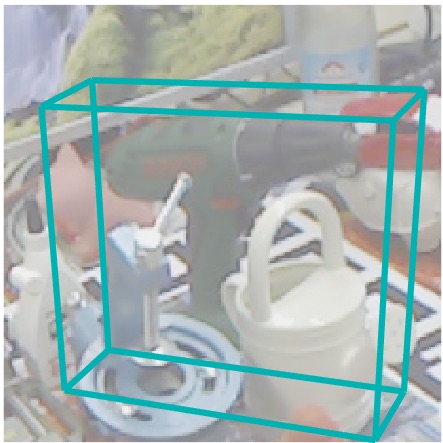}
				\end{subfigure}
				\\
				\begin{subfigure} %{width=0.1\linewidth}
					\centering
					\includegraphics[height=1.85cm,width=0.1\linewidth]{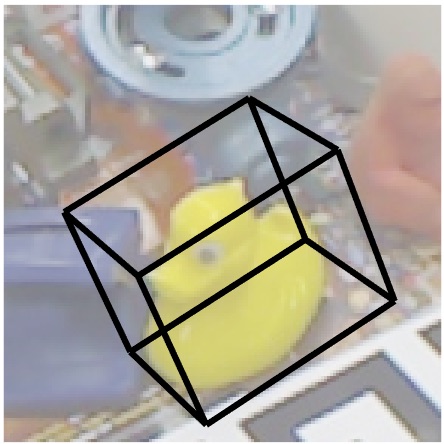}
				\end{subfigure} \hspace{-6mm}
				&\begin{subfigure} %{width=0.1\linewidth}
					\centering
					\includegraphics[height=1.85cm,width=0.1\linewidth]{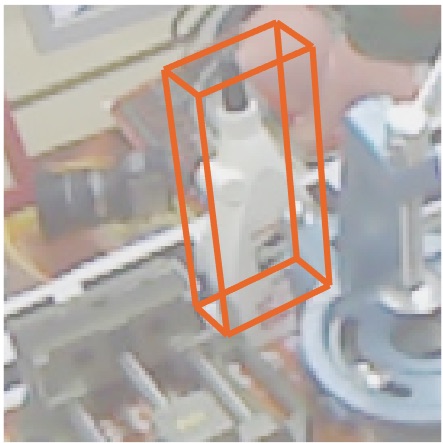}
				\end{subfigure} \hspace{-6mm}
				&\begin{subfigure} %{width=0.1\linewidth}
					\centering
					\includegraphics[height=1.85cm,width=0.1\linewidth]{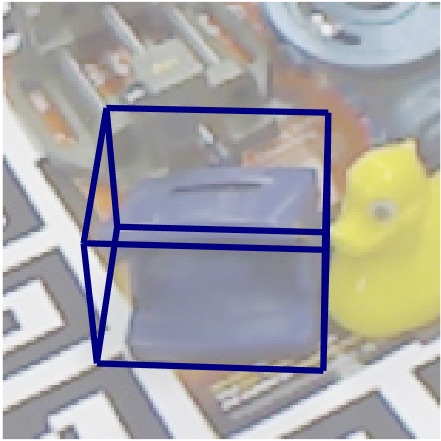}
				\end{subfigure}
		\end{tabular}}\\ %\vspace{2mm}
		%2	
		\begin{subfigure} %{width=0.32\linewidth}
			\centering
			\smallskip
			\includegraphics[height=4cm,width=0.3\linewidth]{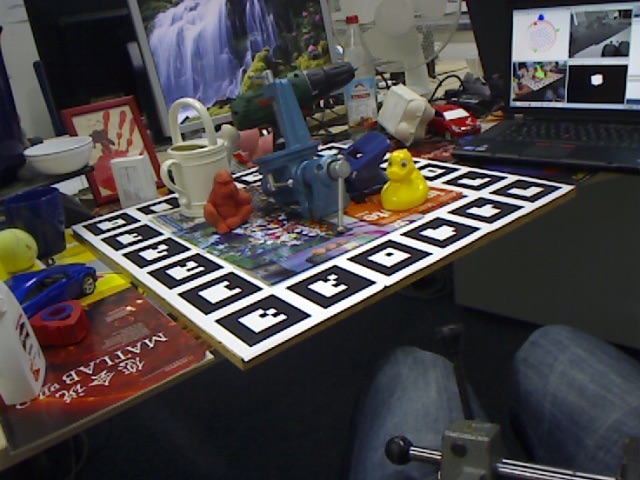}
		\end{subfigure}
		&
		\begin{subfigure} %{width=0.32\linewidth}
			\centering
			\smallskip
			\includegraphics[height=4cm,width=0.3\linewidth]{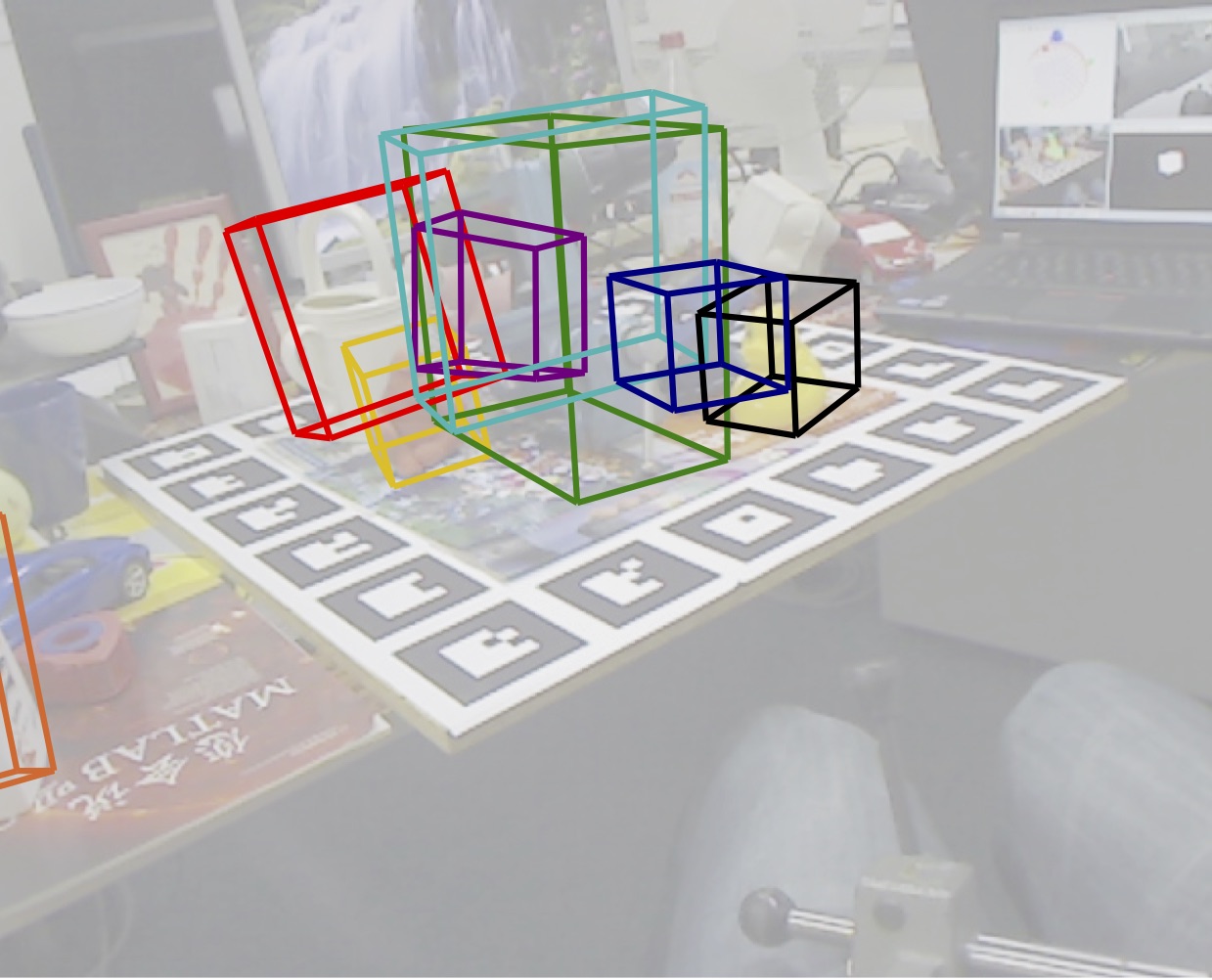}
		\end{subfigure}
		&
		\raisebox{1.8cm}{
			\begin{tabular}{ccc}% if you add [t], than sub images are pushed down
				\smallskip
				\begin{subfigure} %{width=0.1\linewidth}
					\centering
					\includegraphics[height=1.85cm,width=0.1\linewidth]{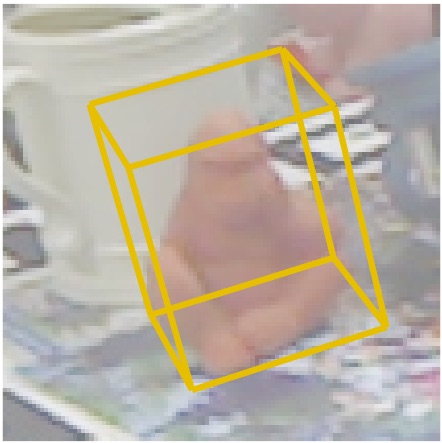}
				\end{subfigure} \hspace{-6mm}
				&\begin{subfigure} %{width=0.1\linewidth}
					\centering
					\includegraphics[height=1.85cm,width=0.1\linewidth]{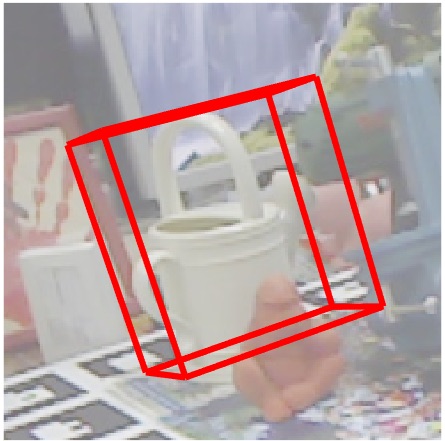}
				\end{subfigure} \hspace{-6mm}
				&\begin{subfigure}%{width=0.1\linewidth}
					\centering
					\includegraphics[height=1.85cm,width=0.1\linewidth]{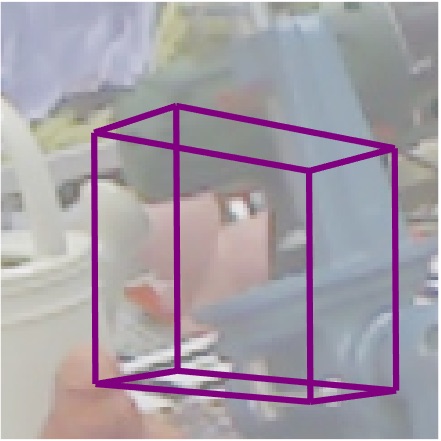}
				\end{subfigure}
				\\
				\begin{subfigure} %{width=0.1\linewidth}
					\centering
					\includegraphics[height=1.85cm,width=0.1\linewidth]{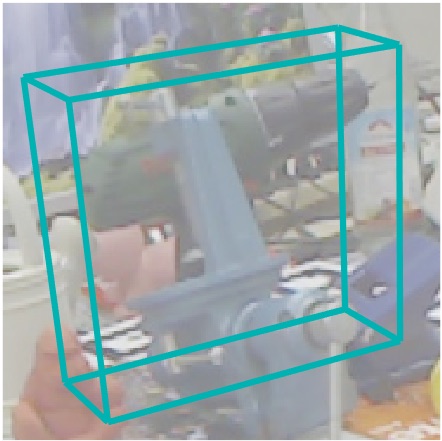}
				\end{subfigure} \hspace{-6mm}
				&\begin{subfigure} %{width=0.1\linewidth}
					\centering
					\includegraphics[height=1.85cm,width=0.1\linewidth]{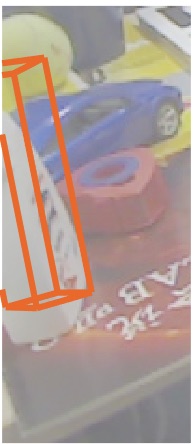}
				\end{subfigure} \hspace{-6mm}
				&\begin{subfigure} %{width=0.1\linewidth}
					\centering
					\includegraphics[height=1.85cm,width=0.1\linewidth]{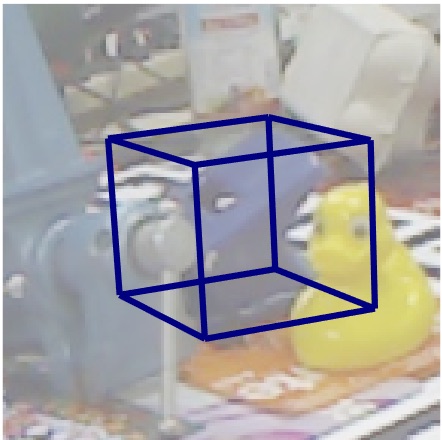}
				\end{subfigure}
		\end{tabular}}\\ %\vspace{2mm}
		%3
		\begin{subfigure} %{width=0.32\linewidth}
			\centering
			\smallskip
			\includegraphics[height=4cm,width=0.3\linewidth]{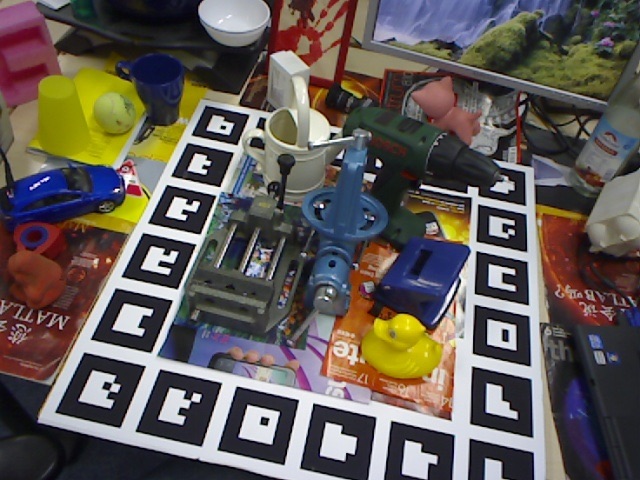}
		\end{subfigure}
		&
		\begin{subfigure} %{width=0.32\linewidth}
			\centering
			\smallskip
			\includegraphics[height=4cm,width=0.3\linewidth]{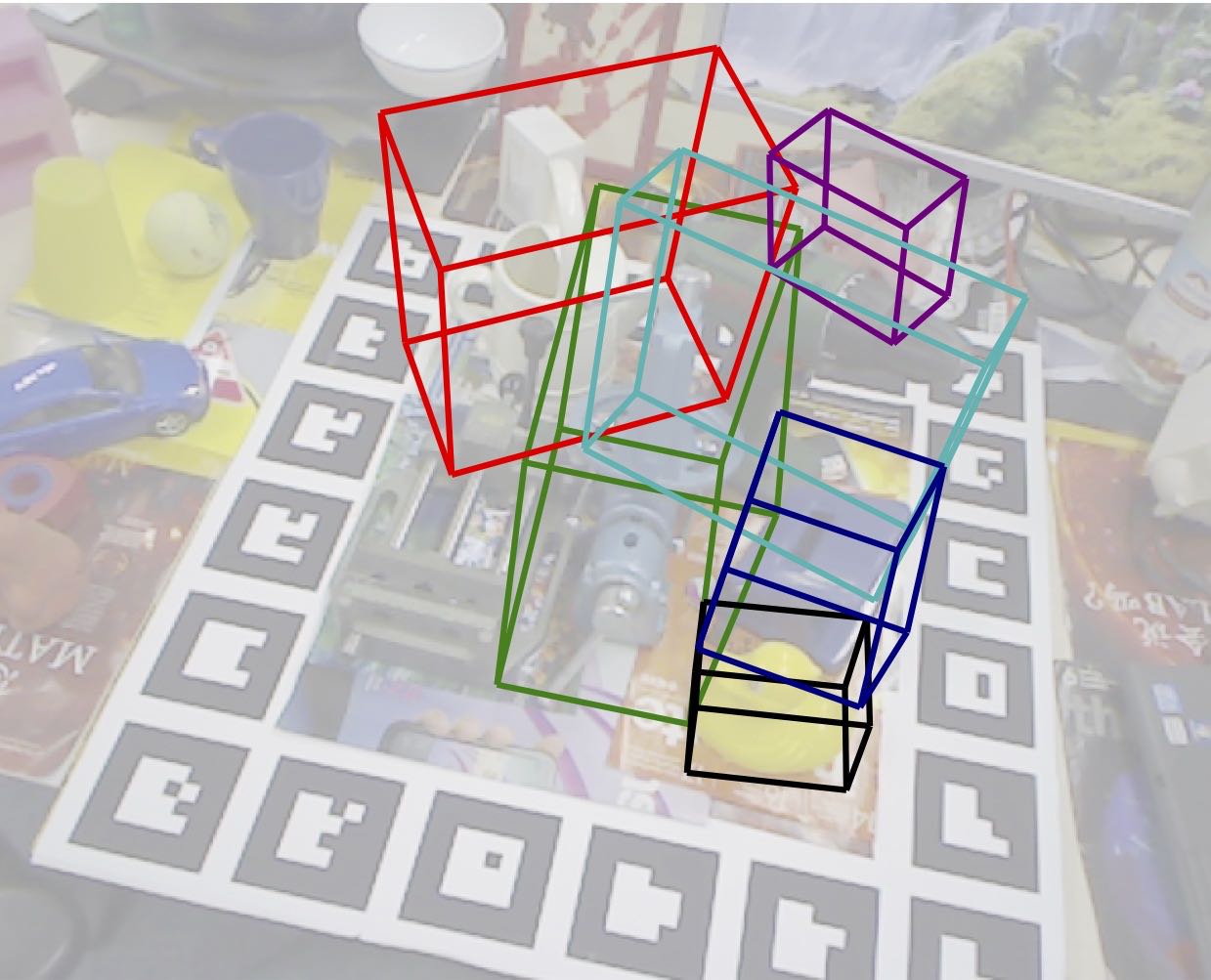}
		\end{subfigure}
		&
		\raisebox{1.8cm}{
			\begin{tabular}{ccc}% if you add [t], than sub images are pushed down
				\smallskip
				\begin{subfigure} %{width=0.1\linewidth}
					\centering
					\includegraphics[height=1.85cm,width=0.1\linewidth]{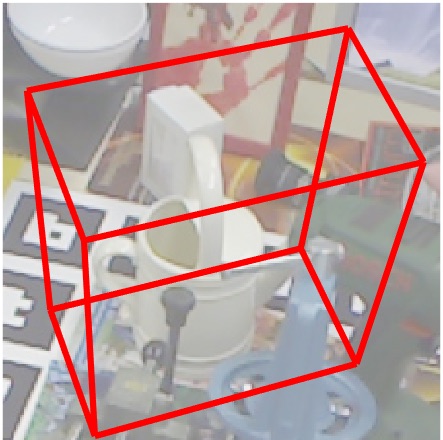}
				\end{subfigure} \hspace{-6mm}
				&\begin{subfigure} %{width=0.1\linewidth}
					\centering
					\includegraphics[height=1.85cm,width=0.1\linewidth]{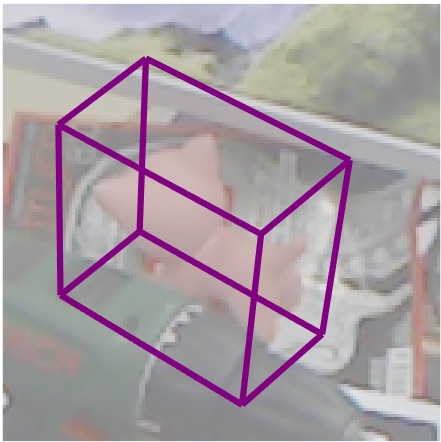}
				\end{subfigure} \hspace{-6mm}
				&\begin{subfigure}%{width=0.1\linewidth}
					\centering
					\includegraphics[height=1.85cm,width=0.1\linewidth]{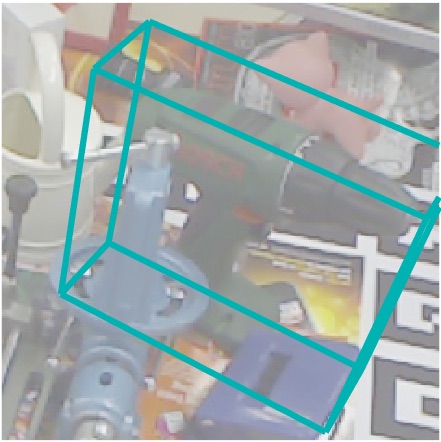}
				\end{subfigure}
				\\
				\begin{subfigure} %{width=0.1\linewidth}
					\centering
					\includegraphics[height=1.85cm,width=0.1\linewidth]{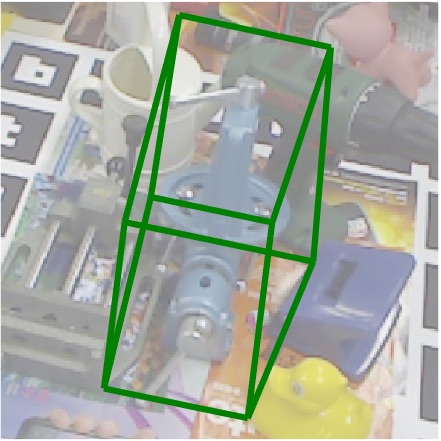}
				\end{subfigure} \hspace{-6mm}
				&\begin{subfigure} %{width=0.1\linewidth}
					\centering
					\includegraphics[height=1.85cm,width=0.1\linewidth]{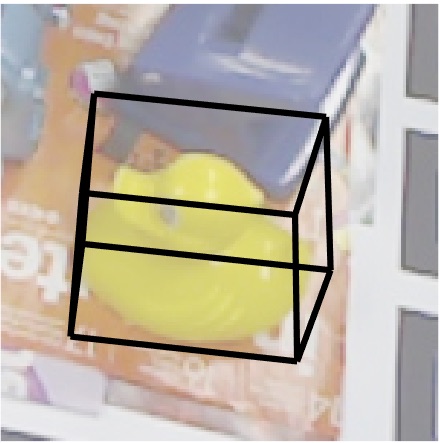}
				\end{subfigure} \hspace{-6mm}
				&\begin{subfigure} %{width=0.1\linewidth}
					\centering
					\includegraphics[height=1.85cm,width=0.1\linewidth]{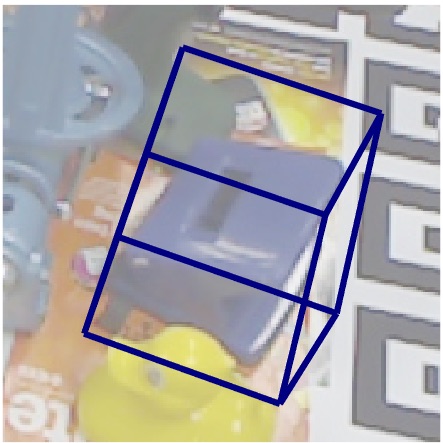}
				\end{subfigure}
		\end{tabular}}\\ %\vspace{2mm}
		%4
		\begin{subfigure} %{width=0.32\linewidth}
			\centering
			\smallskip
			\includegraphics[height=4cm,width=0.3\linewidth]{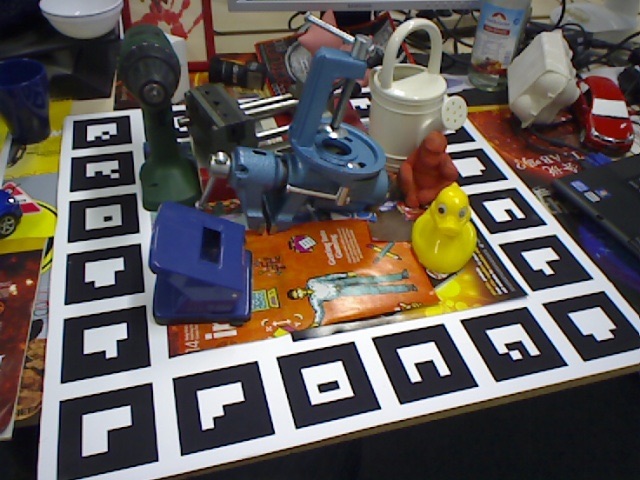}
		\end{subfigure}
		&
		\begin{subfigure} %{width=0.32\linewidth}
			\centering
			\smallskip
			\includegraphics[height=4cm,width=0.3\linewidth]{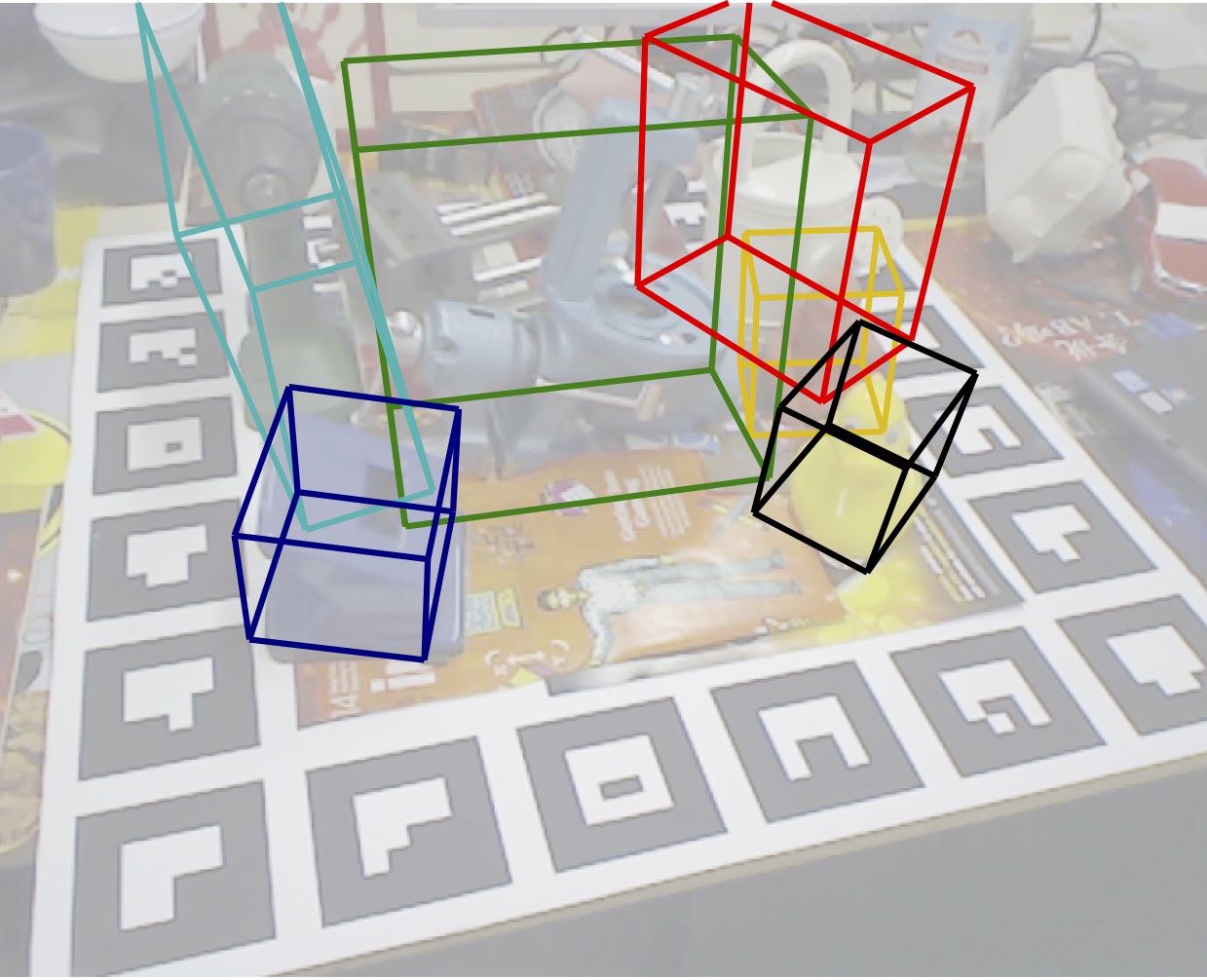}
		\end{subfigure}
		&
		\raisebox{1.8cm}{
			\begin{tabular}{ccc}% if you add [t], than sub images are pushed down
				\smallskip
				\begin{subfigure} %{width=0.1\linewidth}
					\centering
					\includegraphics[height=1.85cm,width=0.1\linewidth]{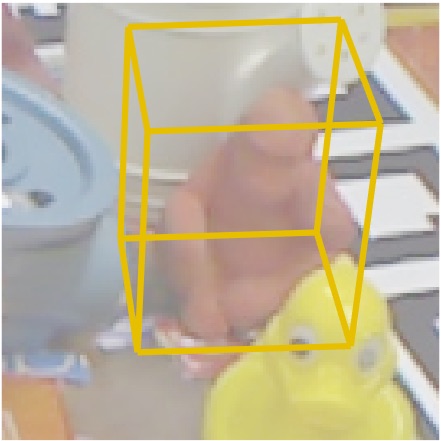}
				\end{subfigure} \hspace{-6mm}
				&\begin{subfigure} %{width=0.1\linewidth}
					\centering
					\includegraphics[height=1.85cm,width=0.1\linewidth]{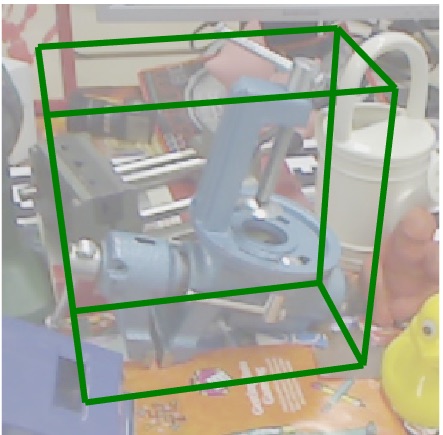}
				\end{subfigure} \hspace{-6mm}
				&\begin{subfigure}%{width=0.1\linewidth}
					\centering
					\includegraphics[height=1.85cm,width=0.1\linewidth]{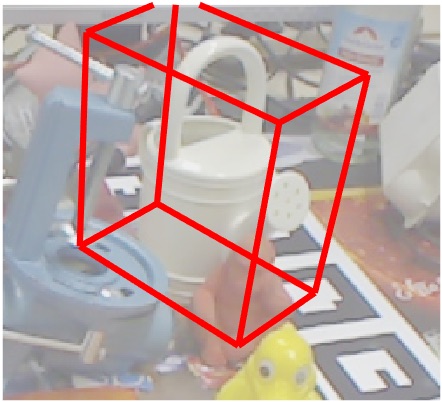}
				\end{subfigure}
				\\
				\begin{subfigure} %{width=0.1\linewidth}
					\centering
					\includegraphics[height=1.85cm,width=0.1\linewidth]{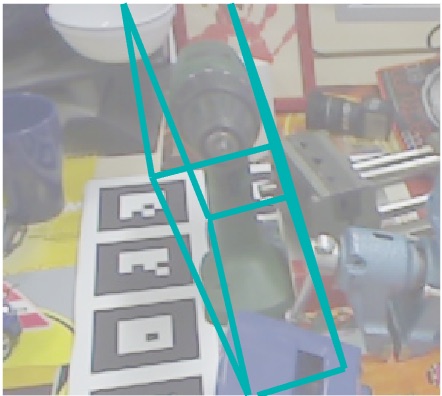}
				\end{subfigure} \hspace{-6mm}
				&\begin{subfigure} %{width=0.1\linewidth}
					\centering
					\includegraphics[height=1.85cm,width=0.1\linewidth]{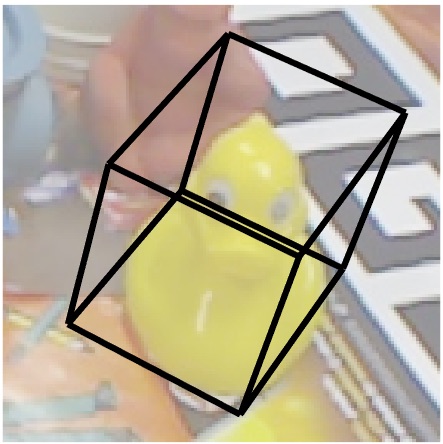}
				\end{subfigure} \hspace{-6mm}
				&\begin{subfigure} %{width=0.1\linewidth}
					\centering
					\includegraphics[height=1.85cm,width=0.1\linewidth]{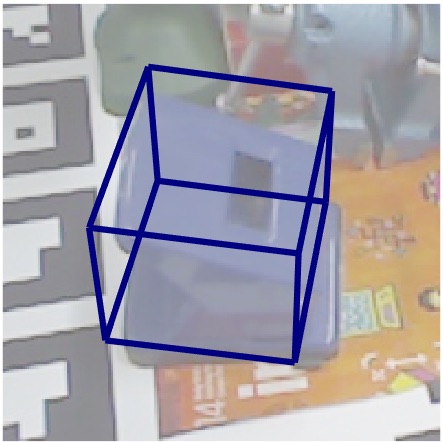}
				\end{subfigure}
		\end{tabular}} \\
		\vspace{-7.5mm}
	\end{tabular}
	\caption{Results on the \textsc{Occlusion} dataset. Our method is quite robust against severe occlusions in the presence of scene clutter and rotational pose ambiguity for symmetric objects. (left) Input images, (middle) 6D pose predictions of multiple objects, (right) A magnified view of the individual 6D pose estimates of six different objects is shown for clarity.
		In each case, the 3D bounding box is rendered on the input image. The following color coding is used -- \textsc{Ape} (gold), \textsc{Benchvise} (green), \textsc{Can} (red), \textsc{Cat} (purple), \textsc{Driller} (cyan), \textsc{Duck} (black), \textsc{Glue} (orange), \textsc{Holepuncher} (blue). In addition to the objects from the~\textsc{Occlusion} dataset, we 
		also visualize the pose predictions of the~\emph{Benchvise} object from the~\textsc{LineMod} dataset. As in~\cite{rad2017}, we do not evaluate on the \emph{Eggbox} object, as more than $70 \%$ of close poses are not seen in the training sequence. This image is best viewed on a computer screen.}
	\vspace{-5mm}
	\label{fig:supp_occlusion2}
\end{figure*}

\begin{figure*}[!htb]
	\centering
	\begin{tabular}[t]{cccccc}
		%1
		\begin{subfigure} %{width=0.32\linewidth}
			\centering
			\smallskip
			\includegraphics[height=3cm,width=0.15\linewidth]{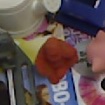}
		\end{subfigure} \hspace{-5.5mm}
		&
		\begin{subfigure} %{width=0.32\linewidth}
			\centering
			\smallskip
			\includegraphics[height=3cm,width=0.15\linewidth]{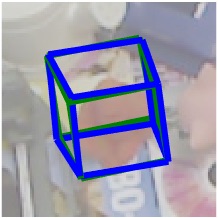}
		\end{subfigure}
		&
		\begin{subfigure} %{width=0.32\linewidth}
			\centering
			\smallskip
			\includegraphics[height=3cm,width=0.15\linewidth]{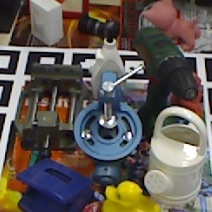}
		\end{subfigure} \hspace{-5.5mm}
		&
		\begin{subfigure} %{width=0.32\linewidth}
			\centering
			\smallskip
			\includegraphics[height=3cm,width=0.15\linewidth]{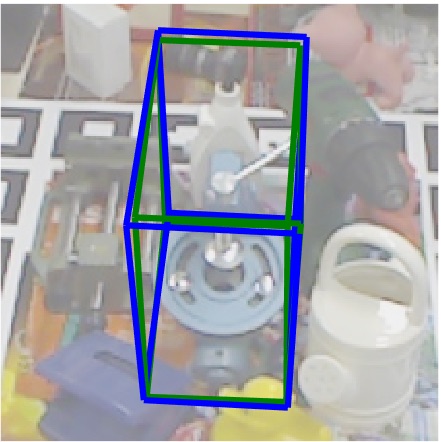}
		\end{subfigure}
		&
		\begin{subfigure} %{width=0.32\linewidth}
			\centering
			\smallskip
			\includegraphics[height=3cm,width=0.15\linewidth]{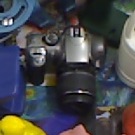}
		\end{subfigure} \hspace{-5.5mm}
		&
		\begin{subfigure} %{width=0.32\linewidth}
			\centering
			\smallskip
			\includegraphics[height=3cm,width=0.15\linewidth]{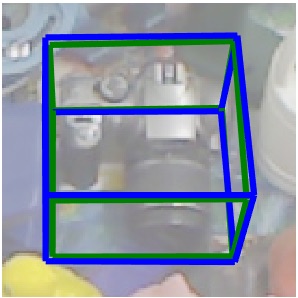}
		\end{subfigure} \\[-5mm]
		%2
		\begin{subfigure} %{width=0.32\linewidth}
			\centering
			\smallskip
			\includegraphics[height=3cm,width=0.15\linewidth]{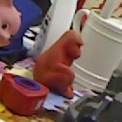}
		\end{subfigure} \hspace{-5.5mm}
		&
		\begin{subfigure} %{width=0.32\linewidth}
			\centering
			\smallskip
			\includegraphics[height=3cm,width=0.15\linewidth]{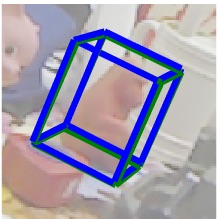}
		\end{subfigure}
		&
		\begin{subfigure} %{width=0.32\linewidth}
			\centering
			\smallskip
			\includegraphics[height=3cm,width=0.15\linewidth]{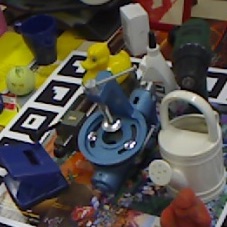}
		\end{subfigure} \hspace{-5.5mm}
		&
		\begin{subfigure} %{width=0.32\linewidth}
			\centering
			\smallskip
			\includegraphics[height=3cm,width=0.15\linewidth]{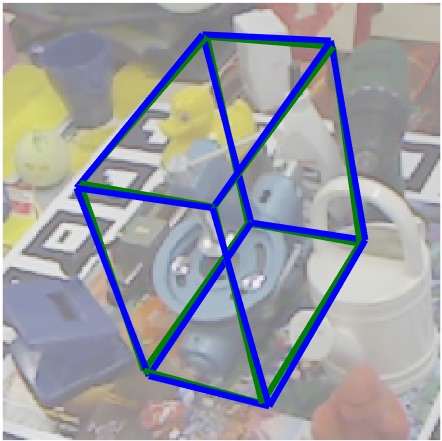}
		\end{subfigure}
		&
		\begin{subfigure} %{width=0.32\linewidth}
			\centering
			\smallskip
			\includegraphics[height=3cm,width=0.15\linewidth]{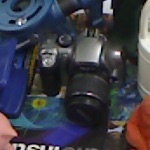}
		\end{subfigure} \hspace{-5.5mm}
		&
		\begin{subfigure} %{width=0.32\linewidth}
			\centering
			\smallskip
			\includegraphics[height=3cm,width=0.15\linewidth]{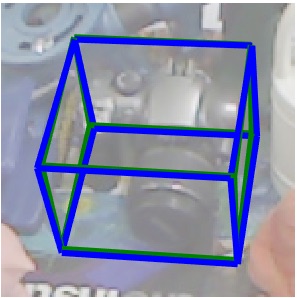}
		\end{subfigure} \\[-5mm]
		%3
		\begin{subfigure} %{width=0.32\linewidth}
			\centering
			\smallskip
			\includegraphics[height=3cm,width=0.15\linewidth]{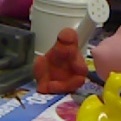}
		\end{subfigure} \hspace{-5.5mm}
		&
		\begin{subfigure} %{width=0.32\linewidth}
			\centering
			\smallskip
			\includegraphics[height=3cm,width=0.15\linewidth]{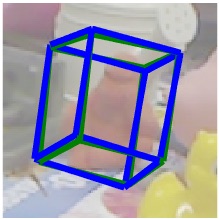}
		\end{subfigure}
		&
		\begin{subfigure} %{width=0.32\linewidth}
			\centering
			\smallskip
			\includegraphics[height=3cm,width=0.15\linewidth]{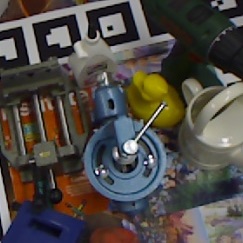}
		\end{subfigure} \hspace{-5.5mm}
		&
		\begin{subfigure} %{width=0.32\linewidth}
			\centering
			\smallskip
			\includegraphics[height=3cm,width=0.15\linewidth]{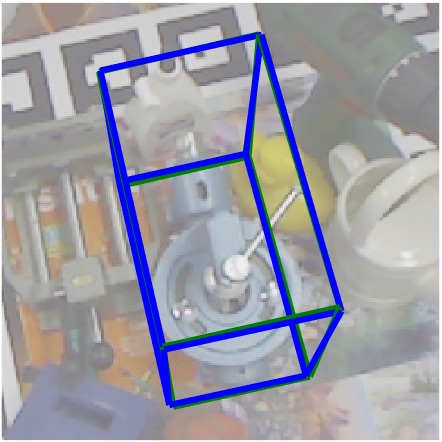}
		\end{subfigure}
		&
		\begin{subfigure} %{width=0.32\linewidth}
			\centering
			\smallskip
			\includegraphics[height=3cm,width=0.15\linewidth]{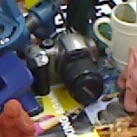}
		\end{subfigure} \hspace{-5.5mm}
		&
		\begin{subfigure} %{width=0.32\linewidth}
			\centering
			\smallskip
			\includegraphics[height=3cm,width=0.15\linewidth]{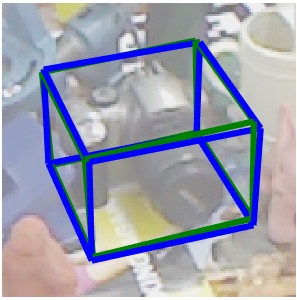}
		\end{subfigure} \\[-5mm]
		%4
		\begin{subfigure} %{width=0.32\linewidth}
			\centering
			\smallskip
			\includegraphics[height=3cm,width=0.15\linewidth]{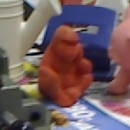}
		\end{subfigure} \hspace{-5.5mm}
		&
		\begin{subfigure} %{width=0.32\linewidth}
			\centering
			\smallskip
			\includegraphics[height=3cm,width=0.15\linewidth]{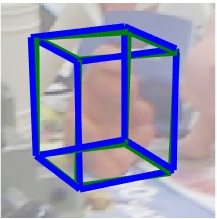}
		\end{subfigure}
		&
		\begin{subfigure} %{width=0.32\linewidth}
			\centering
			\smallskip
			\includegraphics[height=3cm,width=0.15\linewidth]{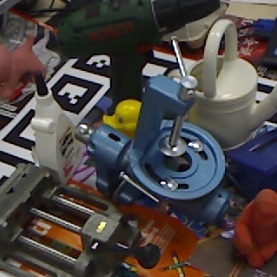}
		\end{subfigure} \hspace{-5.5mm}
		&
		\begin{subfigure} %{width=0.32\linewidth}
			\centering
			\smallskip
			\includegraphics[height=3cm,width=0.15\linewidth]{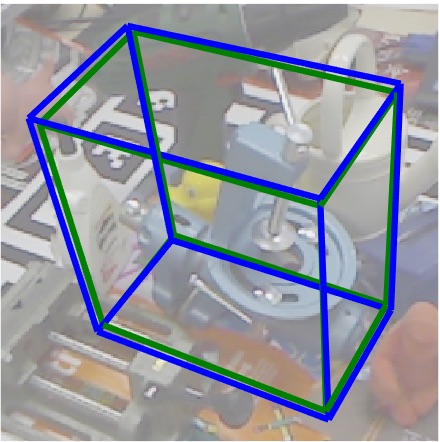}
		\end{subfigure}
		&
		\begin{subfigure} %{width=0.32\linewidth}
			\centering
			\smallskip
			\includegraphics[height=3cm,width=0.15\linewidth]{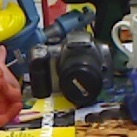}
		\end{subfigure} \hspace{-5.5mm}
		&
		\begin{subfigure} %{width=0.32\linewidth}
			\centering
			\smallskip
			\includegraphics[height=3cm,width=0.15\linewidth]{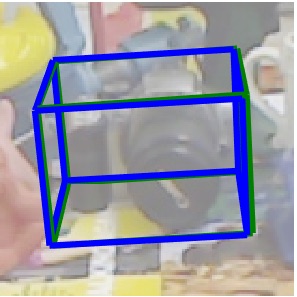}
		\end{subfigure} \\[-5mm]
		%5
		\begin{subfigure} %{width=0.32\linewidth}
			\centering
			\smallskip
			\includegraphics[height=3cm,width=0.15\linewidth]{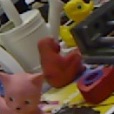}
		\end{subfigure} \hspace{-5.5mm}
		&
		\begin{subfigure} %{width=0.32\linewidth}
			\centering
			\smallskip
			\includegraphics[height=3cm,width=0.15\linewidth]{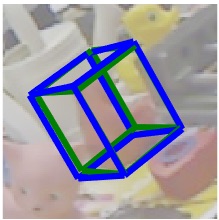}
		\end{subfigure}
		&
		\begin{subfigure} %{width=0.32\linewidth}
			\centering
			\smallskip
			\includegraphics[height=3cm,width=0.15\linewidth]{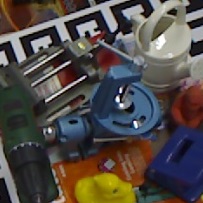}
		\end{subfigure} \hspace{-5.5mm}
		&
		\begin{subfigure} %{width=0.32\linewidth}
			\centering
			\smallskip
			\includegraphics[height=3cm,width=0.15\linewidth]{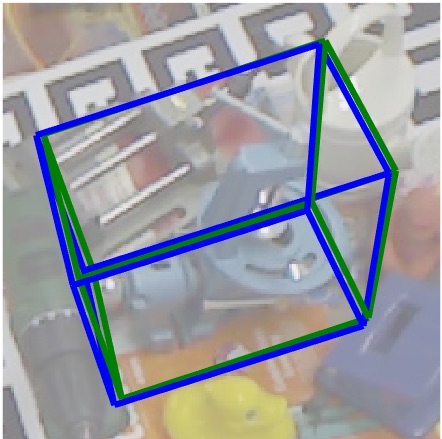}
		\end{subfigure}
		&
		\begin{subfigure} %{width=0.32\linewidth}
			\centering
			\smallskip
			\includegraphics[height=3cm,width=0.15\linewidth]{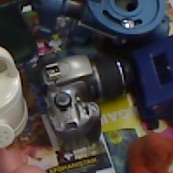}
		\end{subfigure} \hspace{-5.5mm}
		&
		\begin{subfigure} %{width=0.32\linewidth}
			\centering
			\smallskip
			\includegraphics[height=3cm,width=0.15\linewidth]{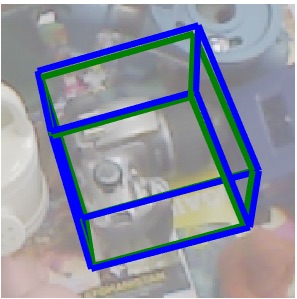}
		\end{subfigure} \\[-5mm]
		%6
		\begin{subfigure} %{width=0.32\linewidth}
			\centering
			\smallskip
			\includegraphics[height=3cm,width=0.15\linewidth]{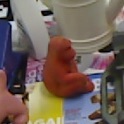}
		\end{subfigure} \hspace{-5.5mm}
		&
		\begin{subfigure} %{width=0.32\linewidth}
			\centering
			\smallskip
			\includegraphics[height=3cm,width=0.15\linewidth]{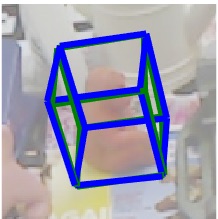}
		\end{subfigure}
		&
		\begin{subfigure} %{width=0.32\linewidth}
			\centering
			\smallskip
			\includegraphics[height=3cm,width=0.15\linewidth]{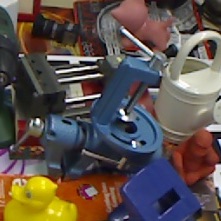}
		\end{subfigure} \hspace{-5.5mm}
		&
		\begin{subfigure} %{width=0.32\linewidth}
			\centering
			\smallskip
			\includegraphics[height=3cm,width=0.15\linewidth]{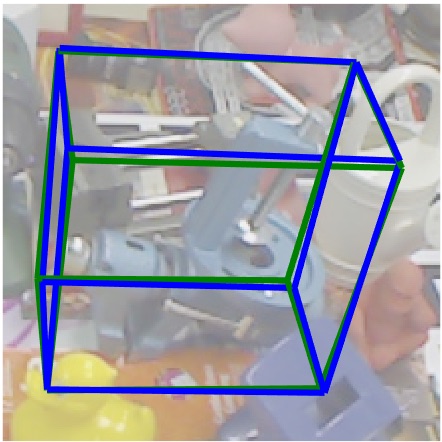}
		\end{subfigure}
		&
		\begin{subfigure} %{width=0.32\linewidth}
			\centering
			\smallskip
			\includegraphics[height=3cm,width=0.15\linewidth]{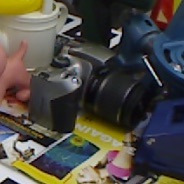}
		\end{subfigure} \hspace{-5.5mm}
		&
		\begin{subfigure} %{width=0.32\linewidth}
			\centering
			\smallskip
			\includegraphics[height=3cm,width=0.15\linewidth]{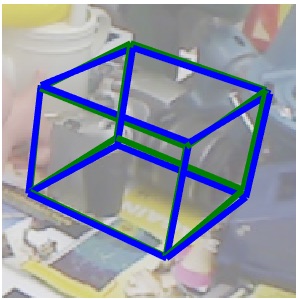}
		\end{subfigure} \\
	\end{tabular}
	
	\caption{Example results on the \textsc{LineMod} dataset: (left) \textsc{Ape}, (middle) \textsc{Benchvise}, (right) ~\textsc{Cam}. The projected 3D bounding boxes are rendered over the image and they have been cropped and resized for ease of visualization. The blue cuboid is rendered using our pose estimate whereas the green cuboid is rendered using the ground truth object pose. Note that the input image dimension is 640 $\times$ 480 pixels and the objects are often quite small. Noticeable scene clutter and occlusion makes these examples challenging.}
	\label{fig:supp_linemod1}
\end{figure*}

\begin{figure*}[!htb]
	\centering
	\begin{tabular}[t]{cccccc}
		%1
		\begin{subfigure} %{width=0.32\linewidth}
			\centering
			\smallskip
			\includegraphics[height=3cm,width=0.15\linewidth]{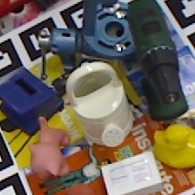}
		\end{subfigure} \hspace{-5.5mm}
		&
		\begin{subfigure} %{width=0.32\linewidth}
			\centering
			\smallskip
			\includegraphics[height=3cm,width=0.15\linewidth]{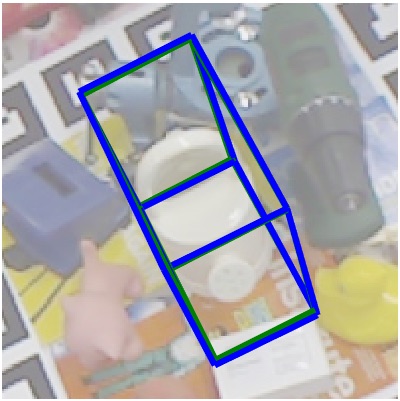}
		\end{subfigure}
		&
		\begin{subfigure} %{width=0.32\linewidth}
			\centering
			\smallskip
			\includegraphics[height=3cm,width=0.15\linewidth]{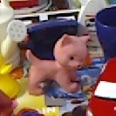}
		\end{subfigure} \hspace{-5.5mm}
		&
		\begin{subfigure} %{width=0.32\linewidth}
			\centering
			\smallskip
			\includegraphics[height=3cm,width=0.15\linewidth]{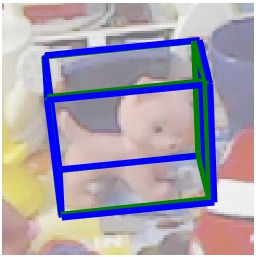}
		\end{subfigure}
		&
		\begin{subfigure} %{width=0.32\linewidth}
			\centering
			\smallskip
			\includegraphics[height=3cm,width=0.15\linewidth]{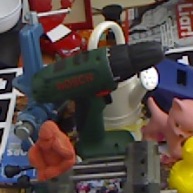}
		\end{subfigure} \hspace{-5.5mm}
		&
		\begin{subfigure} %{width=0.32\linewidth}
			\centering
			\smallskip
			\includegraphics[height=3cm,width=0.15\linewidth]{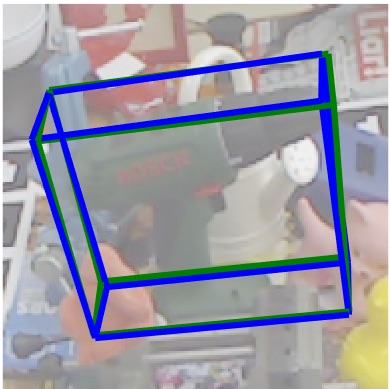}
		\end{subfigure} \\[-5mm]
		%2
		\begin{subfigure} %{width=0.32\linewidth}
			\centering
			\smallskip
			\includegraphics[height=3cm,width=0.15\linewidth]{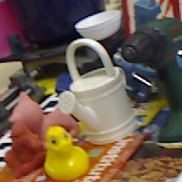}
		\end{subfigure} \hspace{-5.5mm}
		&
		\begin{subfigure} %{width=0.32\linewidth}
			\centering
			\smallskip
			\includegraphics[height=3cm,width=0.15\linewidth]{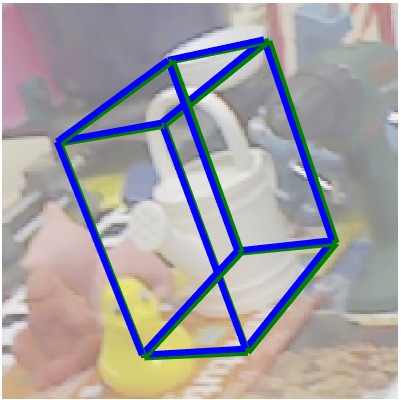}
		\end{subfigure}
		&
		\begin{subfigure} %{width=0.32\linewidth}
			\centering
			\smallskip
			\includegraphics[height=3cm,width=0.15\linewidth]{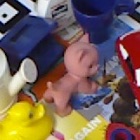}
		\end{subfigure} \hspace{-5.5mm}
		&
		\begin{subfigure} %{width=0.32\linewidth}
			\centering
			\smallskip
			\includegraphics[height=3cm,width=0.15\linewidth]{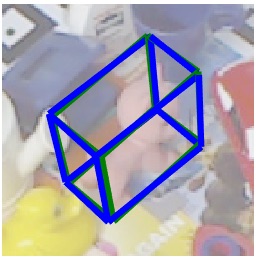}
		\end{subfigure}
		&
		\begin{subfigure} %{width=0.32\linewidth}
			\centering
			\smallskip
			\includegraphics[height=3cm,width=0.15\linewidth]{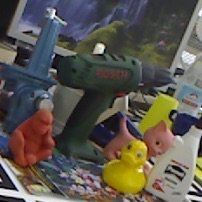}
		\end{subfigure} \hspace{-5.5mm}
		&
		\begin{subfigure} %{width=0.32\linewidth}
			\centering
			\smallskip
			\includegraphics[height=3cm,width=0.15\linewidth]{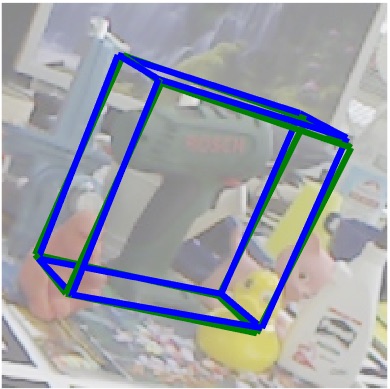}
		\end{subfigure} \\[-5mm]
		%3
		\begin{subfigure} %{width=0.32\linewidth}
			\centering
			\smallskip
			\includegraphics[height=3cm,width=0.15\linewidth]{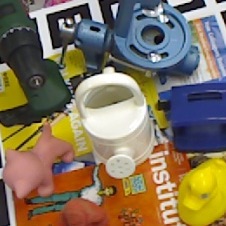}
		\end{subfigure} \hspace{-5.5mm}
		&
		\begin{subfigure} %{width=0.32\linewidth}
			\centering
			\smallskip
			\includegraphics[height=3cm,width=0.15\linewidth]{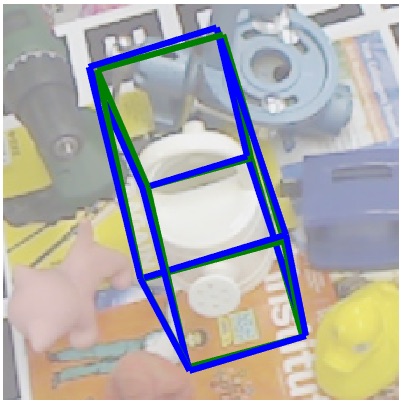}
		\end{subfigure}
		&
		\begin{subfigure} %{width=0.32\linewidth}
			\centering
			\smallskip
			\includegraphics[height=3cm,width=0.15\linewidth]{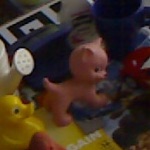}
		\end{subfigure} \hspace{-5.5mm}
		&
		\begin{subfigure} %{width=0.32\linewidth}
			\centering
			\smallskip
			\includegraphics[height=3cm,width=0.15\linewidth]{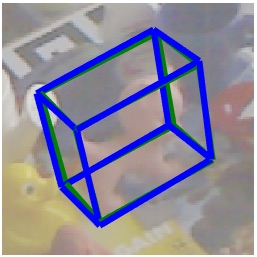}
		\end{subfigure}
		&
		\begin{subfigure} %{width=0.32\linewidth}
			\centering
			\smallskip
			\includegraphics[height=3cm,width=0.15\linewidth]{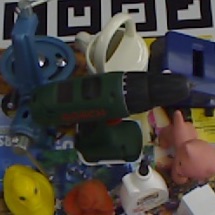}
		\end{subfigure} \hspace{-5.5mm}
		&
		\begin{subfigure} %{width=0.32\linewidth}
			\centering
			\smallskip
			\includegraphics[height=3cm,width=0.15\linewidth]{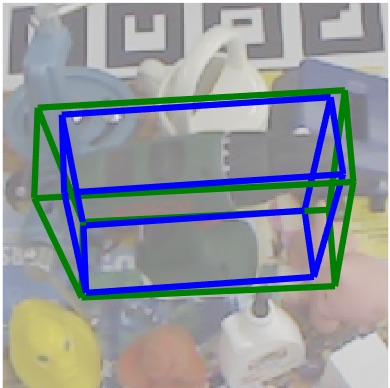}
		\end{subfigure} \\[-5mm]
		%4
		\begin{subfigure} %{width=0.32\linewidth}
			\centering
			\smallskip
			\includegraphics[height=3cm,width=0.15\linewidth]{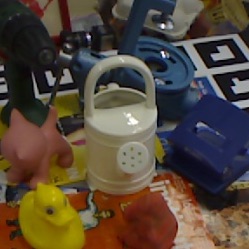}
		\end{subfigure} \hspace{-5.5mm}
		&
		\begin{subfigure} %{width=0.32\linewidth}
			\centering
			\smallskip
			\includegraphics[height=3cm,width=0.15\linewidth]{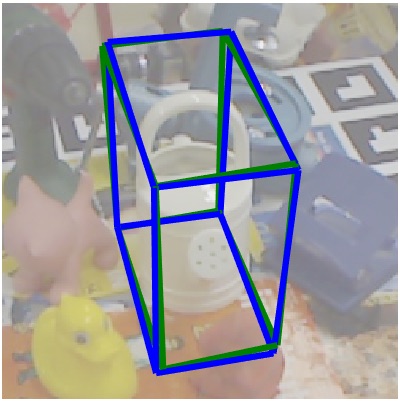}
		\end{subfigure}
		&
		\begin{subfigure} %{width=0.32\linewidth}
			\centering
			\smallskip
			\includegraphics[height=3cm,width=0.15\linewidth]{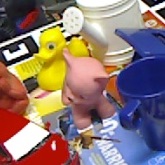}
		\end{subfigure} \hspace{-5.5mm}
		&
		\begin{subfigure} %{width=0.32\linewidth}
			\centering
			\smallskip
			\includegraphics[height=3cm,width=0.15\linewidth]{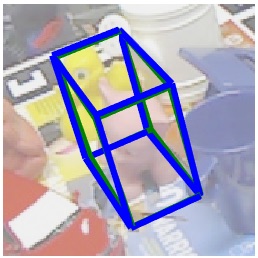}
		\end{subfigure}
		&
		\begin{subfigure} %{width=0.32\linewidth}
			\centering
			\smallskip
			\includegraphics[height=3cm,width=0.15\linewidth]{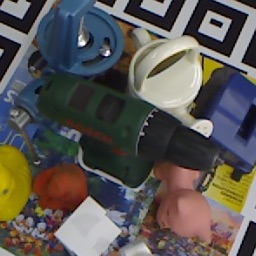}
		\end{subfigure} \hspace{-5.5mm}
		&
		\begin{subfigure} %{width=0.32\linewidth}
			\centering
			\smallskip
			\includegraphics[height=3cm,width=0.15\linewidth]{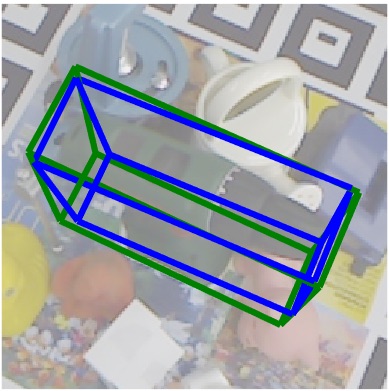}
		\end{subfigure} \\[-5mm]
		%5
		\begin{subfigure} %{width=0.32\linewidth}
			\centering
			\smallskip
			\includegraphics[height=3cm,width=0.15\linewidth]{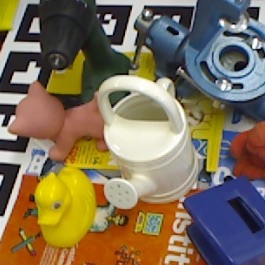}
		\end{subfigure} \hspace{-5.5mm}
		&
		\begin{subfigure} %{width=0.32\linewidth}
			\centering
			\smallskip
			\includegraphics[height=3cm,width=0.15\linewidth]{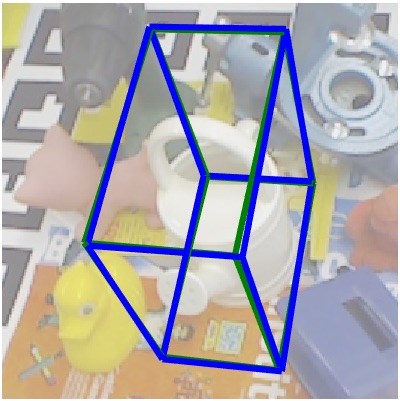}
		\end{subfigure}
		&
		\begin{subfigure} %{width=0.32\linewidth}
			\centering
			\smallskip
			\includegraphics[height=3cm,width=0.15\linewidth]{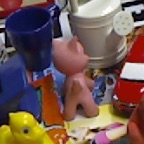}
		\end{subfigure} \hspace{-5.5mm}
		&
		\begin{subfigure} %{width=0.32\linewidth}
			\centering
			\smallskip
			\includegraphics[height=3cm,width=0.15\linewidth]{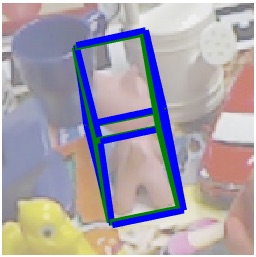}
		\end{subfigure}
		&
		\begin{subfigure} %{width=0.32\linewidth}
			\centering
			\smallskip
			\includegraphics[height=3cm,width=0.15\linewidth]{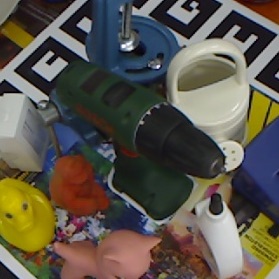}
		\end{subfigure} \hspace{-5.5mm}
		&
		\begin{subfigure} %{width=0.32\linewidth}
			\centering
			\smallskip
			\includegraphics[height=3cm,width=0.15\linewidth]{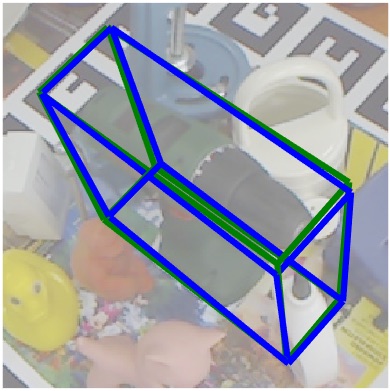}
		\end{subfigure} \\[-5mm]
		%6
		\begin{subfigure} %{width=0.32\linewidth}
			\centering
			\smallskip
			\includegraphics[height=3cm,width=0.15\linewidth]{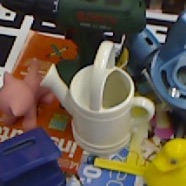}
		\end{subfigure} \hspace{-5.5mm}
		&
		\begin{subfigure} %{width=0.32\linewidth}
			\centering
			\smallskip
			\includegraphics[height=3cm,width=0.15\linewidth]{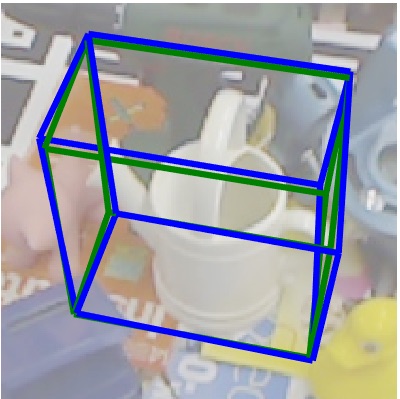}
		\end{subfigure}
		&
		\begin{subfigure} %{width=0.32\linewidth}
			\centering
			\smallskip
			\includegraphics[height=3cm,width=0.15\linewidth]{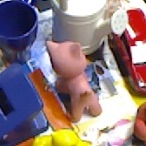}
		\end{subfigure} \hspace{-5.5mm}
		&
		\begin{subfigure} %{width=0.32\linewidth}
			\centering
			\smallskip
			\includegraphics[height=3cm,width=0.15\linewidth]{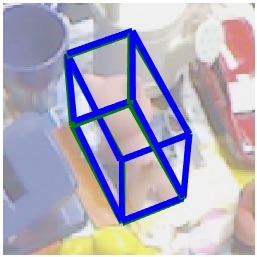}
		\end{subfigure}
		&
		\begin{subfigure} %{width=0.32\linewidth}
			\centering
			\smallskip
			\includegraphics[height=3cm,width=0.15\linewidth]{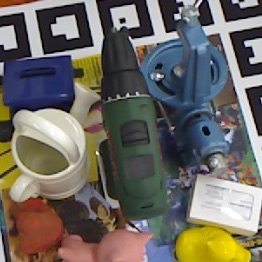}
		\end{subfigure} \hspace{-5.5mm}
		&
		\begin{subfigure} %{width=0.32\linewidth}
			\centering
			\smallskip
			\includegraphics[height=3cm,width=0.15\linewidth]{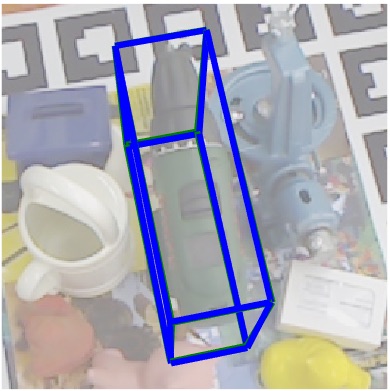}
		\end{subfigure} \\		
	\end{tabular}
	
	\caption{Example results on the \textsc{LineMod} dataset: (left) \textsc{Can}, (middle) \textsc{Cat}, (right) ~\textsc{Driller}. The projected 3D bounding boxes are rendered over the image and they have been cropped and resized for ease of visualization. The blue cuboid is rendered using our pose estimate whereas the green cuboid is rendered using the ground truth object pose. Note that the input image dimension is 640 $\times$ 480 pixels and the objects are often quite small. Noticeable scene clutter and occlusion makes these examples challenging.}
	\label{fig:supp_linemod2}
\end{figure*}

\begin{figure*}[!htb]
	\centering
	\begin{tabular}[t]{cccccc}
		%1
		\begin{subfigure} %{width=0.32\linewidth}
			\centering
			\smallskip
			\includegraphics[height=3cm,width=0.15\linewidth]{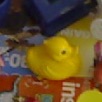}
		\end{subfigure} \hspace{-5.5mm}
		&
		\begin{subfigure} %{width=0.32\linewidth}
			\centering
			\smallskip
			\includegraphics[height=3cm,width=0.15\linewidth]{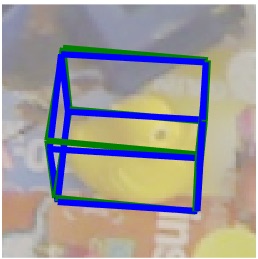}
		\end{subfigure}
		&
		\begin{subfigure} %{width=0.32\linewidth}
			\centering
			\smallskip
			\includegraphics[height=3cm,width=0.15\linewidth]{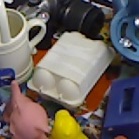}
		\end{subfigure} \hspace{-5.5mm}
		&
		\begin{subfigure} %{width=0.32\linewidth}
			\centering
			\smallskip
			\includegraphics[height=3cm,width=0.15\linewidth]{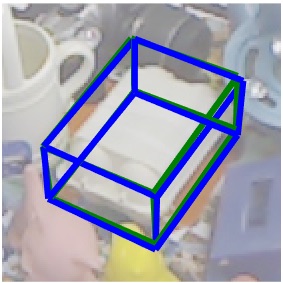}
		\end{subfigure}
		&
		\begin{subfigure} %{width=0.32\linewidth}
			\centering
			\smallskip
			\includegraphics[height=3cm,width=0.15\linewidth]{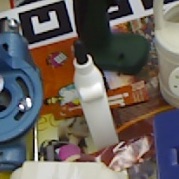}
		\end{subfigure} \hspace{-5.5mm}
		&
		\begin{subfigure} %{width=0.32\linewidth}
			\centering
			\smallskip
			\includegraphics[height=3cm,width=0.15\linewidth]{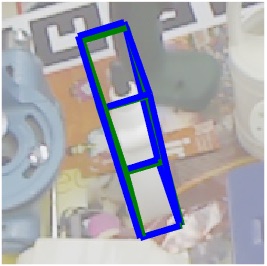}
		\end{subfigure} \\[-5mm]
		%2
		\begin{subfigure} %{width=0.32\linewidth}
			\centering
			\smallskip
			\includegraphics[height=3cm,width=0.15\linewidth]{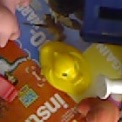}
		\end{subfigure} \hspace{-5.5mm}
		&
		\begin{subfigure} %{width=0.32\linewidth}
			\centering
			\smallskip
			\includegraphics[height=3cm,width=0.15\linewidth]{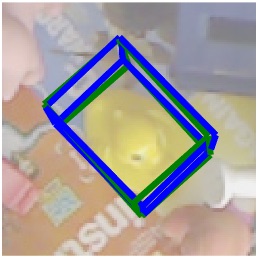}
		\end{subfigure}
		&
		\begin{subfigure} %{width=0.32\linewidth}
			\centering
			\smallskip
			\includegraphics[height=3cm,width=0.15\linewidth]{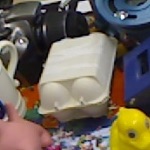}
		\end{subfigure} \hspace{-5.5mm}
		&
		\begin{subfigure} %{width=0.32\linewidth}
			\centering
			\smallskip
			\includegraphics[height=3cm,width=0.15\linewidth]{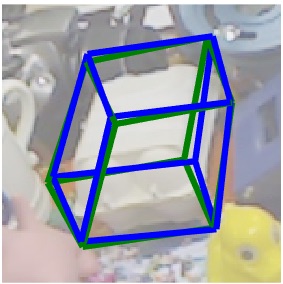}
		\end{subfigure}
		&
		\begin{subfigure} %{width=0.32\linewidth}
			\centering
			\smallskip
			\includegraphics[height=3cm,width=0.15\linewidth]{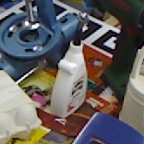}
		\end{subfigure} \hspace{-5.5mm}
		&
		\begin{subfigure} %{width=0.32\linewidth}
			\centering
			\smallskip
			\includegraphics[height=3cm,width=0.15\linewidth]{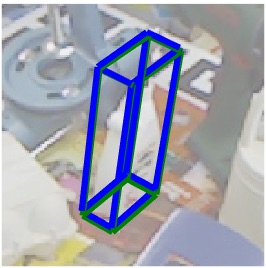}
		\end{subfigure} \\[-5mm]
		%3
		\begin{subfigure} %{width=0.32\linewidth}
			\centering
			\smallskip
			\includegraphics[height=3cm,width=0.15\linewidth]{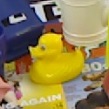}
		\end{subfigure} \hspace{-5.5mm}
		&
		\begin{subfigure} %{width=0.32\linewidth}
			\centering
			\smallskip
			\includegraphics[height=3cm,width=0.15\linewidth]{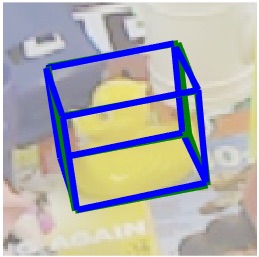}
		\end{subfigure}
		&
		\begin{subfigure} %{width=0.32\linewidth}
			\centering
			\smallskip
			\includegraphics[height=3cm,width=0.15\linewidth]{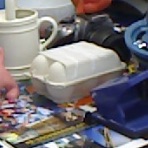}
		\end{subfigure} \hspace{-5.5mm}
		&
		\begin{subfigure} %{width=0.32\linewidth}
			\centering
			\smallskip
			\includegraphics[height=3cm,width=0.15\linewidth]{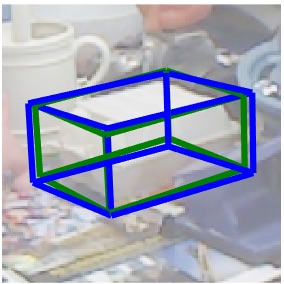}
		\end{subfigure}
		&
		\begin{subfigure} %{width=0.32\linewidth}
			\centering
			\smallskip
			\includegraphics[height=3cm,width=0.15\linewidth]{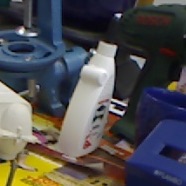}
		\end{subfigure} \hspace{-5.5mm}
		&
		\begin{subfigure} %{width=0.32\linewidth}
			\centering
			\smallskip
			\includegraphics[height=3cm,width=0.15\linewidth]{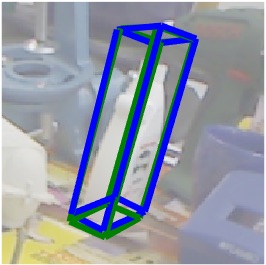}
		\end{subfigure} \\[-5mm]
		%4
		\begin{subfigure} %{width=0.32\linewidth}
			\centering
			\smallskip
			\includegraphics[height=3cm,width=0.15\linewidth]{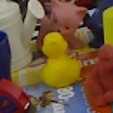}
		\end{subfigure} \hspace{-5.5mm}
		&
		\begin{subfigure} %{width=0.32\linewidth}
			\centering
			\smallskip
			\includegraphics[height=3cm,width=0.15\linewidth]{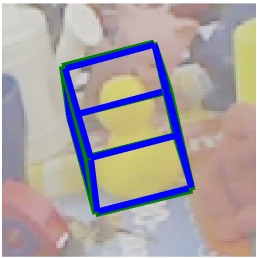}
		\end{subfigure}
		&
		\begin{subfigure} %{width=0.32\linewidth}
			\centering
			\smallskip
			\includegraphics[height=3cm,width=0.15\linewidth]{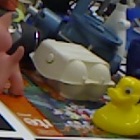}
		\end{subfigure} \hspace{-5.5mm}
		&
		\begin{subfigure} %{width=0.32\linewidth}
			\centering
			\smallskip
			\includegraphics[height=3cm,width=0.15\linewidth]{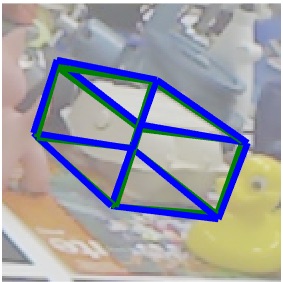}
		\end{subfigure}
		&
		\begin{subfigure} %{width=0.32\linewidth}
			\centering
			\smallskip
			\includegraphics[height=3cm,width=0.15\linewidth]{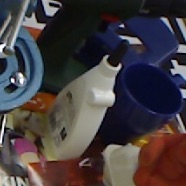}
		\end{subfigure} \hspace{-5.5mm}
		&
		\begin{subfigure} %{width=0.32\linewidth}
			\centering
			\smallskip
			\includegraphics[height=3cm,width=0.15\linewidth]{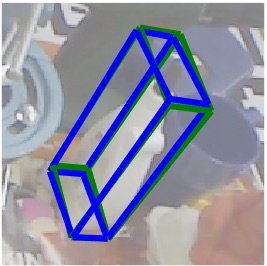}
		\end{subfigure} \\[-5mm]
		%5
		\begin{subfigure} %{width=0.32\linewidth}
			\centering
			\smallskip
			\includegraphics[height=3cm,width=0.15\linewidth]{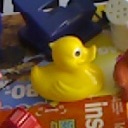}
		\end{subfigure} \hspace{-5.5mm}
		&
		\begin{subfigure} %{width=0.32\linewidth}
			\centering
			\smallskip
			\includegraphics[height=3cm,width=0.15\linewidth]{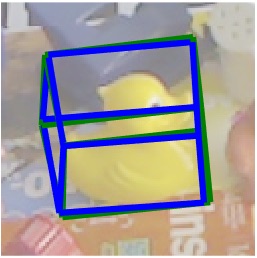}
		\end{subfigure}
		&
		\begin{subfigure} %{width=0.32\linewidth}
			\centering
			\smallskip
			\includegraphics[height=3cm,width=0.15\linewidth]{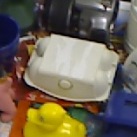}
		\end{subfigure} \hspace{-5.5mm}
		&
		\begin{subfigure} %{width=0.32\linewidth}
			\centering
			\smallskip
			\includegraphics[height=3cm,width=0.15\linewidth]{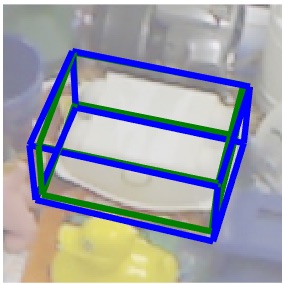}
		\end{subfigure}
		&
		\begin{subfigure} %{width=0.32\linewidth}
			\centering
			\smallskip
			\includegraphics[height=3cm,width=0.15\linewidth]{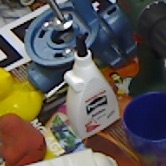}
		\end{subfigure} \hspace{-5.5mm}
		&
		\begin{subfigure} %{width=0.32\linewidth}
			\centering
			\smallskip
			\includegraphics[height=3cm,width=0.15\linewidth]{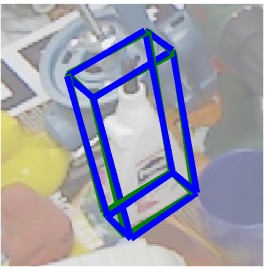}
		\end{subfigure} \\[-5mm]
		%6
		\begin{subfigure} %{width=0.32\linewidth}
			\centering
			\smallskip
			\includegraphics[height=3cm,width=0.15\linewidth]{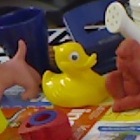}
		\end{subfigure} \hspace{-5.5mm}
		&
		\begin{subfigure} %{width=0.32\linewidth}
			\centering
			\smallskip
			\includegraphics[height=3cm,width=0.15\linewidth]{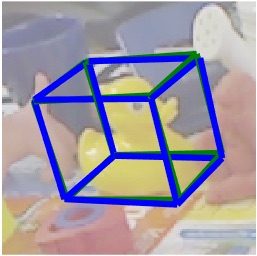}
		\end{subfigure}
		&
		\begin{subfigure} %{width=0.32\linewidth}
			\centering
			\smallskip
			\includegraphics[height=3cm,width=0.15\linewidth]{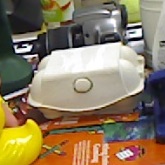}
		\end{subfigure} \hspace{-5.5mm}
		&
		\begin{subfigure} %{width=0.32\linewidth}
			\centering
			\smallskip
			\includegraphics[height=3cm,width=0.15\linewidth]{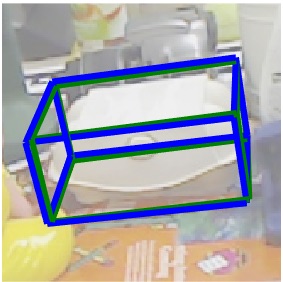}
		\end{subfigure}
		&
		\begin{subfigure} %{width=0.32\linewidth}
			\centering
			\smallskip
			\includegraphics[height=3cm,width=0.15\linewidth]{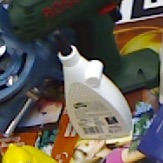}
		\end{subfigure} \hspace{-5.5mm}
		&
		\begin{subfigure} %{width=0.32\linewidth}
			\centering
			\smallskip
			\includegraphics[height=3cm,width=0.15\linewidth]{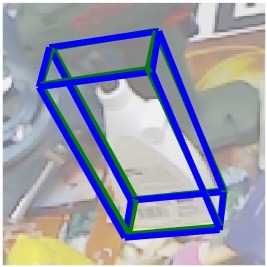}
		\end{subfigure} \\
		
	\end{tabular}
	
	\caption{Example results on the \textsc{LineMod} dataset: (left) \textsc{Duck}, (middle) \textsc{Eggbox}, (right) ~\textsc{Glue}. The projected 3D bounding boxes are rendered over the image and they have been cropped and resized for ease of visualization. The blue cuboid is rendered using our pose estimate whereas the green cuboid is rendered using the ground truth object pose. Note that the input image dimension is 640 $\times$ 480 pixels and the objects are often quite small. Noticeable scene clutter and occlusion makes these examples challenging.}
	\label{fig:supp_linemod3}
\end{figure*}

\begin{figure*}[!htb]
	\centering
	\begin{tabular}[t]{cccccc}
		%1
		\begin{subfigure} %{width=0.32\linewidth}
			\centering
			\smallskip
			\includegraphics[height=3cm,width=0.15\linewidth]{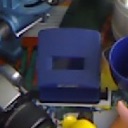}
		\end{subfigure} \hspace{-5.5mm}
		&
		\begin{subfigure} %{width=0.32\linewidth}
			\centering
			\smallskip
			\includegraphics[height=3cm,width=0.15\linewidth]{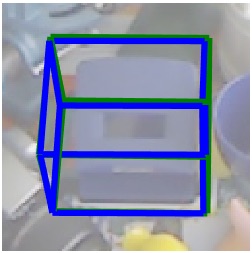}
		\end{subfigure}
		&
		\begin{subfigure} %{width=0.32\linewidth}
			\centering
			\smallskip
			\includegraphics[height=3cm,width=0.15\linewidth]{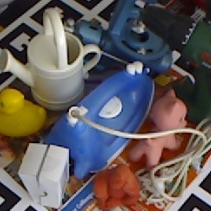}
		\end{subfigure} \hspace{-5.5mm}
		&
		\begin{subfigure} %{width=0.32\linewidth}
			\centering
			\smallskip
			\includegraphics[height=3cm,width=0.15\linewidth]{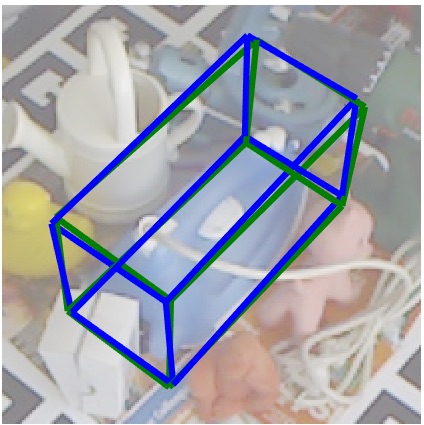}
		\end{subfigure}
		&
		\begin{subfigure} %{width=0.32\linewidth}
			\centering
			\smallskip
			\includegraphics[height=3cm,width=0.15\linewidth]{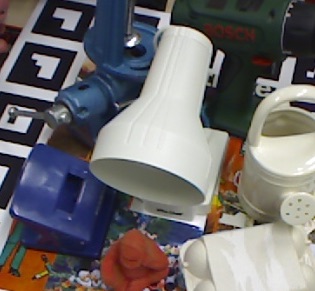}
		\end{subfigure} \hspace{-5.5mm}
		&
		\begin{subfigure} %{width=0.32\linewidth}
			\centering
			\smallskip
			\includegraphics[height=3cm,width=0.15\linewidth]{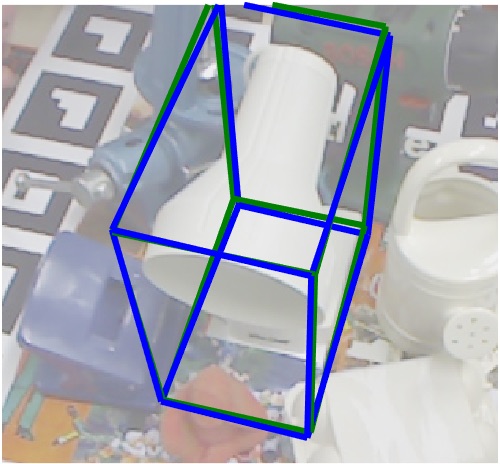}
		\end{subfigure} \\[-5mm]
		%2
		\begin{subfigure} %{width=0.32\linewidth}
			\centering
			\smallskip
			\includegraphics[height=3cm,width=0.15\linewidth]{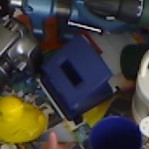}
		\end{subfigure} \hspace{-5.5mm}
		&
		\begin{subfigure} %{width=0.32\linewidth}
			\centering
			\smallskip
			\includegraphics[height=3cm,width=0.15\linewidth]{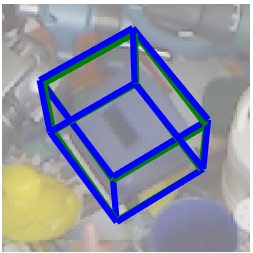}
		\end{subfigure}
		&
		\begin{subfigure} %{width=0.32\linewidth}
			\centering
			\smallskip
			\includegraphics[height=3cm,width=0.15\linewidth]{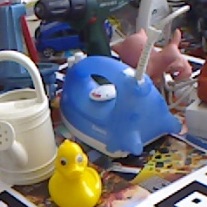}
		\end{subfigure} \hspace{-5.5mm}
		&
		\begin{subfigure} %{width=0.32\linewidth}
			\centering
			\smallskip
			\includegraphics[height=3cm,width=0.15\linewidth]{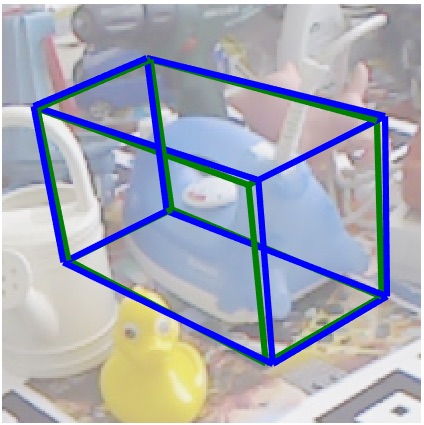}
		\end{subfigure}
		&
		\begin{subfigure} %{width=0.32\linewidth}
			\centering
			\smallskip
			\includegraphics[height=3cm,width=0.15\linewidth]{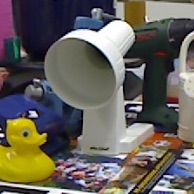}
		\end{subfigure} \hspace{-5.5mm}
		&
		\begin{subfigure} %{width=0.32\linewidth}
			\centering
			\smallskip
			\includegraphics[height=3cm,width=0.15\linewidth]{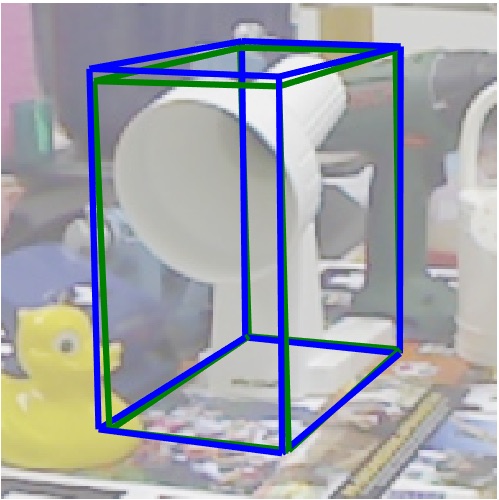}
		\end{subfigure} \\[-5mm]
		%3
		\begin{subfigure} %{width=0.32\linewidth}
			\centering
			\smallskip
			\includegraphics[height=3cm,width=0.15\linewidth]{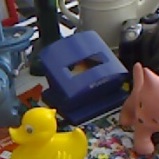}
		\end{subfigure} \hspace{-5.5mm}
		&
		\begin{subfigure} %{width=0.32\linewidth}
			\centering
			\smallskip
			\includegraphics[height=3cm,width=0.15\linewidth]{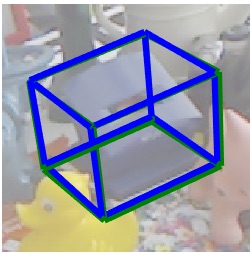}
		\end{subfigure}
		&
		\begin{subfigure} %{width=0.32\linewidth}
			\centering
			\smallskip
			\includegraphics[height=3cm,width=0.15\linewidth]{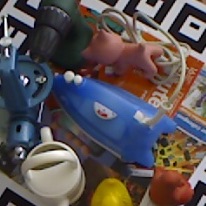}
		\end{subfigure} \hspace{-5.5mm}
		&
		\begin{subfigure} %{width=0.32\linewidth}
			\centering
			\smallskip
			\includegraphics[height=3cm,width=0.15\linewidth]{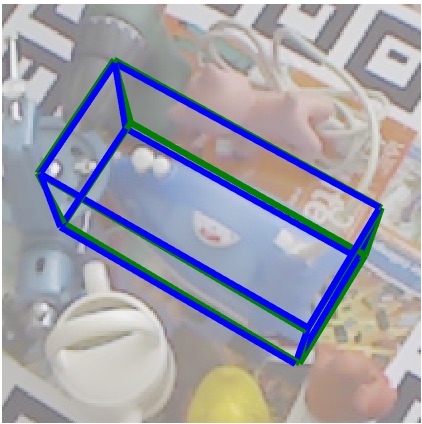}
		\end{subfigure}
		&
		\begin{subfigure} %{width=0.32\linewidth}
			\centering
			\smallskip
			\includegraphics[height=3cm,width=0.15\linewidth]{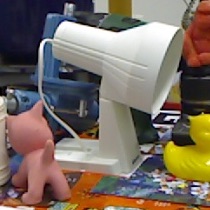}
		\end{subfigure} \hspace{-5.5mm}
		&
		\begin{subfigure} %{width=0.32\linewidth}
			\centering
			\smallskip
			\includegraphics[height=3cm,width=0.15\linewidth]{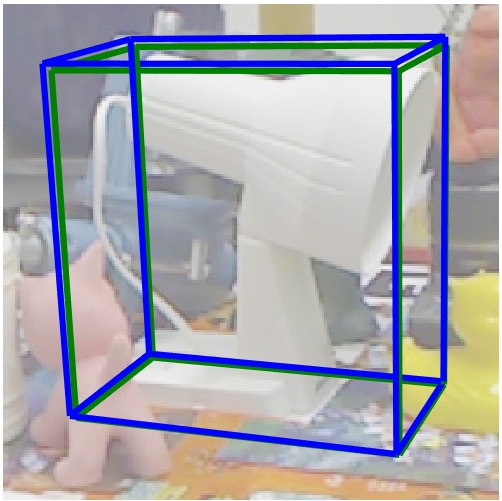}
		\end{subfigure} \\[-5mm]
		%4
		\begin{subfigure} %{width=0.32\linewidth}
			\centering
			\smallskip
			\includegraphics[height=3cm,width=0.15\linewidth]{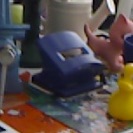}
		\end{subfigure} \hspace{-5.5mm}
		&
		\begin{subfigure} %{width=0.32\linewidth}
			\centering
			\smallskip
			\includegraphics[height=3cm,width=0.15\linewidth]{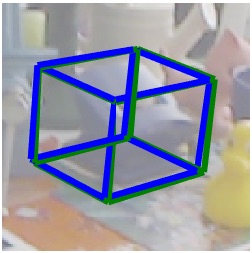}
		\end{subfigure}
		&
		\begin{subfigure} %{width=0.32\linewidth}
			\centering
			\smallskip
			\includegraphics[height=3cm,width=0.15\linewidth]{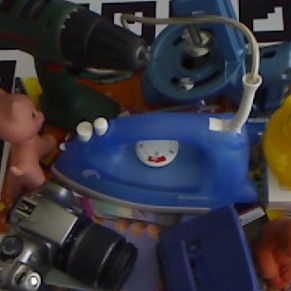}
		\end{subfigure} \hspace{-5.5mm}
		&
		\begin{subfigure} %{width=0.32\linewidth}
			\centering
			\smallskip
			\includegraphics[height=3cm,width=0.15\linewidth]{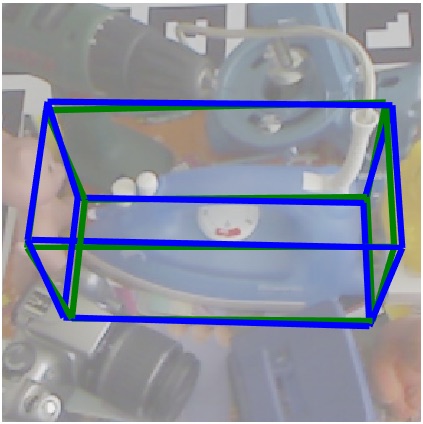}
		\end{subfigure}
		&
		\begin{subfigure} %{width=0.32\linewidth}
			\centering
			\smallskip
			\includegraphics[height=3cm,width=0.15\linewidth]{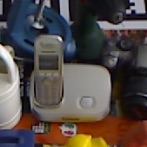}
		\end{subfigure} \hspace{-5.5mm}
		&
		\begin{subfigure} %{width=0.32\linewidth}
			\centering
			\smallskip
			\includegraphics[height=3cm,width=0.15\linewidth]{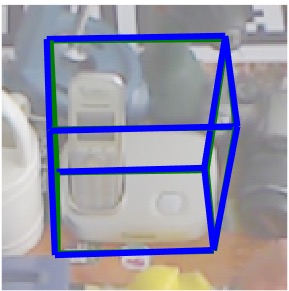}
		\end{subfigure} \\[-5mm]
		%5
		\begin{subfigure} %{width=0.32\linewidth}
			\centering
			\smallskip
			\includegraphics[height=3cm,width=0.15\linewidth]{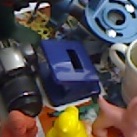}
		\end{subfigure} \hspace{-5.5mm}
		&
		\begin{subfigure} %{width=0.32\linewidth}
			\centering
			\smallskip
			\includegraphics[height=3cm,width=0.15\linewidth]{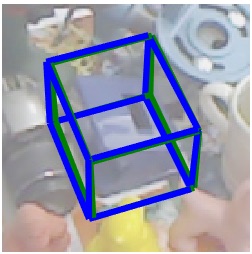}
		\end{subfigure}
		&
		\begin{subfigure} %{width=0.32\linewidth}
			\centering
			\smallskip
			\includegraphics[height=3cm,width=0.15\linewidth]{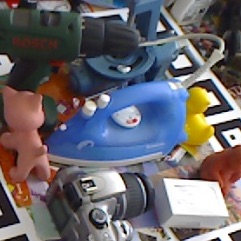}
		\end{subfigure} \hspace{-5.5mm}
		&
		\begin{subfigure} %{width=0.32\linewidth}
			\centering
			\smallskip
			\includegraphics[height=3cm,width=0.15\linewidth]{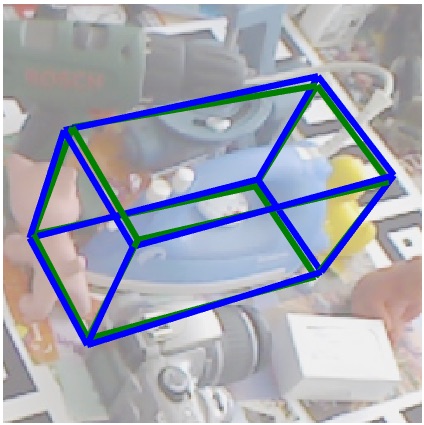}
		\end{subfigure}
		&
		\begin{subfigure} %{width=0.32\linewidth}
			\centering
			\smallskip
			\includegraphics[height=3cm,width=0.15\linewidth]{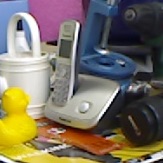}
		\end{subfigure} \hspace{-5.5mm}
		&
		\begin{subfigure} %{width=0.32\linewidth}
			\centering
			\smallskip
			\includegraphics[height=3cm,width=0.15\linewidth]{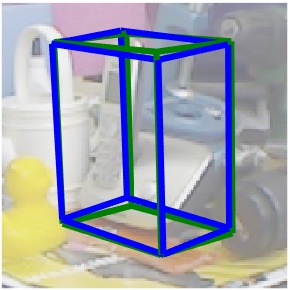}
		\end{subfigure} \\[-5mm]
		%6
		\begin{subfigure} %{width=0.32\linewidth}
			\centering
			\smallskip
			\includegraphics[height=3cm,width=0.15\linewidth]{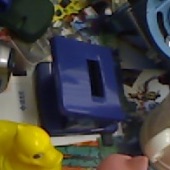}
		\end{subfigure} \hspace{-5.5mm}
		&
		\begin{subfigure} %{width=0.32\linewidth}
			\centering
			\smallskip
			\includegraphics[height=3cm,width=0.15\linewidth]{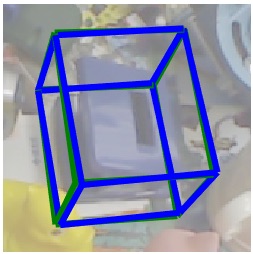}
		\end{subfigure}
		&
		\begin{subfigure} %{width=0.32\linewidth}
			\centering
			\smallskip
			\includegraphics[height=3cm,width=0.15\linewidth]{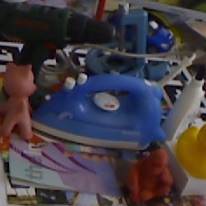}
		\end{subfigure} \hspace{-5.5mm}
		&
		\begin{subfigure} %{width=0.32\linewidth}
			\centering
			\smallskip
			\includegraphics[height=3cm,width=0.15\linewidth]{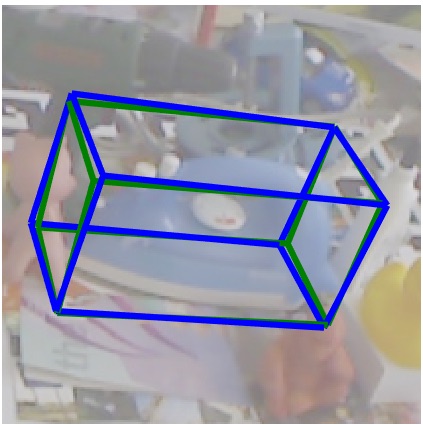}
		\end{subfigure}
		&
		\begin{subfigure} %{width=0.32\linewidth}
			\centering
			\smallskip
			\includegraphics[height=3cm,width=0.15\linewidth]{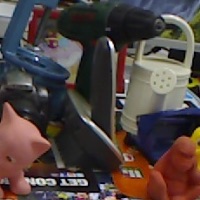}
		\end{subfigure} \hspace{-5.5mm}
		&
		\begin{subfigure} %{width=0.32\linewidth}
			\centering
			\smallskip
			\includegraphics[height=3cm,width=0.15\linewidth]{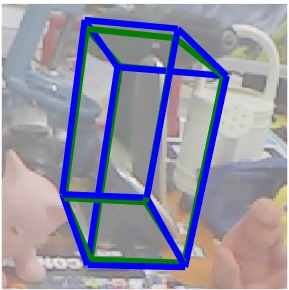}
		\end{subfigure} \\
		
	\end{tabular}
	
	\caption{Example results on the \textsc{LineMod} dataset: (left) \textsc{HolePuncher}, (middle) \textsc{Iron}, (right) ~\textsc{Lamp} and ~\textsc{Phone}. The projected 3D bounding boxes are rendered over the image and they have been cropped and resized for ease of visualization. The blue cuboid is rendered using our pose estimate whereas the green cuboid is rendered using the ground truth object pose. Note that the input image dimension is 640 $\times$ 480 pixels and the objects are often quite small. Noticeable scene clutter and occlusion makes these examples challenging.}
	\label{fig:supp_linemod4}
\end{figure*}

\paragraph{Qualitative Results.} We show qualitative results from the \textsc{Occlusion}~\cite{brachmann2014} and \textsc{LineMod}~\cite{hinterstoisser2012} datasets
in Figures~\ref{fig:supp_occlusion1} to~\ref{fig:supp_linemod4}. These examples show that our method is robust to severe occlusions, rotational
ambiguities in appearance, reflections, viewpoint change and scene clutter.
\vspace{-2mm}

%%\clearpage
%{\footnotesize
%	\bibliographystyle{ieee}
%	\bibliography{refappendix}
%}

\end{document}